# Deep Learning-Based Object Pose Estimation: A Comprehensive Survey

**Jian Liu[1,6] · Wei Sun[1] · Hui Yang[1] · Zhiwen Zeng[1] · Chongpei Liu[1,5] · Jin Zheng[2] · Xingyu Liu[3] · Hossein Rahmani[4] · Nicu Sebe[5] · Ajmal Mian[6]**

**Abstract**

Object pose estimation is a fundamental computer vision problem with broad applications in augmented reality and robotics. Over the past decade, deep learning models, due to their superior accuracy and robustness, have increasingly supplanted conventional algorithms reliant on engineered point pair features. Nevertheless, several challenges persist in contemporary methods, including their dependency on labeled training data, model compactness, robustness under challenging conditions, and their ability to generalize to novel unseen objects. A recent survey discussing the progress made on different aspects of this area, outstanding challenges, and promising future directions, is missing. To fill this gap, we discuss the recent advances in deep learning-based object pose estimation, covering all three formulations of the problem, *i.e.*, instance-level, category-level, and unseen (including both instance-unseen and category-unseen cases) object pose estimation. Our survey also covers multiple input data modalities, degrees-of-freedom of output poses, object properties, and downstream tasks, providing the readers with a holistic understanding of this field. Additionally, it discusses training paradigms of different domains, inference modes, application areas, evaluation metrics, and benchmark datasets, as well as reports the performance of current state-of-the-art methods on these benchmarks, thereby facilitating the readers in selecting the most suitable method for their application. Finally, the survey identifies key challenges, reviews the prevailing trends along with their pros and cons, and identifies promising directions for future research. We cover the literature up to our submission date and will continue to follow the latest works at https://github.com/CNJianLiu/Awesome-Object-Pose-Estimation.

Jian Liu
jianliu@hnu.edu.cn

✉ Wei Sun
wei_sun@hnu.edu.cn

✉ Hui Yang
huiyang@hnu.edu.cn

Zhiwen Zeng
zingaltern@hnu.edu.cn

Chongpei Liu
chongpei56@hnu.edu.cn

Jin Zheng
zheng.jin@csu.edu.cn

Xingyu Liu
liuxy21@mails.tsinghua.edu.cn

Hossein Rahmani
h.rahmani@lancaster.ac.uk

Nicu Sebe
sebe@disi.unitn.it

Ajmal Mian
ajmal.mian@uwa.edu.au

[1] School of Artificial Intelligence and Robotics, Hunan University, Changsha 410082, Hunan, China

[2] School of Architecture and Art, Central South University, Changsha 410083, Hunan, China

[3] Department of Automation, Tsinghua University, Beijing 100084, China

[4] School of Computing and Communications, Lancaster University, Lancaster LA1 4YW, United Kingdom

[5] Department of Information Engineering and Computer Science, University of Trento, Trento 38123, Italy

[6] Department of Computer Science, The University of Western Australia, Perth WA 6009, Australia

## 1 Introduction

Object pose estimation is a fundamental computer vision problem that aims to estimate the pose of an object in a given image relative to the camera that captured the image. Object pose estimation is a crucial technology for augmented reality (Liu et al., 2022f; He et al., 2022a; Wen et al., 2024), robotic manipulation (Liu et al., 2023d, 2024a), hand-object interaction (Lin et al., 2023b; Rezazadeh et al., 2023), etc. Depending on the application needs, the object pose is estimated up to varying degrees of freedom (DoF) such as 3DoF that only includes 3D rotation, 6DoF that additionally includes 3D translation, or 9DoF which includes estimating the 3D size of the object besides the 3D rotation and 3D translation.

In the pre-deep learning era, many hand-crafted feature-based approaches such as SIFT (Lowe, 2004), FPFH (Rusu et al., 2009), VFH (Rusu et al., 2010), and Point Pair Features (PPF) (Drost et al., 2010; Choi & Christensen, 2012; Choi et al., 2013; Birdal & Ilic, 2015) were designed for object pose estimation. However, these methods exhibit deficiencies in accuracy and robustness when confronted with complex scenes (Xiang et al., 2017; Wang et al., 2019a). These traditional methods have now been supplanted by data driven deep learning-based approaches that harness the power of deep neural networks to learn high-dimensional feature representations from data, leading to improved accuracy and robustness to handle complex environments.

Deep learning-based methods can be divided into instance-level, category-level, and unseen (including both instance-unseen and category-unseen cases) object pose estimation according to the problem formulation. Fig. 1 shows a comparison of the three methods. Early methods were mainly instance-level (Wang et al., 2019a; Peng et al., 2019; He et al., 2020; Li et al., 2019; Zakharov et al., 2019), trained to estimate the pose of specific object instances. Instance-level methods can be further divided into correspondence-based, template-based, voting-based, and regression-based methods. Since instance-level methods are trained on instance-specific data, they can estimate pose with high precision for the given object instances. However, their generalization performance is poor because they are meant to be applied only to the instances on which they are trained. Moreover, many instance-level methods (He et al., 2020, 2021) require CAD models of the objects. Recognizing these limitations, Wang et al. (2019b) proposed the first category-level object pose and size estimation method. They generalize to intra-class unseen objects without necessitating retraining and employing CAD models during inference. Subsequent category-level methods (Lin et al., 2022c; Di et al., 2022; Zheng et al., 2023; Liu et al., 2023c, 2025b) can be divided into shape prior-based and shape prior-free methods. While improving the generalization ability within a category, these category-level

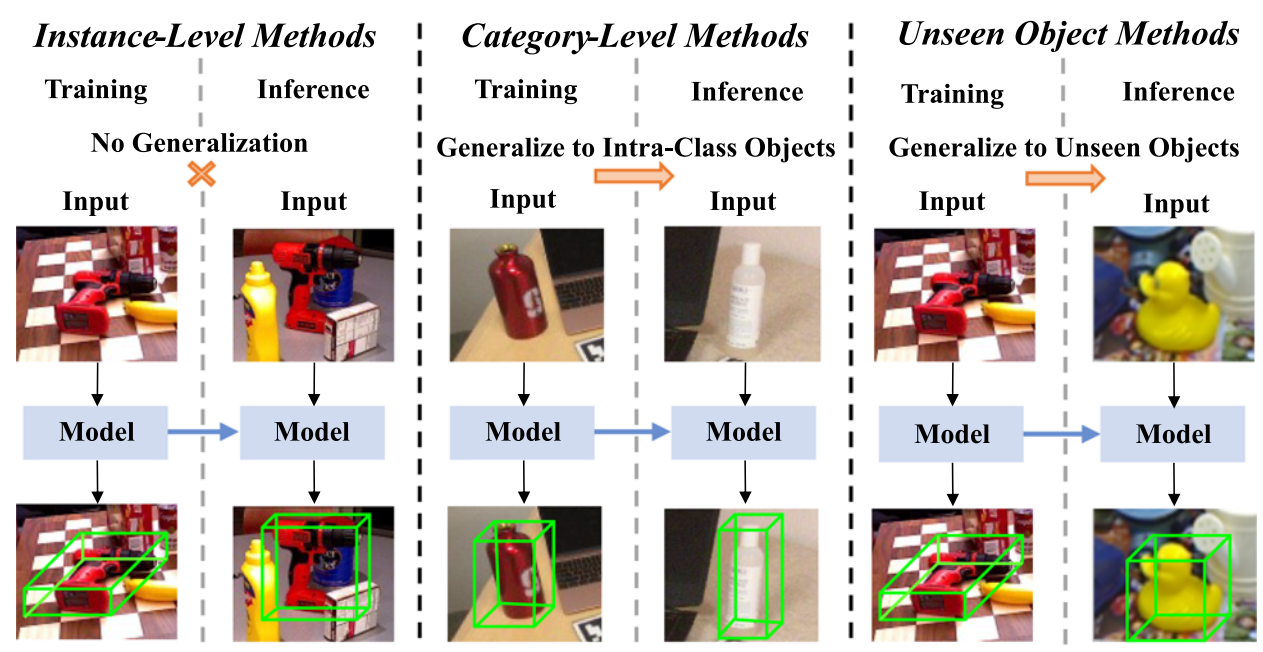

**Fig. 1** Comparison of instance-level, category-level, and unseen (including both instance-unseen and category-unseen cases) object pose estimation methods. Instance-level methods can only estimate the pose of specific object instances on which they are trained. Category-level methods can generalize to unseen instances within a known category, rather than being limited to the specific training instances. In contrast, unseen object pose estimation methods aim for stronger generalization, handling both novel instances and entirely new object categories not encountered during training

methods still need to collect and label extensive training data for each object category. Moreover, these methods cannot generalize to unseen object categories. To this end, some unseen object pose estimation methods have been recently proposed (Liu et al., 2022f; Labbé et al., 2022; Nguyen et al., 2024b; Lin et al., 2024a; Wen et al., 2024), which can be further classified into CAD model-based and manual reference view-based methods. These methods further enhance the generalization of object pose estimation, *i.e.*, they can be generalized to unseen objects without retraining. Nevertheless, they still need to obtain the object CAD model or annotate a few reference images of the object.

Although significant progress has been made in the area of object pose estimation, several challenges persist in current methods, such as the reliance on labeled training data, difficulty in generalizing to novel unseen objects, model compactness, and robustness in challenging scenarios. To enable readers to swiftly grasp the current state-of-the-art (SOTA) in object pose estimation and facilitate further research in this direction, it is crucial to provide a thorough review of all the relevant problem formulations. A close examination of the existing academic literature reveals a significant gap when reviewing the various problem formulations in object pose estimation. Current prevailing reviews (Hoque et al., 2021; Marullo et al., 2023; Fan et al., 2022b; Du & Wang, 2021; Guan et al., 2024) tend to exhibit a narrow focus, either confined to particular input modalities (Marullo et al., 2023; Fan et al., 2022b) or tethered to specific application domains (Du & Wang, 2021; Guan et al., 2024). Furthermore, these reviews predominantly scrutinize instance-level and category-level methods, thus neglecting the exploration of the most practical problem formulation in the domain which is unseen object pose estimation. This hinders readers

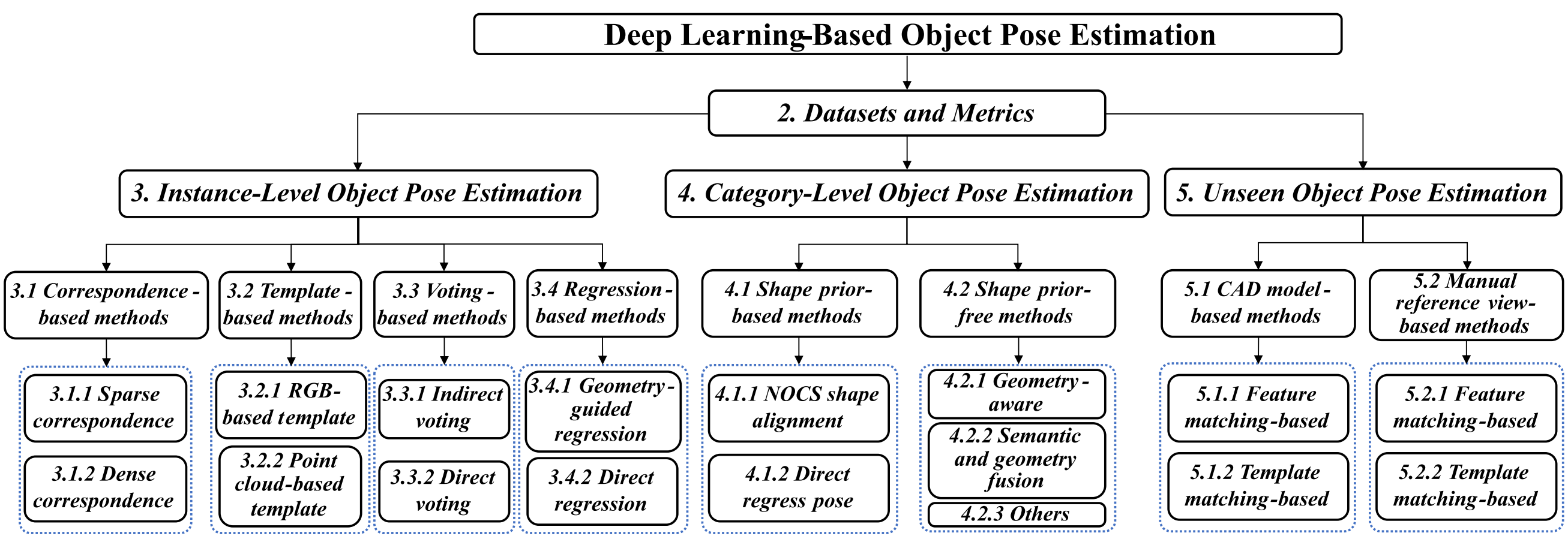

**Fig. 2** A taxonomy of this survey. Firstly, we review the datasets and evaluation metrics used to evaluate object pose estimation. Next, we review the deep learning-based methods by dividing them into three categories: instance-level, category-level, and unseen object pose estimation. Instance-level methods can be further classified into correspondence-based, template-based, voting-based, and regression-based methods. Category-level methods can be further divided into shape prior-based and shape prior-free methods. Unseen object pose estimation methods can be further classified into CAD model-based and manual reference view-based methods

from gaining a comprehensive understanding of the area. For instance, Fan et al. (2022b) provided valuable insights into RGB image-based object pose estimation. However, their focus is limited to a singular modality, hindering readers from comprehensively understanding methods across various input modalities. Conversely, Du and Wang (2021) exclusively examined object pose estimation within the context of the robotic grasping task, which limits the readers to understand object pose estimation only from the perspective of a single specific application.

To address the above problems, we present here a comprehensive survey of recent advancements in deep learning-based methods for object pose estimation. Our survey encompasses all problem formulations, including instance-level, category-level, and unseen object pose estimation, aiming to provide readers with a holistic understanding of this field. Additionally, we discuss different domain training paradigms, application areas, evaluation metrics, and benchmark datasets, as well as report the performance of state-of-the-art methods on these benchmarks, aiding readers in selecting suitable methods for their applications. Furthermore, we also highlight prevailing trends and discuss their strengths and weaknesses, as well as identify key challenges and promising avenues for future research. The taxonomy of this survey is shown in Fig. 2. Our main contributions and highlights are as follows:

- We present a *comprehensive survey* of deep learning-based object pose estimation methods. This is the *first* survey that covers all three problem formulations in the domain, including instance-level, category-level, and unseen object pose estimation.
- Our survey covers popular input data modalities (RGB images, depth images, RGBD images), the different degrees of freedom (3DoF, 6DoF, 9DoF) in output poses, object properties (rigid, articulated) for the task of pose estimation as well as tracking. It is crucial to cover all these aspects in a single survey to give a complete picture to readers, an aspect overlooked by existing surveys which only cover a few of these aspects.
- We discuss different domain training paradigms, inference modes, application areas, evaluation metrics, and benchmark datasets as well as report the performance of existing SOTA methods on these benchmarks to help readers choose the most appropriate ones for deployment in their application.
- We highlight popular trends in the evolution of object pose estimation techniques over the past decade and discuss their strengths and weaknesses. We also identify key challenges that are still outstanding in object pose estimation along with promising research directions to guide future efforts.

The rest of this article is organized as follows. Sec. 2 reviews the datasets and metrics used to evaluate the three categories of object pose estimation methods. We then review instance-level methods in Sec. 3, category-level methods in Sec. 4, and unseen object pose estimation methods in Sec. 5. In the aforementioned three sections, we also discuss the training paradigms, inference modes, challenges, and popular trends associated with representative methods in the particular category. Next, Sec. 6 reviews the common applications of object pose estimation. Finally, Sec. 7 provides an

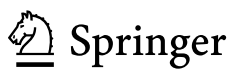

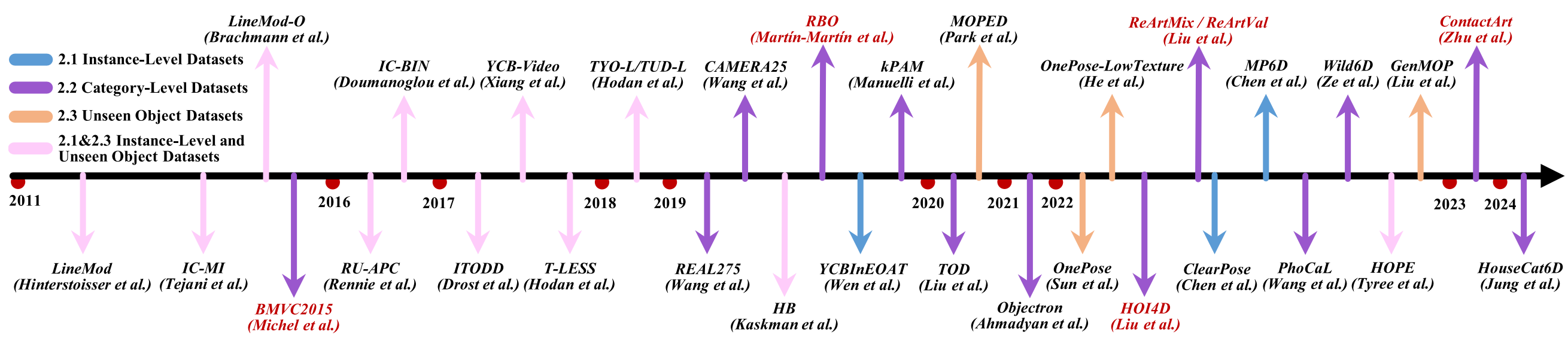

**Fig. 3** Chronological overview of the datasets for object pose estimation evaluation. Notably, the pink arrows represent the BOP Challenge datasets, which can be used to evaluate both instance-level and unseen object pose estimation methods. The red references represent the datasets of articulated objects. From this, we can also see the development trend in the field of object pose estimation, *i.e.*, from instance-level methods to category-level and unseen object pose estimation methods

outlook on emerging trends and future research directions based on the challenges in the field.

## 2 Datasets and Metrics

The advancement of deep learning-based object pose estimation is closely linked to the creation and utilization of challenging and trustworthy large-scale datasets. This section introduces commonly used mainstream object pose estimation datasets, categorized into instance-level, category-level, and unseen object pose estimation methods based on problem formulation. The chronological overview is shown in Fig. 3. In addition, we also conduct an overview of the related evaluation metrics.

### 2.1 Datasets for Instance-Level Methods

Since the BOP Challenge datasets (Hodan et al., 2024) are currently the most popular datasets for the evaluation of instance-level methods, we divide the instance-level datasets into BOP Challenge and other datasets for overview.

#### 2.1.1 BOP Challenge Datasets

**Linemod Dataset (LM)** (Hinterstoisser et al., 2012) comprises 15 RGBD sequences containing annotated RGBD images with ground-truth 6DoF object poses, object CAD models, 2D bounding boxes, and binary masks. Typically following (Brachmann et al., 2016), approximately 15% of images from each sequence are allocated for training, with the remaining 85% reserved for testing. These sequences present challenging scenarios with cluttered scenes, texture-less objects, and varying lighting conditions, making accurate object pose estimation difficult.

**Linemod Occlusion Dataset (LM-O)** (Brachmann et al., 2014) is an extension of the LM dataset (Hinterstoisser et al., 2012) specifically designed to evaluate the performance in occlusion scenarios. This dataset consists of 1214 RGBD images from the basic sequence in the LM dataset for 8 heavily occluded objects. It is critical for evaluating and improving pose estimation algorithms in complex environments characterized by occlusion.

**IC-MI** (Tejani et al., 2014) **/ IC-BIN Dataset** (Doumanoglou et al., 2016) contribute to texture-less object pose estimation. IC-MI comprises six objects: 2 texture-less and 4 textured household item models. IC-BIN dataset is specifically designed to address challenges posed by clutter and occlusion in robot garbage bin picking scenarios. IC-BIN includes 2 objects from the IC-MI.

**RU-APC Dataset** (Rennie et al., 2016) aims to tackle challenges in warehouse picking tasks and provides rich data for evaluating and improving the perception capabilities of robots in a warehouse automation context. The dataset comprises 10,368 registered depth and RGB images, covering 24 types of objects, which are placed in various poses within different boxes on warehouse shelves to simulate diverse experimental conditions.

**YCB-Video Dataset (YCB-V)** (Xiang et al., 2017) comprises 21 objects distributed across 92 RGBD videos, each video containing 3 to 9 objects from the YCB object dataset (Calli et al., 2015) (totaling 50 objects). It includes 133,827 frames with a resolution of 640×480, making it well-suited for both object pose estimation and tracking tasks.

**T-LESS Dataset** (Hodan et al., 2017) is an RGBD dataset designed for texture-less objects commonly found in industrial settings. It includes 30 electrical objects with no obvious texture or distinguishable color properties. In addition, it includes images of varying resolutions. In the training set, images predominantly feature black backgrounds, while the test set showcases diverse backgrounds with varying lighting conditions and occlusions. T-LESS is challenging because of the absence of texture on objects and the intricate environmental settings.

**ITODD Dataset** (Drost et al., 2017) includes 28 real-world industrial objects distributed across over 800 scenes with around 3,500 images. This dataset leverages two industrial 3D sensors and three high-resolution grayscale cameras to enable multi-angle observation of the scenes, providing comprehensive and detailed data for industrial object analysis and evaluation.

**TYO-L / TUD-L Dataset** (Hodan et al., 2024) focus on different lighting conditions. Specifically, TYO-L provides observation of 3 objects under 8 lighting conditions. These scenes are designed to evaluate the robustness of pose estimation algorithms to lighting variations. Unlike TYO-L, the data collection method of TUD-L involves fixing the camera and manually moving the object, providing a more realistic representation of the object's physical movement.

**HB Dataset** (Kaskman et al., 2019) covers various scenes with changes in occlusion and lighting conditions. It comprises 33 objects, including 17 toys, 8 household items, and 8 industry-related objects, distributed across 13 diverse scenes.

**HOPE Dataset** (Tyree et al., 2022) is specifically designed for household objects, containing 28 toy grocery objects. The HOPE-Image dataset includes objects from 50 scenes across 10 home/office environments. Each scene includes up to 5 lighting variations, such as backlit and obliquely directed lighting, with shadow-casting effects. Additionally, the HOPE-Video dataset comprises 10 video sequences totaling 2,038 frames, with each scene showcasing between 5 to 20 objects.

#### 2.1.2 Other Datasets

**YCBInEOAT Dataset** (Wen et al., 2020) is designed for RGBD-based object pose tracking in robotic manipulation. It contains the ego-centric RGBD videos of a dual-arm robot manipulating the YCB objects (Calli et al., 2015). There are 3 types of manipulation: single-arm pick-and-place, within-arm manipulation, and pick-and-place between arms. This dataset comprises annotations of ground-truth poses across 7449 frames, encompassing 5 distinct objects depicted in 9 videos.

**ClearPose Dataset** (Chen et al., 2022d) is designed for transparent objects, which are widely prevalent in daily life, presenting significant challenges to visual perception and sensing systems due to their indistinct texture features and unreliable depth information. It encompasses over 350K real-world RGBD images and 5M instance annotations across 63 household objects.

**MP6D Dataset** (Chen et al., 2022c) is an RGBD dataset designed for object pose estimation of metal parts, featuring 20 texture-less metal components. It consists of 20,100 real-world images with object pose labels collected from various scenarios as well as 50K synthetic images, encompassing cluttered and occluded scenes.

### 2.2 Datasets for Category-Level Methods

In this part, we divide the category-level datasets into rigid and articulated object datasets for elaboration.

#### 2.2.1 Rigid Objects Datasets

**CAMERA25 Dataset** (Wang et al., 2019b) incorporates 1085 instances across 6 object categories: bowl, bottle, can, camera, mug, and laptop. Notably, the object CAD models in CAMERA25 are sourced from the synthetic ShapeNet dataset (Chang et al., 2015). Each image within this dataset contains multiple instances, accompanied by segmentation masks and 9DoF pose labels.

**REAL275 Dataset** (Wang et al., 2019b) is a real-world dataset comprising 18 videos and approximately 8K RGBD images. The dataset is divided into three subsets: a training set (7 videos), a validation set (5 videos), and a testing set (6 videos). It includes 42 object instances across 6 categories, consistent with those in the CAMERA25 dataset. REAL275 is a prominent real-world dataset extensively used for category-level object pose estimation in academic research.

**kPAM Dataset** (Manuelli et al., 2019) is tailored specifically for robotic applications, emphasizing the use of keypoints. Notably, it adopts a methodology involving 3D reconstruction followed by manual keypoint annotation on these reconstructions. With a total of 117 training sequences and 245 testing sequences, kPAM offers a substantial collection of data for training and evaluating algorithms related to robotic perception and manipulation.

**TOD Dataset** (Liu et al., 2020) consists of 15 transparent objects categorized into 6 classes, each annotated with pertinent 3D keypoints. It encompasses a vast collection of 48K stereo and RGBD images capturing both transparent and opaque depth variations. The primary focus of the TOD dataset is on transparent 3D object applications, providing essential resources for tasks such as object detection and pose estimation in challenging scenarios involving transparency.

**Objectron Dataset** (Ahmadyan et al., 2021) contains 15K annotated video clips with over 4M labeled images belonging to categories of bottles, books, bikes, cameras, chairs, cereal boxes, cups, laptops, and shoes. This dataset is sourced from 10 countries spanning 5 continents, ensuring diverse geographic representation. Due to its extensive content, it is highly advantageous for evaluating the RGB-based category-level object pose estimation and tracking methods.

**Wild6D Dataset** (Ze & Wang, 2022) is a substantial real-world dataset used to assess self-supervised category-level object pose estimation methods. It offers annotations exclusively for 486 test videos with diverse backgrounds, showcasing 162 objects across 5 categories (excluding the "can" category found in CAMERA25 and REAL275).

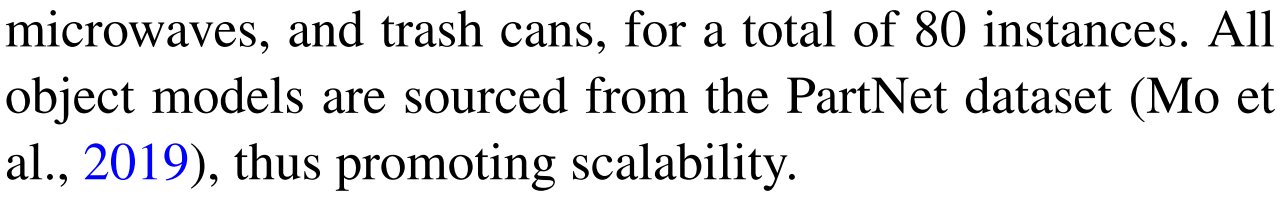

**PhoCaL Dataset** (Wang et al., 2022b) incorporates both RGBD and RGB-P (Polarisation) modalities. It consists of 60 meticulously crafted 3D models representing household objects, including symmetric, transparent, and reflective items. PhoCaL focuses on 8 specific object categories across 24 sequences, deliberately introducing challenges such as occlusion and clutter.
**HouseCat6D Dataset** (Jung et al., 2024) is a comprehensive dataset designed for multi-modal category-level object pose estimation and grasping tasks. The dataset encompasses a wide range of household object categories, featuring 194 high-quality 3D models. It includes objects of varying photometric complexity, such as transparent and reflective items, and spans 41 scenes with diverse viewpoints. The dataset is specifically curated to address challenges in object pose estimation, including occlusions and the absence of markers, making it suitable for evaluating algorithms under real-world conditions.

#### 2.2.2 Articulated Objects Datasets

**BMVC Dataset** (Michel et al., 2015) includes 4 articulated objects: laptop, cabinet, cupboard, and toy train. Each object is modeled as a motion chain comprising components and interconnected heads. Joints are constrained to one rotational and one translational DoF. This dataset provides CAD models and accompanying text files detailing the topology of the underlying motion chain structure for each object.
**RBO Dataset** (Martín-Martín et al., 2019) contains 14 commonly found articulated objects in human environments, with 358 interaction sequences, resulting in a total of 67 minutes of manual manipulation under different experimental conditions, including changes in interaction type, lighting, viewpoint, and background settings.
**HOI4D Dataset** (Liu et al., 2022e) is pivotal for advancing research in category-level human-object interactions. It comprises 2.4M RGBD self-centered video frames depicting interactions between over 9 participants and 800 object instances. These instances are divided into 16 categories, including 7 rigid and 9 articulated objects.
**ReArtMix / ReArtVal Datasets** (Liu et al., 2022c) are formulated to tackle the challenge of partial-level multiple articulated objects pose estimation featuring unknown kinematic structures. The ReArtMix dataset encompasses over 100,000 RGBD images rendered against diverse background scenes. The ReArtVal dataset consists of 6 real-world desktop scenes comprising over 6,000 RGBD frames.
**ContactArt Dataset** (Zhu et al., 2024) is generated using a remote operating system (Qin et al., 2022a) to manipulate articulated objects in a simulation environment. This system utilizes smartphones and laptops to precisely annotate poses and contact information. This dataset contains 5 prevalent categories of articulated objects: laptops, drawers, safes, microwaves, and trash cans, for a total of 80 instances. All object models are sourced from the PartNet dataset (Mo et al., 2019), thus promoting scalability.

### 2.3 Datasets for Unseen Object Pose Estimation

The current mainstream datasets for evaluating unseen objects are the BOP Challenge datasets, as discussed in Sec. 2.1.1. Besides these BOP Challenge datasets, there are also some datasets designed for evaluating manual reference view-based methods as follows.
**MOPED Dataset** (Park et al., 2020a) is a model-free object pose estimation dataset featuring 11 household objects. It includes reference and test images that encompass all views of the objects. Each object in the test sequences is depicted in five distinct environments, with approximately 300 test images per object.
**GenMOP Dataset** (Liu et al., 2022f) includes 10 objects ranging from flat objects to thin structure objects. For each object, there are two video sequences collected from various backgrounds and lighting situations. Each video sequence consists of approximately 200 images.
**OnePose Dataset** (Sun et al., 2022b) comprises over 450 real-world video sequences of 150 objects. These sequences are collected in a variety of background conditions and capture all angles of the objects. Each environment has an average duration of 30 seconds. The dataset is randomly partitioned into training and validation sets.
**OnePose-LowTexture Dataset** (He et al., 2022a) is introduced as a complement to the testing set of the existing OnePose dataset (Sun et al., 2022b), which predominantly features textured objects. This dataset comprises 40 household objects with low texture. For each object, there are two video sequences: one serving as the reference video and the other for testing. Each video is captured at a resolution of 1920×1440, 30 Frames Per Second (FPS), and approximately 30 seconds in duration.

### 2.4 Metrics

In this part, we divide the metrics into 3DoF, 6DoF, 9DoF, and other evaluation metrics for a comprehensive overview.

#### 2.4.1 3DoF Evaluation Metrics

The geodesic distance (Wohlhart and Lepetit, 2015) between the ground-truth and predicted 3D rotations is a commonly used 3DoF pose estimation metric. Calculating the angle error between two rotation matrices can visually evaluate their relative deviation. It can be formulated as follows:

$$d\left(R_{gt}, R\right) = \arccos\left(\frac{tr\left(R_{gt}{}^{\top} R\right)-1}{2}\right)/\pi \,, \tag{1}$$

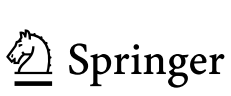

where $R_{gt}$ and $R$ denote the ground-truth and predicted 3D rotations, respectively. $\top$ represents matrix transpose. $tr$ denotes the trace of a matrix, which refers to the sum of the elements on the main diagonal. Typically, 3D rotation estimation accuracy is defined as the percentage of objects whose angle error is below a specific threshold and whose predicted class is correct. It can be expressed as follows:

$$Acc. = \begin{cases} 1, & \text{if } d(R_{gt}, R) < \lambda \text{ and } c = c_{gt} \\ 0, & \text{otherwise} \end{cases}, \quad (2)$$

where $c$, $c_{gt}$, and $\lambda$ denote the predicted class, ground-truth class and predefined threshold, respectively.

### 2.4.2 6DoF Evaluation Metrics

Currently, the BOP metric (*BOP-M*) (Hodan et al., 2024) is the most popular metric, which is the Average Recall (*AR*) of the Visible Surface Discrepancy (VSD), Maximum Symmetry-Aware Surface Distance (MSSD), and Maximum Symmetry-Aware Projection Distance (MSPD) metrics. Specifically, the **VSD** (Hodan et al., 2024) metric treats poses that are indistinguishable in shape as equivalent by only measuring the misalignment of the visible object surface. It can be expressed as follows:

$$e_{VSD}\left(\hat{D}, \bar{D}, \hat{V}, \bar{V}, \tau\right) = avg_{p \in \hat{V} \cup \bar{V}} \begin{cases} 0, & \text{if } p \in \hat{V} \cap \bar{V} \wedge |\hat{D}(p) - \bar{D}(p)| < \tau \\ 1, & \text{otherwise} \end{cases}, \quad (3)$$

where the symbols $\hat{D}$ and $\bar{D}$ represent distance maps generated by rendering the object model $M$ in two different poses: $\hat{P}$ (an estimated pose) and $\bar{P}$ (the ground-truth pose), respectively. In these maps, each pixel $p$ stores the distance from the camera center to a 3D point $\mathrm{x}_p$ that projects onto $p$. These distance values are derived from depth maps, which are typical outputs of sensors like Kinect, containing the $Z$ coordinate of $\mathrm{x}_p$. These distance maps are compared with the distance map $D_I$ of the test image $I$ to derive visibility masks $\hat{V}$ and $\bar{V}$. These masks identify pixels where the model $M$ is visible in the image $I$. The parameter $\tau$ represents the tolerance for misalignment. In addition, the **MSSD** (Hodan et al., 2024) metric is a suitable factor for determining the likelihood of successful robotic manipulation and is not significantly affected by object geometry or surface sampling density. It can be formulated as follows:

$$e_{MSSD}\left(\hat{P}, \bar{P}, S_M, V_M\right) = \min_{S \in S_M} \max_{\mathrm{x} \in V_M} \left\| \hat{P}\mathrm{x} - \bar{P}S\mathrm{x} \right\|_2, \quad (4)$$

where the set $S_M$ comprises global symmetry transformations for the object model $M$, while $V_M$ represents the vertices of the model. Furthermore, the **MSPD** (Hodan et al., 2024) metric is ideal for evaluating RGB-only methods in augmented reality, focusing on perceivable discrepancies and excluding alignment along the optical (Z) axis, which can be represented as follows:

$$e_{MSPD}\left(\hat{P}, \bar{P}, S_M, V_M\right) = \min_{S \in S_M} \max_{\mathrm{x} \in V_M} \left\| proj\left(\hat{P}\mathrm{x}\right) - proj\left(\bar{P}S\mathrm{x}\right) \right\|_2, \quad (5)$$

where the function $proj()$ represents the 2D projection (pixel-level), and the other symbols have the same meanings as in MSSD.

Besides the *BOP-M*, the average point distance (ADD) (Hinterstoisser et al., 2012) and average closest point distance (ADD-S) (Hinterstoisser et al., 2012) are also commonly leveraged to evaluate the performance of 6DoF object pose estimation. They can intuitively quantify the geometric error between the estimated and the ground-truth poses by computing the average distance between corresponding points on the object CAD model. Specifically, the ADD metric is designed for asymmetric objects, while the ADD-S metric is designed explicitly for symmetric objects. Given the ground-truth rotation $R_{gt}$ and translation $t_{gt}$, as well as the estimated rotation $R$ and translation $t$, ADD calculates the average pairwise distance between the 3D model points $x \in O$ corresponding to the transformation between the ground truth and estimated pose:

$$ADD = \underset{x \in O}{avg} \left\| \left(R_{gt}x + t_{gt}\right) - (Rx + t) \right\|. \quad (6)$$

For symmetric objects, the matching between points in certain views is inherently ambiguous. Therefore, the average distance is calculated using the nearest point distance as follows:

$$ADD{-}S = \underset{x_1 \in O}{avg} \min_{x_2 \in O} \left\| \left(R_{gt}x_1 + t_{gt}\right) - (Rx_2 + t) \right\|. \quad (7)$$

Meanwhile, the area under the ADD and the ADD-S curve (AUC) are often leveraged for evaluation. Specifically, if the ADD and ADD-S are smaller than a given threshold, the predicted pose will be considered correct. Moreover, there are many methods (Wang et al., 2020b; Jiang et al., 2022; Li et al., 2022) that evaluate asymmetric and symmetric objects using ADD and ADD-S, respectively. This metric is termed ADD(S).

In addition, $n^\circ\, m$cm (Shotton et al., 2013) is also currently a prevalent evaluation metric (especially in category-level object pose estimation). It directly quantifies the errors in predicted 3D rotation and 3D translation. An object pose prediction is deemed correct if its rotation error is below threshold $n^\circ$ and its translation error is below threshold $m$cm. It can be defined as an indicator function as follows:

$$I_{n^\circ m\text{cm}}(e_R, e_t) = \begin{cases} 1, & \text{if } e_R < n^\circ \text{ and } e_t < m\text{cm} \\ 0, & \text{otherwise} \end{cases}, \quad (8)$$

where $e_R$ and $e_t$ represent the rotation and translation errors between the estimated and ground-truth values, respectively.

Furthermore, compared to directly comparing 6DoF pose in 3D space, the simplicity and practicality of the 2D Projection metric (Brachmann et al., 2016) make it suitable for evaluation as well, which quantifies the average distance between CAD model points when projected under the estimated object pose and the ground-truth pose. A pose is considered correct if the projected distances are less than 5 pixels.

#### 2.4.3 9DoF Evaluation Metric

$IoU_{3D}$ denotes the Intersection-over-Union (IoU) (Wang et al., 2019b) percentage between the ground-truth and predicted 3D bounding boxes, which can evaluate the 6DoF pose estimation as well as the 3DoF size estimation. It can be expressed as:

$$IoU_{3D} = \frac{P_B \cap G_B}{P_B \cup G_B}, \quad (9)$$

where $G_B$ and $P_B$ represent the ground-truth and the predicted 3D bounding boxes, respectively. The symbols $\cap$ and $\cup$ represent the intersection and union, respectively. The correctness of the predicted object pose is determined based on whether the $IoU_{3D}$ value exceeds a predefined threshold.

#### 2.4.4 Other Metric

Since some Normalized Object Coordinate Space (NOCS) shape alignment-based category-level methods reconstruct the 3D object shape before estimating the object pose, the Chamfer Distance (CD) metric (Tian et al., 2020a), which not only captures the global shape deviation but is also sensitive to local shape differences, is commonly leveraged to evaluate the NOCS shape reconstruction accuracy of these methods as follows:

$$D_{cd} = \sum_{x \in N} \min_{y \in M_{gt}} \| x - y \|_2^2 + \sum_{y \in M_{gt}} \min_{x \in N} \| x - y \|_2^2, \quad (10)$$

where $N$ and $M_{gt}$ represent the reconstructed and ground-truth NOCS shape, respectively.

## 3 Instance-Level Object Pose Estimation

Instance-level object pose estimation describes the task of estimating the pose of the objects that have been seen during the training of the model. We classify existing instance-level methods into four categories: correspondence-based (Sec. 3.1), template-based (Sec. 3.2), voting-based (Sec. 3.3), and regression-based (Sec. 3.4) methods.

### 3.1 Correspondence-Based Methods

Correspondence-based object pose estimation refers to techniques that involve identifying correspondences between the input data and the given complete object CAD model. Correspondence-based methods can be divided into sparse (Sec. 3.1.1) and dense (Sec. 3.1.2) correspondences. The illustration of these two types of methods is shown in Fig. 4. The attributes and performance of some representative methods are shown in Table 1.

#### 3.1.1 Sparse Correspondence Methods

Sparse correspondence-based methods involve detecting object keypoints in the input image or point cloud to establish 2D-3D or 3D-3D correspondences between the input data and the object CAD model, followed by the Perspective-n-Point (PnP) algorithm (Fischler & Bolles, 1981) or least square method to determine the object pose. As a representative method, Rad and Lepetit (2017) first used a segmentation method to detect the object of interest in an RGB image. Then, they predicted the 2D projections of the object's 3D

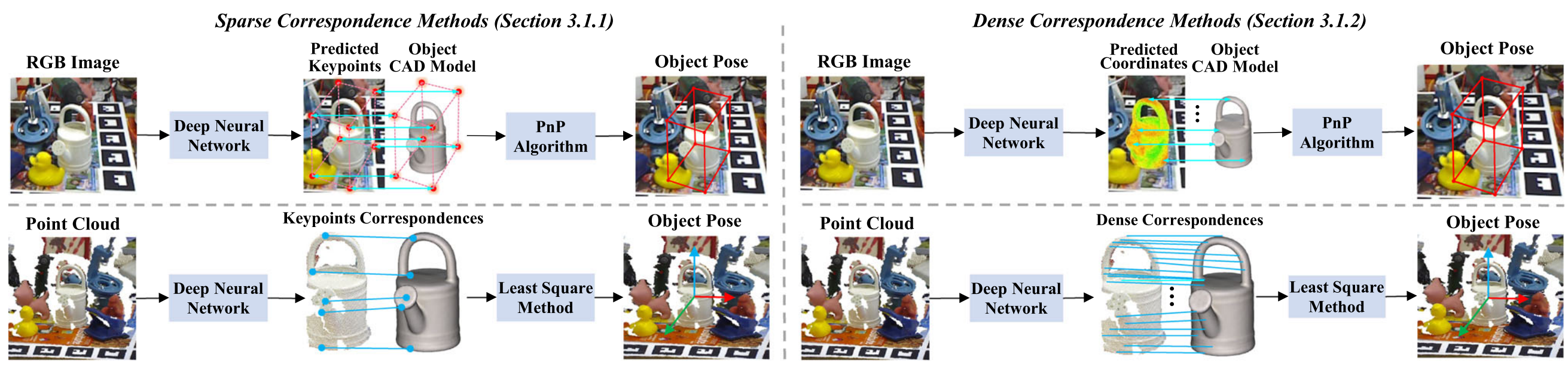

**Fig. 4** Illustration of the correspondence-based methods (Sec. 3.1), which involve establishing correspondences between input data and a provided object CAD model

**Table 1** Representative instance-level methods. For each method, we report its 10 properties: published year, training input, inference input, pose DoF (3DoF, 6DoF, and 9DoF), object property (rigid, articulated), task (estimation, tracking, and refinement), domain training paradigm (source domain, domain adaptation, and domain generalization), inference mode, application area, and its performance of key metrics on key datasets. Notably, for the input of training and inference, we only focus on the input of the pose estimation model, not the input of the front-end segmentation method (because it can be obtained through RGB as well as through depth or RGBD). D, S, C, T, V, P, and R denote object detection, instance segmentation, correspondence prediction, template matching, voting, pose solution/regression, and pose refinement, respectively. We report the average recall of ADD(S) within 10% of the object diameter (termed ADD(S)-0.1d) of LM-O and LM datasets, and the AUC of ADD-S (<0.1m) of YCB-V dataset (Sec. 2)

| Methods | | | Published Year | Training Input | Inference Input | Pose DoF | Object Property | Task | Domain Training Paradigm | Inference Mode | Application Area | LM-O | LM ADD(S)-0.1d | YCB-V ADD-S (<0.1m) |
|---|---|---|---|---|---|---|---|---|---|---|---|---|---|---|
| Correspondence-Based Methods | Sparse correspondence | Rad and Lepetit (2017) | 2017 | RGB, CAD Model | RGB, CAD Model | 6DoF | rigid | estimation | source | four-stage, S+C+P+R | symmetrical objects | - | 62.7 | - |
| | | Tekin et al. (2018) | 2018 | RGB, CAD Model | RGB, CAD Model | 6DoF | rigid | estimation | source | two-stage, C+P | general | - | 56.0 | - |
| | | Hu et al. (2019) | 2019 | RGB, CAD Model | RGB, CAD Model | 6DoF | rigid | estimation | source | two-stage, C+P | occlusion | 27.0 | - | - |
| | | Song et al. (2020) | 2020 | RGB, CAD Model | RGB, CAD Model | 6DoF | rigid | estimation | source | three-stage, C+P+R | symmetrical objects, occlusion | 47.5 | 91.3 | - |
| | | Hu et al. (2021) | 2021 | RGB, CAD Model | RGB, CAD Model | 6DoF | rigid | estimation | source | two-stage, C+P | large scale variations | 48.6 | - | - |
| | | Chang et al. (2021) | 2021 | RGB, CAD Model | RGB, CAD Model | 6DoF | rigid | estimation | source | two-stage, C+P | transparent, symmetrical objects | - | - | - |
| | | Guo et al. (2023) | 2023 | RGB, CAD Model | RGB, CAD Model | 6DoF | rigid | estimation | source | three-stage, D+C+P | general | 44.5 | - | - |
| | Dense correspondence | Li et al. (2019) | 2019 | RGB, CAD Model | RGB, CAD Model | 6DoF | rigid | estimation | source | three-stage, D+C+P | general | - | 89.9 | - |
| | | Hodan et al. (2020) | 2020 | RGB, CAD Model | RGB, CAD Model | 6DoF | rigid | estimation | source | two-stage, C+P | symmetrical objects | - | - | - |
| | | Shugurov et al. (2021) | 2021 | RGB/RGBD, CAD Model | RGB/RGBD, CAD Model | 6DoF | rigid | estimation | source | four-stage, D+C+P+R | general | - | 99.9 | - |
| | | Chen et al. (2022a) | 2022 | RGB, CAD Model | RGB, CAD Model | 6DoF | rigid | estimation | source | three-stage, D+C+P | general | - | 95.8 | - |
| | | Haugaard and Buch (2022) | 2022 | RGB/RGBD, CAD Model | RGB/RGBD, CAD Model | 6DoF | rigid | estimation | generalization | four-stage, D+C+P+R | general | - | - | - |
| | | Li et al. (2023a) | 2023 | RGB, CAD Model | RGB, CAD Model | 6DoF | rigid | estimation | source | three-stage, D+C+P | general | 51.4 | 97.8 | - |
| | | Xu et al. (2024b) | 2024 | RGB, CAD Model | RGB, CAD Model | 6DoF | rigid | refinement | source | two-stage, P+R | occlusion | 60.7 | 97.4 | 85.7 |
| Template-Based Methods | RGB-based | Sundermeyer et al. (2018) | 2018 | RGB, CAD Model | RGB, CAD Model | 6DoF | rigid | estimation | generalization | three-stage, D+T+P | general | - | 31.4 | - |
| | | Papaioannidis et al. (2020) | 2020 | RGB, CAD Model | RGB, CAD Model | 3DoF | rigid | estimation | source | two-stage, D+T | general | - | - | - |
| | | Li and Ji (2020) | 2020 | RGB, CAD Model | RGB, CAD Model | 6DoF | rigid | estimation | source | three-stage, D+T+R | general | - | 88.6 | - |
| | | Deng et al. (2021) | 2021 | RGB, CAD Model | RGB, CAD Model | 6DoF | rigid | tracking | generalization | two-stage, D+T | symmetrical objects | - | - | - |
| | Point cloud | Li et al. (2022) | 2022 | RGBD, CAD Model | RGBD, CAD Model | 6DoF | rigid | estimation | source | three-stage, S+P+R | general | 70.6 | 99.5 | 96.6 |
| | | Jiang et al. (2024) | 2023 | Depth, CAD Model | Depth, CAD Model | 6DoF | rigid | estimation | source | two-stage, S+P | general | - | - | - |
| | | Dang et al. (2024) | 2024 | Depth, CAD Model | Depth, CAD Model | 6DoF | rigid | estimation | source | two-stage, S+P | general | 52.0 | 69.0 | - |

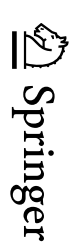

**Table 1** continued

| Methods | | | Published Year | Training Input | Inference Input | Pose DoF | Object Property | Task | Domain Training Paradigm | Inference Mode | Application Area | LM-O | LM ADD(S)-0.1d | YCB-V ADD-S (<0.1m) |
|---|---|---|---|---|---|---|---|---|---|---|---|---|---|---|
| Voting-Based Methods | Indirect voting | Peng et al. (2019) | 2019 | RGB | RGB | 6DoF | rigid | estimation | source | two-stage, V+P | occlusion | 40.8 | 86.3 | - |
| | | He et al. (2020) | 2020 | RGBD, CAD Model | RGBD, CAD Model | 6DoF | rigid | estimation | source | three-stage, S+V+P | general | - | 99.4 | 95.5 |
| | | He et al. (2021) | 2021 | RGBD, CAD Model | RGBD, CAD Model | 6DoF | rigid | estimation | source | three-stage, S+V+P | general | 66.2 | 99.7 | 96.6 |
| | | Cao et al. (2022) | 2022 | RGB, CAD Model | RGB | 6DoF | rigid | estimation | source | end to end | general | 58.7 | - | 90.9 |
| | | Wu et al. (2022) | 2022 | RGBD, CAD Model | RGBD, CAD Model | 6DoF | rigid | estimation | source | three-stage, S+V+P | general | 70.2 | 99.4 | 96.6 |
| | | Zhou et al. (2023a) | 2023 | RGBD, CAD Model | RGBD, CAD Model | 6DoF | rigid | estimation | source | three-stage, S+V+P | general | 77.7 | 99.8 | 96.7 |
| | Direct voting | Wang et al. (2019a) | 2019 | RGBD | RGBD | 6DoF | rigid | estimation | source | two-stage, P+R | general | - | 94.3 | 93.1 |
| | | Tian et al. (2020b) | 2020 | RGBD, CAD Model | RGBD | 6DoF | rigid | estimation | source | three-stage, S+V+P | general | - | 92.9 | 91.8 |
| | | Zhou et al. (2021b) | 2021 | RGBD, CAD Model | RGBD | 6DoF | rigid | estimation | source | two-stage, S+P | general | 65.0 | 99.6 | 95.8 |
| | | Mo et al. (2022) | 2022 | RGBD | RGBD | 6DoF | rigid | estimation | source | two-stage, S+P | general | - | - | 93.6 |
| | | Hong et al. (2024) | 2024 | RGBD | RGBD | 6DoF | rigid | estimation | source | two-stage, S+P | general | 71.1 | 96.7 | 92.7 |
| Regression-Based Methods | Geometry-guided | Chen et al. (2020b) | 2020 | RGBD | RGBD | 6DoF | rigid | estimation | source | end to end | general | - | 98.7 | 92.4 |
| | | Hu et al. (2020) | 2020 | RGBD, CAD Model | RGB | 6DoF | rigid | estimation | source | end to end | general | 43.3 | - | - |
| | | Labbé et al. (2020) | 2020 | RGB, CAD Model | RGB, CAD Model | 6DoF | rigid | estimation | source | three-stage, D+P+R | general | - | - | 93.4 |
| | | Wang et al. (2021c) | 2021 | RGBD, CAD Model | RGB | 6DoF | rigid | estimation | source | two-stage, D+P | general | 62.2 | - | 91.6 |
| | | Di et al. (2021) | 2021 | RGBD, CAD Model | RGB | 6DoF | rigid | estimation | source | two-stage, D+P | occlusion | 62.32 | 96.0 | 90.9 |
| | | Wang et al. (2021b) | 2021 | RGBD, CAD Model | RGB | 6DoF | rigid | estimation | adaptation | two-stage, D+P | occlusion | 59.8 | 85.6 | 90.5 |
| | Direct regression | Xiang et al. (2017) | 2017 | RGB | RGB | 6DoF | rigid | estimation | source | three-stage, S+V+P | cluttered | 24.9 | - | 75.9 |
| | | Li et al. (2018a) | 2018 | RGBD | RGBD | 6DoF | rigid | estimation | source | two-stage, P+R | general | - | - | 94.3 |

**Table 1** continued

| Methods | Published Year | Training Input | Inference Input | Pose DoF | Object Property | Task | Domain Training Paradigm | Inference Mode | Application Area | LM-O \| LM ADD(S)-0.1d | YCB-V ADD-S (<0.1m) |
|---|---|---|---|---|---|---|---|---|---|---|---|
| Li et al. (2018c) | 2018 | RGB, CAD Model | RGB, CAD Model | 6DoF | rigid | refinement, tracking | source | end to end | general | 55.5 88.6 | 81.9 |
| Manhardt et al. (2018) | 2018 | RGB, CAD Model | RGB, CAD Model | 6DoF | rigid | refinement, tracking | generaliztion | end to end | general | - - | - |
| Manhardt et al. (2019) | 2019 | RGB | RGB | 6DoF | rigid | estimation | source | end to end | symmetrical objects | - - | - |
| Papaioannidis and Pitas (2019) | 2019 | RGB | RGB | 3DoF | rigid | estimation | source | two-stage, D+P | general | - - | - |
| Liu et al. (2019b) | 2019 | RGB | RGB | 3DoF | rigid | estimation | source | two-stage, D+P | texture-less | - - | - |
| Wen et al. (2020) | 2020 | RGBD, CAD Model | RGBD | 6DoF | rigid | tracking | generaliztion | end to end | general | - - | 93.9 |
| Wang et al. (2020b) | 2020 | RGBD, CAD Model | RGBD | 6DoF | rigid | estimation | generaliztion | end to end | general | 32.1 58.9 | - |
| Jiang et al. (2022) | 2022 | RGBD | RGBD | 6DoF | rigid | estimation | source | end to end | general | 30.8 97.0 | 95.2 |
| Hai et al. (2023b) | 2023 | RGB, CAD Model | RGB | 6DoF | rigid | refinement | source | end to end | general | 66.4 99.3 | - |
| Li et al. (2024a) | 2024 | RGB, CAD Model | RGB, CAD Model | 6DoF | rigid | refinement | source | two-stage, S+P | general | - - | 97.0 |

bounding box corners. Finally, they used the PnP algorithm (Fischler & Bolles, 1981) to estimate the object pose. Additionally, they employed a classifier to determine the pose range in real-time, addressing the issue of ambiguity in symmetric objects. Tekin et al. (2018) proposed a CNN network inspired by YOLO (Redmon et al., 2016) to integrate object detection and pose estimation, directly predicting the locations of the projected vertices of the 3D object bounding box. Unlike Rad and Lepetit (2017) and Tekin et al. (2018), Pavlakos et al. (2017) predicted the 2D projections of predefined semantic keypoints. Doosti et al. (2020) introduced a compact model comprising two adaptive graph convolutional neural networks (GCNNs) (Kipf & Welling, 2016), collaborating to estimate object and hand poses. To further enhance the robustness of object pose estimation, Song et al. (2020) employed a hybrid intermediate representation to convey geometric details in the input image, encompassing keypoints, edge vectors, and symmetry correspondences. Liu et al. (2021a) proposed a multi-directional feature pyramid network along with a method that calculates object pose estimation confidence by incorporating spatial and plane information. Hu et al. (2021) introduced a single-stage hierarchical end-to-end trainable network to address pose estimation challenges associated with scale variations in aerospace objects. In a recent development, Lian and Ling (2023) increased the number of predefined 3D keypoints to enhance the establishment of correspondences. Moreover, they devised a hierarchical binary encoding approach for localizing keypoints, enabling gradual refinement of correspondences and transforming correspondence regression into a more efficient classification task. To estimate transparent object pose, Chang et al. (2021) used a 3D bounding box prediction network and multi-view geometry. Their method first detects 2D projections of 3D bounding box vertices, and then reconstructs 3D points based on the multi-view detected 2D projections incorporating camera motion data. Additionally, they introduced a generalized pose definition to address pose ambiguity for symmetric objects. To enhance the efficiency of pose estimation networks, Guo et al. (2023) integrated knowledge distillation into object pose estimation by distilling the teacher's distribution of local predictions into the student network. Liu et al. (2023b) argued that differentiable PnP strategies conflict with the averaging nature of the PnP problem, resulting in gradients that may encourage the network to degrade the accuracy of individual correspondences. To mitigate this, they introduced a linear covariance loss, which can be used for both sparse and dense correspondence-based methods.

To mitigate vulnerability caused by large occlusions, Crivellaro et al. (2017) used several control points to represent each object part. Then, they predicted the 2D projections of these control points to calculate the object pose. Some researchers solved the occlusion problem by predicting keypoints using small patches. Oberweger et al. (2018) processed each patch separately to generate heatmaps and then aggregated the results to achieve precise and reliable predictions. Additionally, they offered a straightforward but efficient strategy to resolve ambiguities between patches and heatmaps during training. Hu et al. (2019) unveiled a segmentation-driven pose estimation framework in which every visible object part offers a local pose prediction through 2D keypoint locations. Furthermore, Huang et al. (2021) conceptualized 2D keypoint locations as probabilistic distributions within the loss function and designed a confidence-based network.

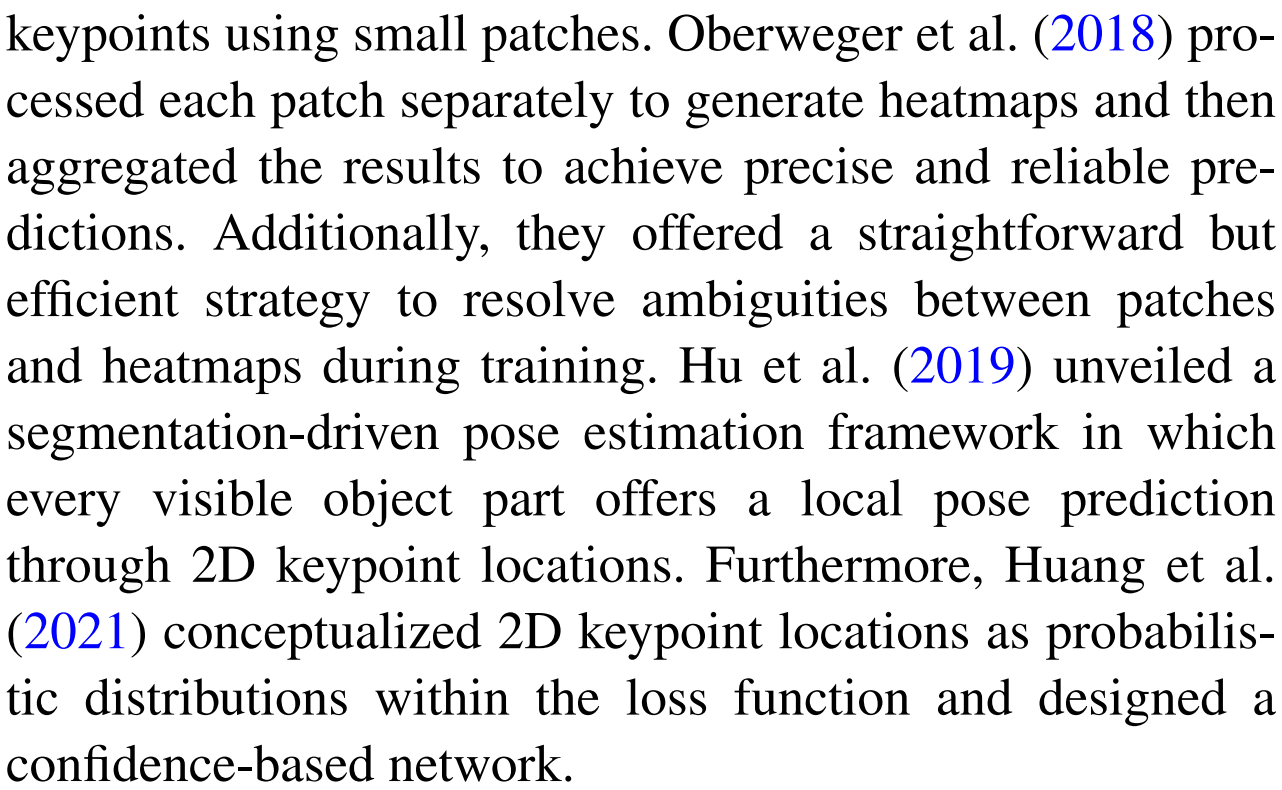

Reducing the reliance on annotated real-world data is also an important task. Some methods exploit geometric consistency as additional information to alleviate the need for annotation. Zhao et al. (2020) employed image pairs with object annotations and relative transformation between viewpoints to automatically identify objects' 3D keypoints that are geometrically and visually consistent. In addition, Yang et al. (2021) used a keypoint consistency regularization for dual-scale images with a labeled 2D bounding box. Using semi-supervised learning, Liu et al. (2021b) developed a unified framework for estimating 3D hand and object poses. They constructed a joint learning framework that conducts explicit contextual reasoning between hand and object representations. To generate pseudo labels in semi-supervised learning, they utilized the spatial-temporal consistency found in large-scale hand-object videos as a constraint. Synthetic data is also a way to solve the annotation problem. Georgakis et al. (2019) reduced the need for expensive 3DoF pose annotations by selecting keypoints and maintaining viewpoint and modality invariance in RGB images and CAD model renderings. Sock et al. (2020) utilized self-supervision to minimize the gap between synthetic and real data and enforced photometric consistency across different object views to fine-tune the model. Further, Zhang et al. (2021) utilized the invariance of geometry relations between keypoints across real and synthetic domains to accomplish domain adaptation. Thalhammer et al. (2021) introduced a specialized feature pyramid network to compute multi-scale features, enabling the simultaneous generation of pose hypotheses across various feature map resolutions.

Overall, sparse correspondence-based methods can estimate object pose efficiently. However, relying on only a few control points can lead to sub-optimal accuracy.

#### 3.1.2 Dense Correspondence Methods

Dense correspondence-based methods aim to establish dense 2D-3D or 3D-3D correspondences. They utilize a significantly larger number of correspondences compared to sparse correspondence-based methods. This enables them to achieve higher accuracy and handle occlusions more effectively. Specifically, for the RGB image, they leverage every pixel or multiple patches to generate pixel-wise correspondences, while for the point cloud, they use the entire point cloud to find point-wise correspondences. Li et al. (2019) argued for the differentiation between rotation and translation, proposing the coordinates-based disentangled pose network. This network separates pose estimation into distinct predictions for rotation and translation. Zakharov et al. (2019) introduced the dense multi-class 2D-3D correspondence-based object pose detector and a tailored deep learning-based refinement process. In addition, Cai and Reid (2020) proposed a technique to automatically identify and match image landmarks consistently across different views, aiming to enhance the process of learning 2D-3D mapping. Wang et al. (2021a) developed a pose estimation pipeline guided by reconstruction, capitalizing on geometric consistency. Further, Shugurov et al. (2021) built upon Zakharov et al. (2019) by developing a unified deep network capable of accommodating multiple image modalities (such as RGB and Depth) and integrating a differentiable rendering-based pose refinement method. Su et al. (2022) introduced a discrete descriptor realized by hierarchical binary grouping, capable of densely representing the object surface. As a result, this method can predict fine-grained correspondences. Chen et al. (2022a) introduced a probabilistic PnP (Fischler & Bolles, 1981) layer designed for general end-to-end pose estimation. This layer generates a pose distribution on the SE(3) manifold. On the other hand, Xu et al. (2022) argued that encoding pose-sensitive local features and modeling the statistical distribution of inlier poses are crucial for accurate and robust 6DoF pose estimation. Inspired by PPF (Drost et al., 2010), they exploited pose-sensitive information carried by each pair of oriented points and an ensemble of redundant pose predictions to achieve robust performance on severe inter-object occlusion and systematic noises in scene point clouds.

Some methods recover object poses by establishing 3D-3D correspondences. Huang et al. (2022) used an RGB image to predict 3D object coordinates in the camera frustum, thus establishing 3D-3D correspondences. Further, Jiang et al. (2023) introduced a center-based decoupled framework, leveraging bird's eye and front views for object center voting. They utilized feature similarity between the center-aligned object and the object CAD model to establish correspondences for Singular Value Decomposition (SVD)-based (Besl & McKay, 1992) rotation estimation. More recently, Lin et al. (2024d) utilized an RGBD image as input and employed point-to-surface matching to estimate the object surface correspondence. They establish 3D-3D correspondences by iteratively constricting the surface, transitioning it into a correspondence point while progressively eliminating outliers.

Some methods put more effort into handling challenging cases, such as symmetric objects (Park et al., 2019b; Hodan et al., 2020; Wu et al., 2023) and texture-less objects (Wu et al., 2021a). Park et al. (2019b) utilized generative adversarial training to reconstruct occluded parts to alleviate the impact of occlusion. They handle symmetric objects by guiding predictions towards the nearest symmetric pose. In addition, Hodan et al. (2020) modeled an object using compact surface fragments to handle symmetries in object modeling effectively. For each pixel, the network predicts: the likelihood of each object's presence, the probability of the fragments conditional on the object's presence, and the exact 3D translation of each fragment. Finally, the object pose is determined using a robust and efficient version of the PnP-RANSAC algorithm (Fischler & Bolles, 1981). Further, Wu et al. (2023) employed a geometric-aware dense matching network to acquire visible dense correspondences. Additionally, they utilized the distance consistency of these correspondences to mitigate ambiguity in symmetrical objects. For texture-less objects, Wu et al. (2021a) leveraged information from the object CAD model and established 2D-3D correspondences using a pseudo-Siamese neural network.

With research development, domain adaptation, weak supervision, and self-supervision techniques have been introduced into pose estimation. Li et al. (2021b) noticed that images with varying levels of realism and semantics exhibit different transferability between synthetic and real domains. Consequently, they decomposed the input image into multi-level semantic representations and merged the strengths of these representations to mitigate the domain gap. Further, Hu et al. (2022) introduced a method exclusively trained on synthetic images, which infers the necessary pose correction for refining rough poses. Haugaard and Buch (2022) utilized learned distributions to sample, score, and refine pose hypotheses. Correspondence distributions are learned using a contrastive loss. This method is unsupervised regarding visual ambiguities. More recently, Li et al. (2023a) introduced a weakly-supervised reconstruction-based pipeline. Initially, they reconstructed the objects from various viewpoints using an implicit neural representation. Subsequently, they trained a network to predict pixel-wise 2D-3D correspondences. Hai et al. (2023a) proposed a refinement strategy that uses the geometry constraint in synthetic-to-real image pairs captured from multiple viewpoints.

There are also methods that focus on pose refinement. Lipson et al. (2022) iteratively refined pose and correspondences in a tightly coupled manner. They incorporated a differentiable layer to refine the pose by solving the bidirectional depth-augmented PnP problem. In addition, Xu et al. (2024b) formulated object pose refinement as a non-linear least squares problem using the estimated correspondence field, *i.e.*, the correspondence between the RGB image and the rendered image using the initial pose. The non-linear least squares problem is then solved by a differentiable

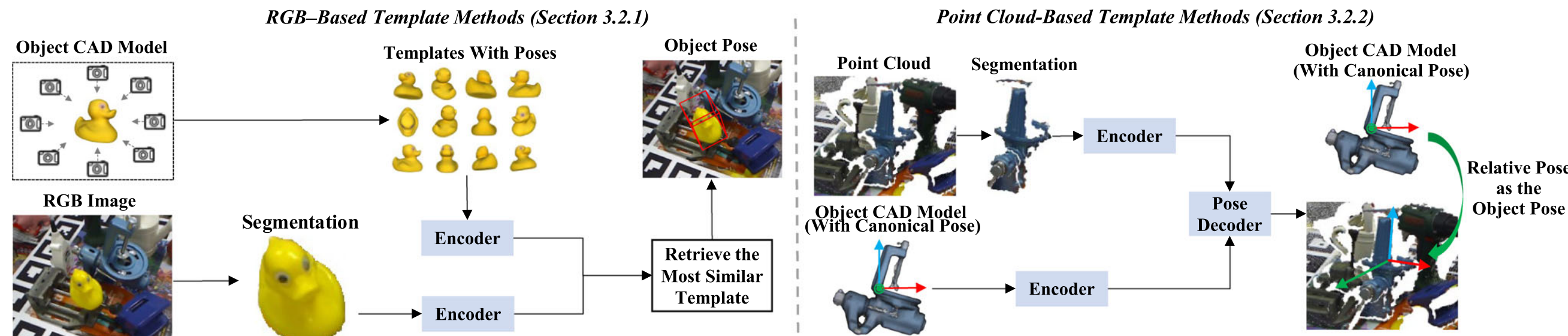

**Fig. 5** Illustration of the template-based methods (Sec. 3.2), which involve identifying the most similar template from a set of templates labeled with ground-truth object poses

levenberg-marquardt algorithm (Moré, 2006), enabling end-to-end training.

**Brief Discussion:** In general, the aforementioned correspondence-based methods exhibit robustness to occlusion since they use local correspondences to predict object pose. However, these methods may encounter challenges when handling objects that lack salient shape features or texture.

### 3.2 Template-Based Methods

By leveraging global information from the image, template-based methods can effectively address the challenges posed by texture-less objects. Template-based methods involve identifying the most similar template from a set of templates labeled with ground-truth object poses. They can be categorized into RGB-based template (Sec. 3.2.1) and point cloud-based template (Sec. 3.2.2) methods. These two methods are illustrated in Fig. 5. The characteristics and performance of some representative methods are shown in Table 1.

#### 3.2.1 RGB-Based Template Methods

When the input is an RGB image, the templates comprise 2D projections extracted from object CAD models, with annotations of ground-truth poses. This process reformulates object pose estimation as an image retrieval task. As a seminal contribution, Sundermeyer et al. (2018) achieved 3D rotation estimation through a variant of denoising autoencoder, which learns an implicit representation of object rotation. If depth is available, it can be used for pose refinement. Liu et al. (2019a) developed a CNN akin to an autoencoder to reconstruct arbitrary scenes featuring the target object and extract the object area. In addition, Zhang et al. (2020b) utilized an object detector and a keypoint extractor to simplify the template search process. Papaioannidis et al. (2020) suggested that estimating object poses in synthetic images is more straightforward. Therefore, they employed a generative adversarial network to convert real images into synthetic ones while preserving the object pose. Li and Ji (2020) utilized a new pose representation (*i.e.*, 3D location field) to guide an auto-encoder to distill pose-related features, thereby enhancing the handling of pose ambiguity. Stevšič and Hilliges (2020) proposed a spatial attention mechanism to identify and utilize spatial details for pose refinement. Different from the above methods, Deng et al. (2021) addressed the 6DoF object pose tracking problem within the Rao-Blackwellized particle filtering (Doucet et al., 2001) framework. They finely discretized the rotation space and trained an autoencoder network to build a codebook of feature embeddings for these discretized rotations. This method efficiently estimates the 3D translation along with the full distribution over the 3D rotation.

RGB cameras are widely used as visual sensors, yet they struggle to capture sufficient information under poor lighting conditions. This results in poor pose estimation performance.

#### 3.2.2 Point Cloud-Based Template Methods

With the popularity of consumer-grade 3D cameras, point cloud-based methods take full advantage of their ability to adapt to poor illumination and capture geometric information. When dealing with a point cloud, the template comprises the object CAD model in canonical pose. Notably, we classify the methods that directly regress the relative pose between the object CAD model and the observed point cloud as template-based methods. This is because these methods can be interpreted as seeking the optimal relative pose that aligns the observed point cloud with the template. Consequently, the determined relative pose serves as the object pose. Li et al. (2022) adopted a feature disentanglement and alignment module to establish part-to-part correspondences between the partial point cloud and object CAD model, enhancing geometric constraint. Jiang et al. (2024) proposed a point cloud registration framework based on the SE(3) diffusion model, which gradually perturbs the optimal rigid transformation of a pair of point clouds by continuously injecting perturbation transformation through the SE(3) forward diffusion process. Then, the SE(3) reverse denoising process is used to gradu-ally denoise, making it closer to the optimal transformation for accurate pose estimation. Dang et al. (2024) proposed two key contributions to enhance pose estimation performance on real-world data. First, they introduced a directly supervised loss function that bypasses the SVD (Besl & McKay, 1992) operation, mitigating the sensitivity of SVD-based loss functions to the rotation range between the input partial point cloud and the object CAD model. Second, they devised a match normalization strategy to address disparities in feature distributions between the partial point cloud and the CAD model.

**Fig. 6** Illustration of the voting-based methods (Sec. 3.3), which determine object pose through a pixel-level or point-level voting scheme

**Brief Discussion:** Overall, template-based methods leverage global information from the image, enabling them to effectively handle texture-less objects. Moreover, they are advantageous in that they can naturally handle the one-to-many mapping from an image to multiple possible poses, which often arises due to pose ambiguity. However, the high pose estimation accuracy comes at the cost of increased memory usage by the templates and a rapid rise in computational complexity. Additionally, they may also exhibit poor performance when confronted with occluded objects.

## 3.3 Voting-Based Methods

Voting-based methods determine object pose through a pixel-level or point-level voting scheme, which can be categorized into two main types: indirect voting (Sec. 3.3.1) and direct voting (Sec. 3.3.2). The illustration of these types of methods is shown in Fig. 6. The attributes and performance of some representative methods are shown in Table 1.

### 3.3.1 Indirect Voting Methods

Indirect voting methods first estimate a set of 2D keypoint hypotheses from an RGB image via pixel-level voting, or a set of 3D keypoint hypotheses from a point cloud via point-level voting, and then estimate the keypoint distributions based on these hypotheses. Subsequently, the object pose is determined through 2D-3D or 3D-3D keypoint correspondences as described in Sec. 3.1. For instance, some researchers predicted 2D keypoints and then derived the object pose through 2D-3D keypoints correspondence. Liu et al. (2021c) introduced a continuous representation method called keypoint distance field (KDF), which extracts 2D keypoints by voting on each KDF. Meanwhile, Cao et al. (2022) proposed a method called dynamic graph PnP (Fischler & Bolles, 1981) to learn the object pose from 2D-3D correspondence, enabling end-to-end training. Moreover, Liu et al. (2023f) introduced a bidirectional depth residual fusion network to fuse RGBD information, thereby estimating 2D keypoints precisely. Inspired by the diffusion model, Xu et al. (2024a) proposed a diffusion-based framework to formulate 2D keypoint detection as a denoising process to establish more accurate 2D-3D correspondences.

Unlike the aforementioned methods that predict 2D keypoints, He et al. (2020) proposed a depth hough voting network to predict 3D keypoints. Subsequently, they estimated the object pose through levenberg-marquardt algorithm (Moré, 2006). Furthermore, He et al. (2021) introduced a bidirectional fusion network to complement RGB and depth heterogeneous data, thereby better predicting the 3D keypoints. To better capture features among object points in 3D space, Mei et al. (2022) utilized graph convolutional networks to facilitate feature exchange among points in 3D space, aiming to improve the accuracy of predicting 3D keypoints. Wu et al. (2022) proposed a 3D keypoint voting scheme based on cross-spherical surfaces, allowing for generating smaller and more dispersed 3D keypoint sets, thus improving estimation efficiency. To obtain more accurate 3D keypoints, Wang et al. (2023a) presented an iterative 3D keypoint voting network to refine the initial localization of 3D keypoints. Most recently, Zhou et al. (2023a) introduced a novel weighted vector 3D keypoints voting algorithm, which adopts a non-iterative global optimization strategy to precisely localize 3D keypoints, while also achieving near real-time inference speed.

In response to challenging scenarios such as cluttered or occlusion, Peng et al. (2019) introduced a pixel-wise voting network to regress pixel-level vectors pointing to 3D keypoints. These vectors create a flexible representation for locating occluded or truncated 3D keypoints. Since most industrial parts are parameterized, Zeng et al. (2021b) defined 3D keypoints linked to parameters through driven parameters and symmetries. This approach effectively addresses the pose estimation of objects in stacking scenes.

Rather than utilizing a single-view RGBD image as input, Duffhauss et al. (2022) employed multi-view RGBD images as input. They extracted visual features from each RGB image, while geometric features were extracted from the object point cloud (generated by fusing all depth images). This multi-view RGBD feature fusion-based method can accurately predict object pose in cluttered scenes.

Some researchers have proposed new training strategies to improve pose estimation performance. Yu et al. (2020) developed a differentiable proxy voting loss that simulates hypothesis selection during the voting process, enabling end-to-end training. In addition, Lin et al. (2022b) proposed a novel learning framework, which utilizes the accurate result of the RGBD-based pose refinement method to supervise the RGB-based pose estimator. To bridge the domain gap between synthetic and real data, Ikeda et al. (2022) introduced a method to transfer object style transfer from synthetic to realistic without manual intervention.

In general, indirect voting-based methods provide an excellent solution for instance-level object pose estimation. However, the accuracy of pose estimation heavily relies on the quality of the keypoints, which can result in lower robustness.

#### 3.3.2 Direct Voting Methods

Since the performance of the indirect voting methods heavily depends on the selection of keypoints, direct voting methods have been proposed as an alternative solution, which directly predict the pose and confidence at the pixel-level or point-level, then select the pose with the highest confidence as the object pose. Tian et al. (2020b) uniformly sampled rotation anchors in SO(3). Subsequently, they predicted constraint deviations for each anchor towards the target, using the uncertainty score to select the best prediction. Then, they detected the 3D translation by aggregating point-to-center vectors towards the object center to recover the 6DoF pose. Wang et al. (2019a) fused RGB and depth features on a per-pixel basis and utilized a pose predictor to generate 6DoF pose and confidence for each pixel. Subsequently, they selected the pose of the pixel with the highest confidence as the final pose. Zhou et al. (2020) employed CNNs (LeCun et al., 1989) to extract RGB features, which are then integrated into the point cloud to obtain fused features. Unlike (Wang et al., 2019a), the fused features take the form of point sets rather than feature mappings.

However, the aforementioned RGBD fusion methods merely concatenate RGB and depth features without delving into their intrinsic relationship. Therefore, Zhou et al. (2021b) proposed a new multi-modal fusion graph convolutional network to enhance the fusion of RGB and depth images, capturing the inter-modality correlations through local information propagation. Liu et al. (2024b) decoupled scale-related and scale-invariant information in the depth image to guide the network in perceiving the scene's 3D structure and provide scene texture for the RGB image feature extraction. Unlike the aforementioned approaches that use still images, Mu et al. (2021) proposed a time fusion model integrating temporal motion information from RGBD images for 6DoF object pose estimation. This method effectively captures object motion and changes, thereby enhancing pose estimation accuracy and stability.

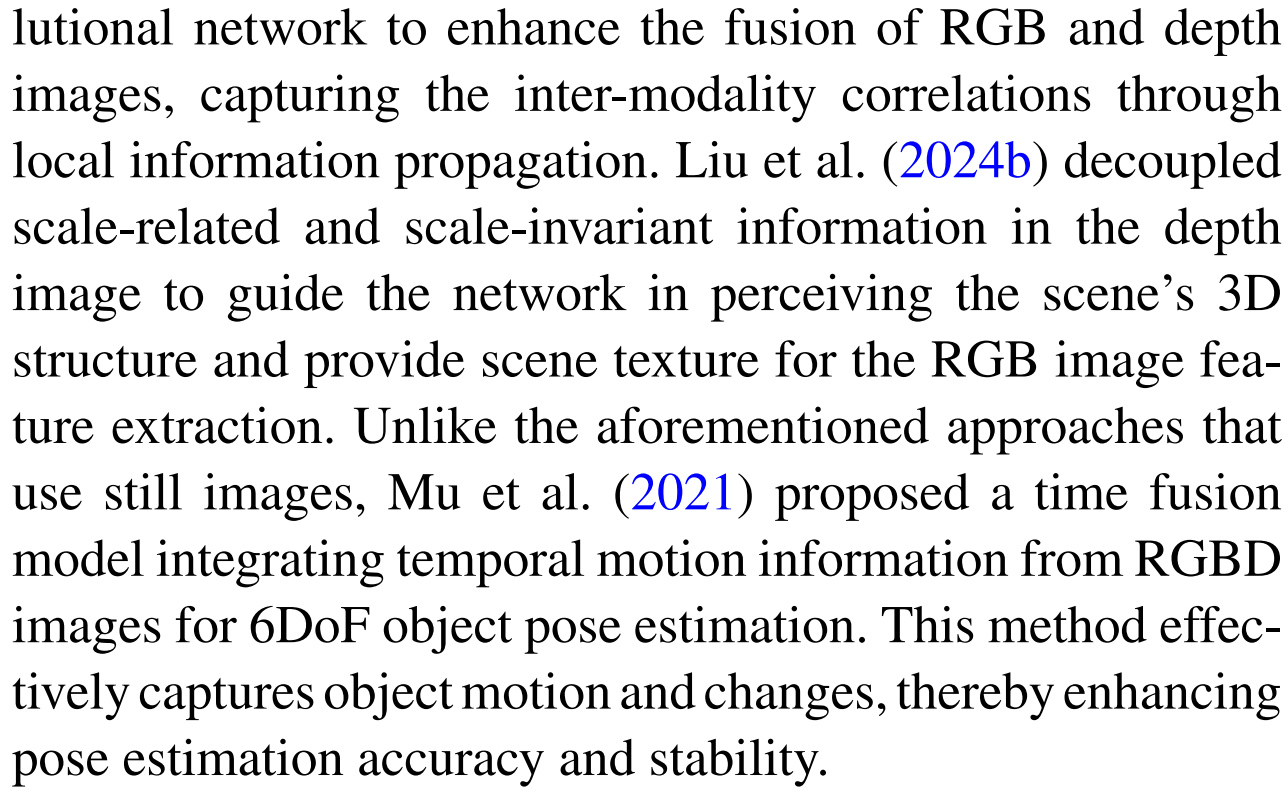

Symmetric objects may have multiple true poses, leading to ambiguity in pose estimation. To address this issue, Mo et al. (2022) designed a symmetric-invariant pose distance metric, which enables the network to estimate symmetric objects accurately. Cai et al. (2022b) introduced a 3D rotation representation to learn the object implicit symmetry, eliminating the need for additional prior knowledge about object symmetry.

To reduce dependency on annotated real data, Zeng et al. (2021a) trained their model solely on synthetic dataset. Then, they utilized a sim-to-real learning network to improve their generalization ability. During pose estimation, they transformed scene points into centroid space and obtained object pose through clustering and voting.

**Brief Discussion:** Overall, voting-based methods have demonstrated superior performance in pose estimation tasks. However, the voting process is time-consuming and increases computational complexity (Bukschat & Vetter, 2020).

### 3.4 Regression-Based Methods

Regression-based methods aim to directly obtain the object pose from the learned features. They can be divided into two main types: geometry-guided regression (Sec. 3.4.1) and direct regression (Sec. 3.4.2). The illustration of these two types of methods is shown in Fig. 7. The attributes and performance of some representative methods are shown in Table 1.

#### 3.4.1 Geometry-Guided Regression Methods

Geometry-guided regression methods leverage geometric information from RGBD images (such as object 3D structural features or 2D-3D geometric constraints) to assist in object pose estimation. Gao et al. (2020) employed decoupled networks for rotation and translation regression from the object point cloud. Meanwhile, Chen et al. (2020b) introduced a rotation residual estimator to estimate the residual between the predicted rotation and the ground truth, enhancing the accuracy of rotation prediction. Lin et al. (2021c) used a network to extract the geometric features of the object point cloud. Then, they enhanced the pairwise consistency of geometric features by applying spectral convolution on pairwise compatibility graphs. Additionally, Shi et al. (2021) learned geometric and contextual features within point cloud blocks. Then, they trained a sub-block network to predict the pose of each point cloud block. Finally, the most reliable block pose is selected as the object pose. To address the challenge of point cloud-based object pose tracking, Liu et al. (2024c) proposed a shifted point convolution operation between the point clouds of adjacent frames to facilitate the local context interaction.

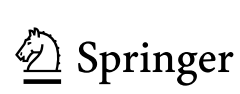

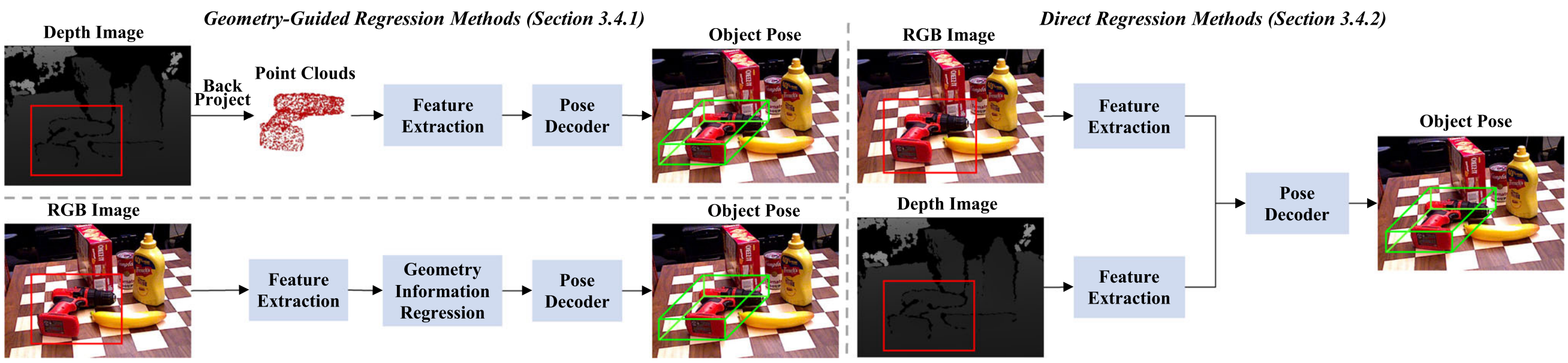

**Fig. 7** Illustration of the regression-based methods (Sec. 3.4), which aim to obtain the object pose directly from the learned features

Approaches solely relying on object point cloud often overlook the object texture details. Therefore, Wen et al. (2021) and An et al. (2024) leveraged the complementary nature of RGB and depth information. They improved cross-modal fusion strategies by employing attention mechanisms to effectively align and integrate these two heterogeneous data sources, resulting in enhanced performance.

In contrast to the aforementioned methods that directly derive geometric information from the depth image or the object CAD model, many researchers focused more on generating geometric constraints from the RGB image. Hu et al. (2020) learned the 2D offset from the CAD model center to the 3D bounding box corners from the RGB image, and then directly regressed the object pose from the 2D-3D correspondence. Di et al. (2021) used a shared encoder and two independent decoders to generate 2D-3D correspondence and self-occlusion information, improving the robustness of object pose estimation under occlusion. Further, Wang et al. (2021c) proposed a Geometry-Guided Direct Regression Network (GDR-Net) to learn object pose from dense 2D-3D correspondence in an end-to-end manner. Wang et al. (2021b) introduced noise-augmented student training and differentiable rendering based on GDR-Net (Wang et al., 2021c), enabling robustness to occlusion scenes through self-supervised learning with multiple geometric constraints. Zhang et al. (2022d) proposed a transformer-based pose estimation approach that consists of a patch-aware feature fusion module and a transformer-based pose refinement module to address the limitation of CNN-based networks in capturing global dependencies. Most recently, Feng et al. (2024) decoupled rotation into two sets of corresponding 3D normals. This decoupling strategy significantly improves the rotation accuracy.

Given the labor-intensive nature of real-world data annotation, some methods leverage synthetic data training to generalize to the real world. Gao et al. (2021) constructed a lightweight synthetic point cloud generation pipeline and leveraged an enhanced point cloud-based autoencoder to learn latent object pose information to regress object pose. To improve generalization to real-world scenes, Zhou et al. (2021a) utilized annotated synthetic data to supervise the network convergence. They proposed a self-supervised pipeline for unannotated real data by minimizing the distance between the CAD model transformed from the predicted pose and the input point cloud. Tan and Dong (2023) proposed a self-supervised monocular object pose estimation network consisting of teacher and student modules. The teacher module is trained on synthetic data for initial object pose estimation, and the student model predicts camera pose from the unannotated real image. The student module acquires knowledge of object pose estimation from the teacher module by imposing geometric constraints derived from the camera pose.

Geometry-guided regression methods typically require additional processing steps to extract and handle geometric information, which increases computational costs and complexity.

#### 3.4.2 Direct Regression Methods

Direct regression methods aim to directly recover the object pose from the RGBD image without additional transformation steps, thus reducing complexity. These methods encompass various strategies, including coupled pose output, decoupled pose output, and 3D rotation (3DoF pose) output. Coupled pose involves predicting object rotation and translation together, while decoupled pose involves predicting them separately. Moreover, the 3DoF pose output focuses solely on predicting object rotation without considering translation. These strategies are discussed in detail below.

**Coupled Pose:** To overcome lighting variations in the environment, Rambach et al. (2018) used a pencil filter to normalize the input image into light-invariant representations, and then directly regressed the object coupled pose using a CNN network. Additionally, Kleeberger and Huber (2020) introduced a robust framework to handle occlusions between objects and estimate the multiple objects pose in the image. This framework is capable of running in real-time at 65 FPS. Sarode et al. (2019) introduced a PointNet-based (Qi et al., 2017) framework to align point clouds for pose estimation, aiming to reduce sensitivity to pose misalignment. Estimating object pose from a single RGB image introduces an inherent ambiguity problem. Manhardt et al. (2019) suggested explicitly addressing these ambiguities. They predicted multiple 6DoF poses for each object to estimate specific pose distributions caused by symmetry and repetitive textures. Inspired by the visible surface difference metric, Bengtson et al. (2021) relied on a differentiable renderer and the CAD model to generate multiple weighted poses, avoiding falling into local minima. Moreover, Park and Cho (2022) proposed a method for pose estimation based on the local grid in object space. The method locates the grid region of interest on a ray in camera space and transforms the grid into object space via the estimated pose. The transformed grid is a new standard for sampling mesh and estimating pose.

For object pose tracking, Garon and Lalonde (2017) proposed a real-time tracking method that learns transformation relationships from consecutive frames during training, and used FCN (Long et al., 2015) to obtain the relative pose between two frames for training and inference.

Coupled pose may lead to information coupling between rotation and translation, making it difficult to distinguish their relationship during the optimization process, thus affecting estimation accuracy.

**Decoupled Pose:** Decoupling the 6DoF object pose enables explicit modeling of the dependencies and independencies between object rotation and translation (Xiang et al., 2017).

In object pose estimation, Xiang et al. (2017) estimated the 3D translation by locating the object center in the image and predicting the distance from the object center to the camera. They further estimated the 3D rotation by regressing to a quaternion representation, and introduced a novel loss function to handle symmetric objects better. Meanwhile, Kehl et al. (2017) extended the SSD framework (Liu et al., 2016) to generate 2D bounding boxes, as well as confidence scores for each viewpoint and in-plane rotation. Then, they chose the 2D bounding box through non-maximum suppression, along with the highest confidence viewpoint and in-plane rotation to infer the 3D translation, resulting in the full 6DoF object pose. Wu et al. (2018), Do et al. (2018) and Bukschat and Vetter (2020) used two parallel FCN (Long et al., 2015) branches to regress the object rotation and translation independently. To eliminate the dependence on annotations of real data, Wang et al. (2020b) used synthetic RGB data for fully supervised training, and then leveraged neural rendering for self-supervised learning on unannotated real RGBD data. Moreover, Jiang et al. (2022) fused RGBD, built-in 2D-pixel coordinate encoding, and depth normal vector features to better estimate the object rotation and translation. Single-view methods suffer from ambiguity, therefore, Li et al. (2018a) proposed a multi-view fusion framework to reduce the ambiguity inherent in single-view frameworks. Further, Labbé et al. (2020) proposed a unified approach for multi-view, multi-object pose estimation. Initially, they utilized a single-view, single-object pose estimation technique to derive pose hypotheses for individual objects. Then, they aligned these object pose hypotheses across multiple input images to collectively infer both the camera viewpoints and object pose within a unified scene. Most recently, Hsiao et al. (2024) introduced a score-based diffusion method to solve the pose ambiguity problem in RGB-based object pose estimation.

For object pose tracking, Wen et al. (2020) proposed a data-driven optimization strategy to stabilize the 6DoF object pose tracking. Specifically, they predicted the 6DoF pose by predicting the relative pose between the adjacent frames. Liu et al. (2022b) proposed a new subtraction feature fusion module based on (Wen et al., 2020) to establish sufficient spatiotemporal information interaction between adjacent frames, improving the robustness of object pose tracking in complex scenes. Different from the method based on RGBD input, Ge and Loianno (2021) designed a novel deep neural network architecture that integrates visual and inertial features to predict the relative object pose between consecutive image frames.

In object pose refinement, Li et al. (2018c) iteratively refined pose through aligning RGB image with rendered image of object CAD model. Additionally, they predicted optical flow and foreground masks to stabilize the training procedure. Manhardt et al. (2018) refined the 6DoF pose by aligning object contour between the RGB image and rendered contour. The rendered contour is obtained from the object CAD model using the initial pose. Hai et al. (2023b) proposed a shape-constraint recursive matching framework to refine the initial pose. They first computed a pose-induced flow based on the initial and currently estimated pose, and then directly decoupled the 6DoF pose from the pose-induced flow. To address the low running efficiency of the pose refinement methods, Iwase et al. (2021) introduced a deep texture rendering-based pose refinement method for fast feature extraction using an object CAD model with a learnable texture. Most recently, Li et al. (2024a) proposed a two-stage method. The first stage performs pose classification and renders the object CAD model in the classified poses. The second stage performs regression to predict fine-grained residual in the classified poses. This method improves robustness by guiding residual pose regression through pose classification.
**3DoF Pose:** Some researchers pursue more efficient and practical pose estimation by solely regressing the 3D rotation. Papaioannidis and Pitas (2019) proposed a novel quaternion-based multi-objective loss function, which integrates manifold learning and regression for learning 3DoF pose descriptors. They obtained the 3DoF pose through the regression of the learned descriptors. Liu et al. (2019b) trained a triple network based on convolutional neural networks to extract discriminative features from binary images. They incorporated pose-guided methods and regression constraints into the constructed triple network to adapt the features for the regression task, enhancing robustness. In addition, Josifovski et al. (2018) estimated the camera viewpoint related to the object coordinate system by constructing a viewpoint estimation model, thereby obtaining the 3DoF pose appearing in the bounding box.
**Brief Discussion:** Overall, direct regression methods simplify the object pose estimation process and further enhance the performance of instance-level methods.

## 3.5 Discussion

The methods reviewed above demonstrate diverse strategies for instance-level object pose estimation. To place them in a broader context, we discuss approaches from two complementary perspectives: the input modality (RGB vs. RGBD vs. RGB + Point Cloud) that supplies information and the methodological paradigm (Correspondence vs. Template vs. Voting vs. Regression) that defines how it is exploited.

### 3.5.1 Input Modalities

RGB-based methods rely on semantic richness and the wide availability of cameras, which makes them efficient and practical for large-scale deployment. Their limitation is sensitivity to illumination changes and surface texture. RGBD pipelines address this by explicitly fusing photometric and geometric information. When the fusion is designed carefully, for example by exploiting inter-modality relations, separating depth-dependent scale signals, or integrating temporal RGBD motion, they consistently achieve higher accuracy and stability. Point cloud–centric methods directly leverage dense geometry. They are robust in low-light conditions and can capture fine structural details, yet they face challenges with partial views, missing regions, and sensor noise. In summary, RGB inputs remain attractive when efficiency or hardware simplicity is the priority, whereas RGBD and point cloud inputs provide significant advantages due to their active and reliable sensing of the geometric data.

### 3.5.2 Methodological Paradigms

Correspondence-based methods, whether sparse or dense, are effective under occlusion because they rely on local matches and geometric consistency. However, they are less reliable on texture-less or weakly distinctive objects, although recent work shows improvements through stronger feature descriptors and symmetry-aware modeling. Template-based methods exploit global image or shape evidence, which allows them to handle texture-less objects and pose ambiguities, but they remain sensitive to occlusion, background clutter, and illumination changes, and become computationally heavy when matching against a large number of templates. Voting-based methods achieve high accuracy by aggregating pixel or point hypotheses. Additionally, indirect voting depends strongly on keypoint quality, and both indirect and direct variants require considerable computation. Regression-based methods provide a simplified pipeline by predicting pose end-to-end. Geometry-guided regression, in particular, benefits from explicit 2D–3D constraints or point cloud structures to improve stability. Overall, correspondence is suitable for scenarios with partial visibility, template methods are effective on texture-less objects, voting methods are appropriate when accuracy is prioritized over speed, and regression is preferred when efficiency and throughput are critical.

In general, instance-level methods can only estimate specific object instances in the training data, limiting their generalization to unseen objects. Additionally, most instance-level methods require accurate object CAD models, which is a challenge, especially for objects with complex shapes and textures.

# 4 Category-Level Object Pose Estimation

Research on category-level methods has garnered significant attention due to their potential for generalizing to unseen objects within established categories (Sahin & Kim, 2018). In this section, we review category-level methods by dividing them into shape prior-based (Sec. 4.1) and shape prior-free (Sec. 4.2) methods. The illustration of these two categories is shown in Fig. 8 and Fig. 9, respectively. The characteristics and performance of some representative SOTA methods are shown in Table 2.

## 4.1 Shape Prior-Based Methods

Shape prior-based methods first learn a neural network using CAD models of intra-class seen objects in offline mode to derive shape priors, and then utilize them as 3D geometry prior information to guide intra-class unseen object pose estimation. In this part, we divide the shape prior-based methods

**Fig. 8** Illustration of the shape prior-based category-level methods (Sec. 4.1). The dashed arrows indicate offline training, which means that we need to train a model offline using the category-level model library to obtain shape priors. Taking RGBD input as an example, NOCS shape alignment methods (Sec. 4.1.1) first learn a model to predict the NOCS shape/map of the object, and then align the object point cloud with the NOCS shape/map through a non-differentiable pose solution method such as the Umeyama algorithm (Umeyama, 1991) to solve the object pose. In contrast, direct regress pose methods (Sec. 4.1.2) directly regress the object pose from the extracted input features

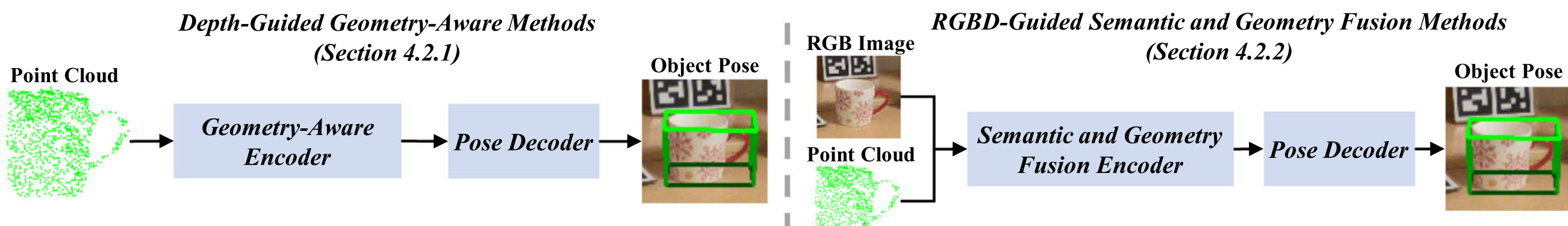

**Fig. 9** Illustration of the shape prior-free category-level methods (Sec. 4.2), which do not have the process of shape priors regression: Depth-guided geometry-aware methods (Sec. 4.2.1) focus on perceiving the global and local geometric information of the object and leverage these 3D geometric features to estimate the object pose. Conversely, RGBD-guided semantic and geometry fusion methods (Sec. 4.2.2) regress the object pose by fusing the 2D semantic and 3D geometric information of the object

into two categories based on their approach to addressing object pose estimation. The first category is Normalized Object Coordinate Space (NOCS) shape alignment methods (Sec. 4.1.1), and the other is the direct pose regression methods (Sec. 4.1.2). The illustration of the two categories is shown in Fig. 8.

#### 4.1.1 NOCS Shape Alignment Methods

The NOCS shape alignment methods first predict the NOCS shape/map, and then use an offline pose solution method (such as the Umeyama algorithm (Umeyama, 1991)) to align the object point cloud with the predicted NOCS shape/map to obtain the object pose. As a pioneering work, Tian et al. (2020a) first extracted the shape prior in offline mode, which is used to represent the mean shape of a category of objects. For example, mugs are composed of a cylindrical cup body and an arc-shaped handle. Next, they introduced a shape prior deformation network for the intra-class unseen object to reconstruct its NOCS shape. Finally, the Umeyama algorithm (Umeyama, 1991) is employed to solve the object pose by aligning the NOCS shape and the object point cloud.

Following Tian et al. (2020a), some methods aim to reconstruct the NOCS shape more accurately. Specifically, Wang et al. (2021d) designed a recurrent reconstruction network to iteratively refine the reconstructed NOCS shape. Further, Chen and Dou (2021) adjusted the shape prior dynamically by using the structure similarity between the RGBD image and the shape prior. Zou et al. (2022) proposed two multi-scale transformer-based networks (Pixelformer and Pointformer) for extracting RGB and point cloud features, and subsequently merging them for shape prior deformation. Different from the previous methods, Fan et al. (2024b) introduced an adversarial canonical representation reconstruction framework, which includes a reconstructor and a discriminator of NOCS representation. Specifically, the reconstructor mainly consists of a pose-irrelevant module and a relational reconstruction module to reduce the sensitivity to rotation and translation, as well as to generate high-quality features, respectively. Then, the discriminator is used to guide the reconstructor to generate realistic NOCS representations. Nie et al. (2023) improved the accuracy of pose estimation via geometry-informed instance-specific priors and multi-stage shape reconstruction. More recently,

**Table 2** Representative category-level methods. For each method, we report its 10 properties, which have the same meanings as described in Table. 1. D, S, N, K, and P denote object detection, instance segmentation, NOCS shape/map regression, keypoints detection, and pose solution/regression, respectively. Moreover, we report the 5°5cm metric of CAMERA25 and REAL275 datasets (Sec. 2)

| Methods | | | Published Year | Training Input | Inference Input | Pose DoF | Object Property | Task | Domain Training Paradigm | Inference Mode | Application Area | CAMERA25 5°5cm (mAP) | REAL275 5°5cm (mAP) |
|---|---|---|---|---|---|---|---|---|---|---|---|---|---|
| Shape Prior-Based Methods | NOCS shape alignment | Tian et al. (2020a) | 2020 | RGBD, CAD Model | RGBD | 9DoF | rigid | estimation | source | three-stage, S+N+P | general | 59.0 | 21.4 |
| | | Wang et al. (2021d) | 2021 | RGBD, CAD Model | RGBD | 9DoF | rigid | estimation | source | three-stage, S+N+P | general | 76.4 | 34.3 |
| | | Chen and Dou (2021) | 2021 | RGBD, CAD Model | RGBD | 9DoF | rigid | estimation | source | three-stage, S+N+P | general | 74.5 | 39.6 |
| | | Zou et al. (2022) | 2022 | RGBD, CAD Model | RGBD | 9DoF | rigid | estimation | source | three-stage, S+N+P | general | 76.7 | 41.9 |
| | | Fan et al. (2022a) | 2022 | RGB, CAD Model | RGB | 9DoF | rigid | estimation | source | three-stage, S+N+P | general | - | - |
| | | Wei et al. (2023) | 2023 | RGB, CAD Model | RGB | 9DoF | rigid | estimation | source | three-stage, S+N+P | general | - | - |
| | Direct regress pose | Irshad et al. (2022) | 2022 | RGBD, CAD Model | RGBD | 9DoF | rigid | estimation | source | end-to-end | general | 66.2 | 29.1 |
| | | Lin et al. (2022a) | 2022 | Depth | Depth | 9DoF | rigid | estimation | generalization | two-stage, S+P | general | 70.9 | 42.3 |
| | | Zhang et al. (2022c) | 2022 | Depth, CAD Model | Depth | 9DoF | rigid | estimation | source | two-stage, S+P | general | 75.5 | 44.6 |
| | | Zhang et al. (2022b) | 2022 | Depth, CAD Model | Depth | 9DoF | rigid | estimation | source | two-stage, S+P | general | 79.6 | 48.1 |
| | | Liu et al. (2022d) | 2022 | Depth | Depth | 9DoF | rigid | refinement, tracking | source | two-stage, S+P | general | 80.3 | 54.4 |
| | | Lin et al. (2022c) | 2022 | RGBD, CAD Model | RGBD | 9DoF | rigid | estimation | generalization | two-stage, S+P | general | - | 45.0 |
| | | Ze and Wang (2022) | 2022 | RGBD, CAD Model | RGBD | 9DoF | rigid | estimation | adaptation | two-stage, S+P | general | - | 33.9 |
| | | Liu et al. (2025a) | 2024 | RGBD, CAD Model | RGBD | 9DoF | rigid | estimation | generalization | two-stage, S+P | general | - | 50.1 |
| Shape Prior-Free Methods | Depth-guided geometry-aware | Li et al. (2020) | 2020 | Depth | Depth | 9DoF | articulated | estimation | generalization | two-stage, S+P | general | - | - |
| | | Chen et al. (2021) | 2021 | Depth | Depth | 9DoF | rigid | estimation | source | two-stage, D+P | general | - | 28.2 |
| | | Weng et al. (2021) | 2021 | Depth | Depth | 9DoF | rigid, articulated | tracking | source | two-stage, N+P | general | - | 62.2 |
| | | Di et al. (2022) | 2022 | Depth | Depth | 9DoF | rigid | estimation | source | two-stage, S+P | general | 79.1 | 42.9 |
| | | You et al. (2022) | 2022 | Depth | Depth | 9DoF | rigid | estimation | generalization | two-stage, S+P | general | - | 16.9 |
| | | Zheng et al. (2023) | 2023 | Depth | Depth | 9DoF | rigid | estimation | source | two-stage, S+P | general | 80.5 | 55.2 |
| | | Zhang et al. (2023) | 2023 | Depth | Depth | 6DoF | rigid | estimation, tracking | source | two-stage, S+P | general | - | 60.9 |

**Table 2** continued

| Methods | | Published Year | Training Input | Inference Input | Pose DoF | Object Property | Task | Domain Training Paradigm | Inference Mode | Application Area | CAMERA25 5°5cm (mAP) | REAL275 5°5cm (mAP) |
|---|---|---|---|---|---|---|---|---|---|---|---|---|
| RGBD-guided semantic & geometry fusion | Wang et al. (2019b) | 2019 | RGBD | RGBD | 9DoF | rigid | estimation | source | two-stage, (S+N)+P | general | 40.9 | 10.0 |
| | Wang et al. (2020a) | 2020 | RGBD | RGBD | 6DoF | rigid | tracking | source | two-stage, K+P | general | - | 33.3 |
| | Lin et al. (2021b) | 2021 | RGBD | RGBD | 9DoF | rigid | estimation | source | two-stage, S+P | general | 70.7 | 35.9 |
| | Wen and Bekris (2021) | 2021 | RGBD | RGBD | 9DoF | rigid | tracking | source | three-stage, S+K+P | general | - | 87.4 |
| | Peng et al. (2022) | 2022 | RGBD, CAD Model | RGBD | 9DoF | rigid | estimation | adaptation | two-stage, S+P | general | - | 33.4 |
| | Lee et al. (2022) | 2022 | RGBD, CAD Model | RGBD | 9DoF | rigid | estimation | adaptation | three-stage, S+N+P | general | - | 34.8 |
| | Lee et al. (2023) | 2023 | RGBD, CAD Model | RGBD | 9DoF | rigid | estimation | generalization | three-stage, S+N+P | general | - | 35.9 |
| | Liu et al. (2023c) | 2023 | RGBD | RGBD | 9DoF | rigid | estimation | source | two-stage, S+P | general | 79.9 | 53.4 |
| | Lin et al. (2023a) | 2023 | RGBD or Depth | RGBD or Depth | 9DoF | rigid | estimation | source | two-stage, S+P | general | 81.4 | 57.6 |
| | Chen et al. (2024) | 2024 | RGBD | RGBD | 9DoF | rigid | estimation | source | two-stage, S+P | general | - | 63.6 |
| Others | Lee et al. (2021) | 2021 | RGB, CAD Model | RGB | 9DoF | rigid | estimation | generalization | three-stage, S+N+P | general | - | - |
| | Lin et al. (2024c) | 2024 | RGBD, Text | RGBD, Text | 9DoF | rigid | estimation | source | two-stage, S+P | general | 82.2 | 58.3 |

Zhou et al. (2023b) designed a two-stage pipeline consisting of deformation and registration to improve accuracy. Zou et al. (2023) introduced a graph-guided point transformer consisting of a graph-guided attention encoder and an iterative non-parametric decoder to further extract the point cloud feature. In addition, Li et al. (2023c) leveraged discrepancies in instance-category structures alongside potential geometric-semantic associations to better investigate intra-class shape information. Yu et al. (2024) further divided the NOCS shape reconstruction process into three parts: coarse deformation, fine deformation, and recurrent refinement to enhance the accuracy of NOCS shape reconstruction.

Given that the annotation of ground-truth object pose is time-consuming, He et al. (2022b) explored a self-supervised method via enforcing the geometric consistency between point cloud and category prior mesh, avoiding using the real-world pose annotation. Further, Li et al. (2023b) first extracted semantic primitives via a part segmentation network, and leveraged semantic primitives to compute SIM(3)-invariant shape descriptor to generate the optimized shape. Then, the Umeyama algorithm (Umeyama, 1991) is utilized to recover the object pose. Through this approach, they achieved domain generalization, bridging the gap between synthesis and real-world application.

Depth images may be unavailable in some challenging scenes (*e.g.*, under strong or low light conditions). Therefore, achieving monocular category-level object pose estimation is of great significance across various applications. Fan et al. (2022a) directly predicted object-level depth and NOCS shape from a monocular RGB image by deforming the shape prior, and subsequently leveraged the Umeyama algorithm (Umeyama, 1991) to solve the object pose. Unlike Fan et al. (2022a), Wei et al. (2023) estimated the 2.5D sketch and separated scale recovery using the shape prior. They then reconstructed the NOCS shape, employing the RANSAC (Fischler & Bolles, 1981) algorithm to remove outliers, before utilizing the PnP algorithm for recovering the object pose. For transparent objects, Chen et al. (2023c) proposed a new solution based on stereo vision, which defines a back-view NOCS map to tackle the problem of image content aliasing.

In general, although these NOCS shape alignment methods can recover the object pose, the alignment process is non-differentiable and is not integrated into the learning process. Thus, errors in predicting the NOCS shape/map have a significant impact on the accuracy of pose estimation.

#### 4.1.2 Direct Regress Pose Methods

Due to the non-differentiable nature of the NOCS shape alignment process, several direct regression-based pose methods have been proposed recently to enable end-to-end training. They directly regress the object pose from the feature level, making the pose acquisition process differentiable. Irshad et al. (2022) treated object instances as spatial centers and proposed an end-to-end method that combines object detection, reconstruction, and pose estimation. Wang et al. (2023b) developed a deformable template field to decouple shape and pose deformation, improving the accuracy of shape reconstruction and pose estimation. On the other hand, Zhang et al. (2022c) proposed a symmetry-aware shape prior deformation method, which integrates shape prior into a direct pose estimation network. Further, Zhang et al. (2022b) introduced a geometry-guided residual object bounding box projection framework to address the challenge of insufficient pose-sensitive feature extraction. In order to obtain a more precise object pose, Liu et al. (2022d) designed CATRE, a pose refinement method based on the alignment of the shape prior and the object point cloud to refine the object pose estimated by the above methods. Zheng et al. (2024) extended CATRE (Liu et al., 2022d) to address the geometric variation problem by integrating hybrid scope layers and learnable affine transformations.

Due to the extensive manual effort required for annotating real-world training data, Lin et al. (2022a) explored the shape alignment of each intra-class unseen instance against its corresponding category-level shape prior, implicitly representing its 3D rotation. This approach facilitates domain generalization from synthesis to real-world scenarios. Further, Ze and Wang (2022) proposed a novel framework based on pose and shape differentiable rendering to achieve domain adaptation object pose estimation. In addition, they collected a large Wild6D dataset for category-level object pose estimation in the wild. Following Ze and Wang (2022), Zhang and Fu (2023) introduced 2D-3D and 3D-2D geometry correspondences to enhance the ability of domain adaptation. Different from the previous approaches, Remus and D'Avella (2023) leveraged instance-level methods for domain-generalized category-level object pose estimation via a single RGB image. Lin et al. (2022c) proposed a deep prior deformation-based network and leveraged a parallel learning scheme to achieve domain generalization. More recently, Liu et al. (2025a) designed a multi-hypothesis consistency learning framework. This framework addresses the uncertainty problem and reduces the domain gap between synthetic and real-world datasets by employing multiple feature extraction and fusion techniques.

**Brief Discussion:** Overall, while the shape prior-based methods mentioned above significantly improve pose estimation performance, obtaining the shape priors requires constructing category-level CAD model libraries and subsequently training a network, which is both cumbersome and time-consuming.

### 4.2 Shape Prior-Free Methods

Shape prior-free methods do not rely on using shape priors and thus have better generalization capabilities. These methods can be divided into three main categories: depth-guided geometry-aware (Sec. 4.2.1), RGBD-guided semantic and geometry fusion (Sec. 4.2.2), and other (Sec. 4.2.3) methods. The illustration of the first two categories is shown in Fig. 9.

#### 4.2.1 Depth-Guided Geometry-Aware Methods

Depth-guided geometry-aware methods primarily exploit 3D geometric cues derived from depth or point cloud data to estimate object pose. By leveraging structures such as 3D Graph Convolution (3DGC), point-wise voting, or geometric reconstruction, these approaches learn pose-sensitive features directly from shape and spatial relationships, enabling robust estimation without explicit reliance on shape priors. Chen et al. (2021) leveraged 3DGC and introduced a fast shape-based method, which consists of an RGB-based network to achieve 2D object detection, a shape-based network for 3D segmentation and rotation regression, and a residual-based network for translation and size regression. Inspired by Chen et al. (2021), Liu et al. (2022a) improved the network with structure encoder and reasoning attention. Further, Di et al. (2022) proposed a geometry-guided point-wise voting method that exploits geometric insights to enhance the learning of pose-sensitive features. Specifically, they designed a symmetry-aware point cloud reconstruction network and introduced a point-wise bounding box voting mechanism during training to add additional geometric guidance. Due to the translation and scale invariant properties of the 3DGC, these methods are limited in perceiving object translation and size information. Based on this, Zheng et al. (2023) further designed a hybrid scope feature extraction layer, which can simultaneously perceive global and local geometric structures and encode size and translation information.

Besides the above 3DGC-based methods, Deng et al. (2022) combined a category-level auto-encoder with a particle filter framework to achieve object pose estimation and tracking. Wang et al. (2023c) leveraged learnable sparse queries as implicit prior to perform deformation and matching for pose estimation. In addition, Wan et al. (2023) developed a semantically-aware object coordinate space to address the semantically incoherent problem of NOCS (Wang et al., 2019b). More recently, Zhang et al. (2023) proposed a scored-based diffusion model to address the multi-hypothesis problem in symmetric objects and partial point clouds. They first leveraged the scored-based diffusion model to generate multiple pose candidates, and then utilized an energy-based diffusion model to remove abnormal poses. On the other hand, Lin et al. (2024b) first introduced an instance-adaptive keypoints detection method and then designed a geometric-aware global and local features aggregation network based on the detected keypoints for pose and size estimation. Li et al. (2024b) leveraged category-level method to determine part object poses for assembling multi-part multi-joint 3D shape.

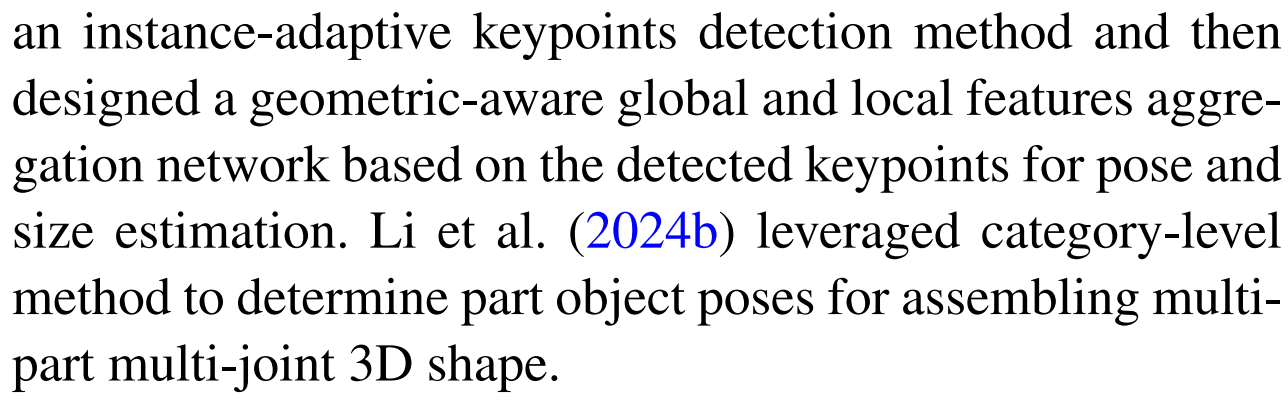

To perform pose estimation on articulated objects, Li et al. (2020) inspired by Wang et al. (2019b), introduced a standard representation for different articulated objects within a category by designing an articulation-aware normalized coordinate space hierarchy, which simultaneously constructs a canonical object space and a set of canonical part spaces. Weng et al. (2021) further proposed CAPTRA, a unified framework that enables 9DoF pose tracking of rigid and articulated objects simultaneously. Due to the nearly unlimited freedom of garments and extreme self-occlusion, Chi and Song (2021) introduce GarmentNets, which conceptualizes deformable object pose estimation as a shape completion problem within a canonical space. More recently, Liu et al. (2023e) developed a reinforcement learning-based pipeline to predict 9DoF articulated object pose via fitting joint states through reinforced agent training. Further, Liu and Zhang (2023) learned part-level SE(3)-equivariant features via a pose-aware equivariant point convolution operator to address the issue of self-supervised articulated object pose estimation.

To avoid using extensive real-world labeled data for training, Li et al. (2021a) used SE(3) equivariant point cloud networks for self-supervised object pose estimation. You et al. (2022) introduced a category-level point pair feature voting method to reduce the impact of synthetic to real-world domain gap, achieving generalizable object pose estimation in the wild.

In general, these methods fully extract the pose-related geometric features. However, the absence of semantic information limits their better performance. Appropriate fusion of semantic and geometric information can significantly improve the robustness of pose estimation.

#### 4.2.2 RGBD-Guided Semantic and Geometry Fusion Methods

RGBD-guided semantic and geometry fusion methods combine the complementary strengths of RGB-based semantic features and depth-based geometric representations within a unified framework. These methods typically predict canonical object coordinates, fuse semantic and structural features, or adopt attention-based integration strategies to enhance robustness. By jointly leveraging semantic context and geometric precision, they achieve higher generalization across diverse intra-class objects and challenging real-world conditions. As a groundbreaking research, Wang et al. (2019b) designed a normalized object coordinate space to provide a canonical representation for a category of objects. They first predicted the class label, mask, and NOCS map of the intra-class unseen object. Then, they utilized the Umeyama algorithm (Umeyama, 1991) to solve object pose by aligning the NOCS map with the object point cloud. To handle various shape changes of intra-class objects, Chen et al. (2020a) learned a canonical shape space as a unified representation. On the other hand, Lin et al. (2021a) explored the applicability of sparse steerable convolution (SSC) to object pose estimation and proposed an SSC-based pipeline. Further, Lin et al. (2021b) proposed a dual pose network, which consists of a shared pose encoder and two parallel explicit and implicit pose decoders. The implicit decoder can enforce predicted pose consistency when there are no CAD models during inference. Wang et al. (2022a) designed an attention-guided network with relation-aware and structure-aware for RGB image and point cloud features fusion. Very recently, Liu et al. (2023c) explored the necessity of shape priors for shape reconstruction of intra-class unseen objects. They demonstrated that the deformation process is more important than the shape prior and proposed a prior-free implicit space transformation network. Lin et al. (2023a) addressed the poor rotation estimation accuracy by decoupling the rotation estimation into viewpoint and in-plane rotation. In addition, they also proposed a spherical feature pyramid network based on spatial spherical convolution to process spherical signals. With the rapid development of the Large Vision Model (LVM), Chen et al. (2024) further leveraged the LVM DINOv2 (Oquab et al., 2023) to extract the SE(3)-consistent semantic features and fused them with object-specific hierarchical geometric features to encapsulate category-level information for rotation estimation.

Since the above methods still require a large amount of real-world annotated training data, their applicability in real-world scenes is limited. To this end, Peng et al. (2022) proposed a real-world self-supervised training framework based on deep implicit shape representation. They leveraged the deep signed distance function (Park et al., 2019a) as a 3D representation to achieve domain adaptation from synthesis to the real world. In addition, Lee et al. (2022) introduced a teacher-student self-supervised learning mechanism. They used supervised training in the source domain and self-supervised training in the target domain, effectively achieving domain adaptation. Recently, Lee et al. (2023) further proposed a test-time adaptation framework for domain-generalized category-level object pose estimation. Specifically, they first trained the model using labeled synthetic data and then leveraged the pre-trained model for test-time adaptation in the real world during inference.

To improve the running speed of the object pose estimation method, once the object pose of the first frame is acquired, continuous spatio-temporal information can be utilized to track the object pose. Wang et al. (2020a) proposed an anchors-based object pose tracking method. They first detected the anchors of each frame as the keypoints, and then solved the relative object pose through the keypoints correspondence. Wen and Bekris (2021) first obtained continuous frame RGBD masks through the video segmentation network, transductive-VOS (Zhang et al., 2020a), and then leveraged LF-Net (Ono et al., 2018) for generalized keypoints detection. Next, they matched keypoints between consecutive frames and performed coarse registration to estimate the initial relative pose. Finally, a memory-augmented pose graph optimization method is proposed for continuous pose tracking.

Overall, these RGBD-guided semantic and geometry fusion methods achieve superior performance. However, if the input depth image contains errors, the accuracy of pose estimation can significantly decrease. Hence, ensuring robustness in pose estimation when dealing with erroneous or missing depth images is crucial.

#### 4.2.3 Others

Since most mobile devices are not equipped with depth cameras Yang et al. (2025a, c); Liu et al. (2025c), Chen et al. (2020c) incorporated a neural synthesis module with a gradient-based fitting procedure to simultaneously predict object shape and pose, achieving monocular object pose estimation. Lee et al. (2021) estimated the NOCS shape and the metric scale shape of the object, and performed a similarity transformation between them to solve the object pose and size. Further, Yen-Chen et al. (2021) inverted neural radiance fields for monocular category-level pose estimation. Different from the previous methods, Lin et al. (2022e) proposed a keypoint-based single-stage pipeline via a single RGB image. Guo et al. (2022) redefined the monocular category-level object pose estimation problem from a long-horizon visual navigation perspective. On the other hand, Ma et al. (2022) enhanced the robustness of the monocular method in occlusion scenes through coarse-to-fine rendering of neural features. Given that transparent instances lack both color and depth information, Zhang et al. (2022a) proposed to utilize depth completion and surface normal estimation to achieve category-level pose estimation for transparent instances. More recently, Wei et al. (2024) decoupled 6D pose from size, leveraged a pretrained monocular estimator for inlier 2D to 3D correspondences and a separate branch for metric scale, and solved pose with RANSAC-PnP, improving rotation accuracy.

In order to improve the running efficiency of the monocular method, Lin et al. (2022d) developed a keypoint-based monocular object pose tracking approach. This approach demonstrates the significance of integrating uncertainty estimation using a tracklet-conditioned deep network and probabilistic filtering. Following Lin et al. (2022d), Yu et al.

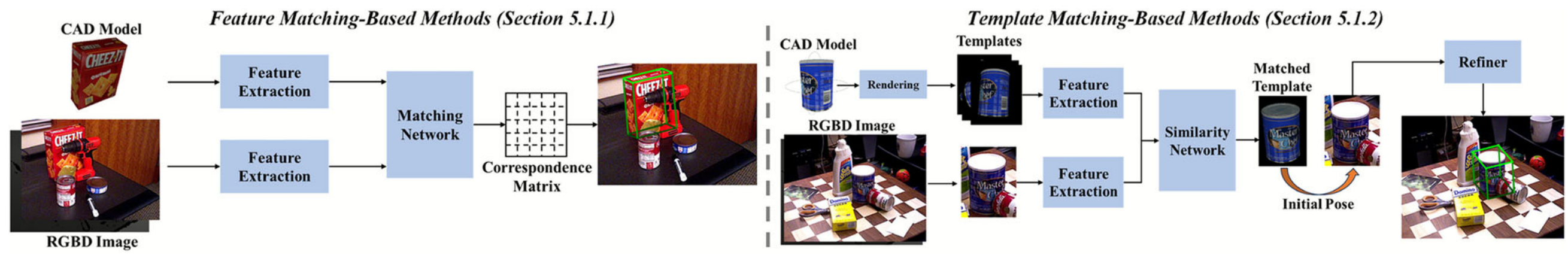

**Fig. 10** Illustration of the CAD model-based methods (Sec. 5.1) for unseen object pose estimation: The feature matching-based methods (Sec. 5.1.1) focus on designing a network to match features between the CAD model and the query image, establishing correspondences (2D-3D or 3D-3D), and solving the pose using the PnP algorithm or least squares method. The template matching-based methods (Sec. 5.1.2) utilize rendered templates from the CAD model for retrieval. The initial pose is acquired based on the most similar template, and further refinement is necessary using a refiner to obtain a more accurate pose

(2023c) further improved the pose tracking accuracy through a network that combines convolutions and transformers.

To further improve the generalization of category-level methods, Goodwin et al. (2022) introduced a reference image-based zero-shot approach, which first extracts spatial feature descriptors and builds cyclical descriptor distances. Then, they established the top-k semantic correspondences for pose estimation. Zaccaria and Manhardt (2023) proposed a self-supervised framework via optical flow consistency. Very recently, Cai et al. (2024b) developed an open-vocabulary framework that aims to generalize to unseen categories using textual prompts in unseen scene images. Di Felice et al. (2024) explored zero-shot novel view synthesis based on a diffusion model for 3D object reconstruction, and recovered the object pose through correspondences. Lin et al. (2024c) used a pre-trained vision-language model to make full use of rich semantic knowledge and align the representations of the three modalities (image, point cloud, and text) in the feature space through multi-modal contrastive learning.

**Brief Discussion:** On the whole, these shape prior-free methods circumvent the reliance on shape priors and further improve the generalization ability of category-level object pose estimation methods. Nevertheless, these methods are limited to generalizing within intra-class unseen objects. For objects of different categories, additional training data must be collected and the model must be retrained. This remains a significant limitation.

# 5 Unseen Object Pose Estimation

Unseen object pose estimation methods can further generalize to object categories not encountered during training, without retraining the model. Point Pair Features (PPF) (Drost et al., 2010) is a classical method for unseen object pose estimation that utilizes oriented point pair features to build global model description and a fast voting scheme to match locally. The final pose is solved by pose clustering and iterative closest point (Besl & McKay, 1992) refinement. However, PPF suffers from low accuracy and slow runtime, limiting its applicability. In contrast, deep learning-based methods leverage neural networks to learn more complex features from data without specifically designed feature engineering, thus enhancing accuracy and efficiency. In this section, we review the deep learning-based unseen object pose estimation methods and classify them into CAD model-based (Sec. 5.1) and manual reference view-based (Sec. 5.2) methods. The illustration of these two categories of methods is shown in Fig. 10 and Fig. 11, respectively.

## 5.1 CAD Model-Based Methods

The CAD model-based methods involve utilizing the object CAD model as prior knowledge during the process of estimating the pose of an unseen object. These methods can be further categorized into feature matching-based (Sec. 5.1.1) and template matching-based (Sec. 5.1.2) methods. The illustration of these two categories of methods is shown in Fig. 10. The characteristics and performance of some representative methods are shown in Table 3.

### 5.1.1 Feature Matching-Based Methods

Feature matching-based methods focus on designing a network to match features between the CAD model and the query image, establishing 2D-3D or 3D-3D correspondences, and solving the pose by the PnP algorithm or least squares method. As an early exploratory work, Pitteri et al. (2019) proposed a 3DoF pose estimation approach that approximates object's geometry using only the corner points of the CAD model. Nonetheless, it only works effectively on objects having specific corners. Hence, Pitteri et al. (2020) further introduced an embedding that captures the local geometry of 3D points on the object surface. Matching these embeddings can create 2D-3D correspondences, and the pose is then determined using the PnP+RANSAC (Fischler &

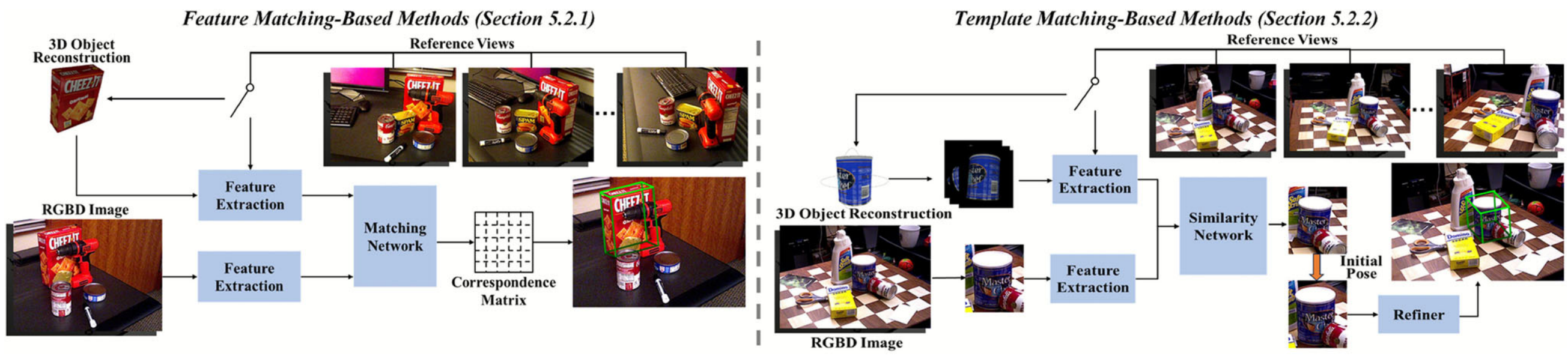

**Fig. 11** Illustration of the manual reference view-based methods (Sec. 5.2) for unseen object pose estimation: There are two types of feature matching-based methods (Sec. 5.2.1). One involves extracting features from reference views and the query image, obtaining 3D-3D correspondences through a feature matching network. The other initially reconstructs the 3D object representation using the reference views and establishes the 2D-3D correspondences between the query image and the 3D representation. The object pose is solved using correspondence-based algorithms, like PnP or the least squares method. Template matching-based methods (Sec. 5.2.2) also have two types. One reconstructs the 3D object representation using the reference views and then renders multiple templates. The initial pose is acquired by retrieving the most similar template and then refining it to get the final pose. The other directly uses the reference views as the templates for template matching

Bolles, 1981) algorithm. However, these methods only estimate the 3DoF pose.

Gou et al. (2022) defined the challenge of estimating the 6DoF pose of unseen objects, offering a baseline solution through the identification of 3D correspondences between object and scene point clouds. Similarly, Hagelskjær and Haugaard (2023) trained a network to match keypoints from the CAD model to the object point cloud. Yet, it focuses on bin picking with homogeneous bins, which only demonstrates that generalized pose estimation can achieve outstanding performance in restricted scenarios.

Inspired by point cloud registration methods on unseen objects, Zhao et al. (2023) proposed a geometry correspondence-based method using generic and object-agnostic geometry features to establish unambiguous and robust 3D-3D correspondences. Nevertheless, it still needs to get the class label and segmentation mask of unseen objects through other methods such as Mask-RCNN (He et al., 2017). To this end, Chen et al. (2023b) explored a framework named ZeroPose, which realizes joint instance segmentation and pose estimation of unseen objects. Specifically, they utilized the foundation model SAM (Kirillov et al., 2023) to generate possible object proposals and adopted a template matching method to accomplish instance segmentation. After that, they developed a hierarchical geometric feature matching network based on GeoTransformer (Qin et al., 2022b) to establish correspondences. Following ZeroPose, Lin et al. (2024a) devised a novel matching score in terms of semantics, appearance, and geometry to obtain better segmentation. As for pose estimation, they proposed a two-stage partial-to-partial point matching model to construct dense 3D-3D correspondence effectively.

Besides these methods employing geometry features, Caraffa et al. (2024) devised a method that fuses visual and geometric features extracted from different pre-trained models to enhance pose prediction stability and accuracy. It is the first technique to estimate the unseen object pose by utilizing the synergy between geometric and vision foundation models. Additionally, Huang et al. (2024) proposed a method for object pose prediction from RGBD images by combining 2D texture and 3D geometric cues.

To sum up, feature matching-based methods aim to extract generic object-agnostic features and achieve strong correspondences by matching these features. However, these methods require not only robust feature matching models but also tailored designs to enhance the representation of object features, presenting a significant challenge.

### 5.1.2 Template Matching-Based Methods

Template matching has been widely used in computer vision and stands as an effective solution for tackling the pose estimation challenges posed by unseen objects. Specifically, they utilize rendered templates from the CAD model for retrieval. The initial pose is acquired based on the most similar template, and further refinement is necessary using a refiner to obtain a more accurate pose. Wohlhart and Lepetit (2015) and Balntas et al. (2017) were pioneers in using deep pose descriptors for object matching and pose retrieval. However, their descriptors are tailored to specific orientations and categories, limiting their utility to objects with similar appearances. In contrast, Sundermeyer et al. (2020) proposed a single-encoder-multi-decoder network for jointly estimating the 3D rotation of multiple objects. This approach eliminates the need to segregate views of different objects in the latent space and enables the sharing of common features in the encoder. Yet, it still requires training multiple decoders. Wen et al. (2022) addressed this problem by decoupling object shape and pose in the latent representation, enabling auto-encoding without the necessity of multi-path decoders for different objects, thus enhancing scalability.

**Table 3** Representative CAD-based methods. Since the domain training paradigm of most unseen object pose estimation methods is domain generalization, we report 9 properties for each method, which have the same meanings as described in the caption of Table. 1. D, S, F, T, P, R, and V denote object detection, instance segmentation, feature matching to build correspondences, template matching to retrieve pose, pose solution/regression, pose refinement, and pose voting, respectively. We report the *BOP-M* across the LM-O and YCB-V datasets (Sec. 2) for various methods. Notably, these methods use Mask-RCNN (He et al., 2017) (normal font), CNOS (Nguyen et al., 2023) or their own proposed methods (Chen et al., 2023b; Lin et al., 2024a; Shugurov et al., 2022) (**bold**), and a combination of PPF and SIFT (Okorn et al., 2021) (*italics*) for unseen object location, respectively. Moreover, Örnek et al. (2023) and Caraffa et al. (2024) don't require any task-specific training, we use "×" to denote it

| Methods | | | Published Year | Training Input | Inference Input | Pose DoF | Object Property | Task | Inference Mode | Application Area | LM-O *BOP-M* | YCB-V *BOP-M* |
|---|---|---|---|---|---|---|---|---|---|---|---|---|
| CAD Model-Based Methods | Feature matching | Pitteri et al. (2020) | 2020 | RGB, CAD Model | RGB, CAD Model | 3DoF | rigid | estimation | three-stage, S+F+P | general | - | - |
| | | Zhao et al. (2023) | 2023 | Depth, CAD Model | Depth, CAD Model | 6DoF | rigid | estimation | three-stage, S+F+P | general | 65.2 | - |
| | | Chen et al. (2023b) | 2023 | RGBD, CAD Model | RGBD, CAD Model | 6DoF | rigid | estimation | four-stage, S+F+P+R | general | **49.1** | **57.7** |
| | | Lin et al. (2024a) | 2024 | RGBD, CAD Model | RGBD, CAD Model | 6DoF | rigid | estimation | three-stage, S+F+P | general | **69.9** | **84.5** |
| | | Huang et al. (2024) | 2024 | RGBD, CAD Model | RGBD, CAD Model | 6DoF | rigid | estimation | three-stage, S+F+P | general | **56.2** | **60.8** |
| | | Caraffa et al. (2024) | 2024 | × | RGBD, CAD Model | 6DoF | rigid | estimation | four-stage, S+F+P+R | general | **69.0** | **85.3** |
| | Template matching | Sundermeyer et al. (2020) | 2020 | RGB, CAD Model | RGB, CAD Model | 6DoF | rigid | estimation | three-stage, D/S+T+R | general | - | - |
| | | Okorn et al. (2021) | 2021 | RGBD, CAD Model | RGBD, CAD Model | 6DoF | rigid | estimation | two-stage, P+V | general | *59.8* | *51.6* |
| | | Shugurov et al. (2022) | 2022 | RGB/RGBD, CAD Model | RGB/RGBD, CAD Model | 6DoF | rigid | estimation | three-stage, S+T+P | general | **46.2** | **54.2** |
| | | Nguyen et al. (2022b) | 2022 | RGB, CAD Model | RGB, CAD Model | 3DoF | rigid | estimation | two-stage, D+T | occlusion | - | - |
| | | Labbé et al. (2022) | 2022 | RGB/RGBD, CAD Model | RGB/RGBD, CAD Model | 6DoF | rigid | estimation, refinement | three-stage, D+T+R | general | 58.3 | 63.3 |
| | | Örnek et al. (2023) | 2023 | × | RGB, CAD Model | 6DoF | rigid | estimation | four-stage, S+T+P+R | general | **61.0** | **69.0** |
| | | Nguyen et al. (2024b) | 2024 | RGB, CAD Model | RGB, CAD Model | 6DoF | rigid | estimation | four-stage, D+T+P+R | general | **63.1** | **65.2** |
| | | Moon et al. (2024) | 2024 | RGB/RGBD, CAD Model | RGB/RGBD, CAD Model | 6DoF | rigid | refinement | two-stage, D+R | general | **62.9** | **82.5** |
| | | Wang et al. (2024) | 2024 | RGB, CAD Model | RGB, CAD Model | 3DoF | rigid | estimation | two-stage, D+T | general | - | - |
| | | Wen et al. (2024) | 2024 | RGBD, CAD Model | RGBD, CAD Model | 6DoF | rigid | estimation, tracking | four-stage, D+T+R+V | general | 78.8 | 88.0 |

Instead of training a network to learn features across objects, Okorn et al. (2021) first generated candidate poses by PPF (Drost et al., 2010) and projected each into the scene. Later, they designed a scoring network to evaluate the hypothesis by comparing color and geometry differences between the projected object point cloud and RGBD image. Busam et al. (2020) reformulated 6DoF pose retrieval as an action decision process and determined the final pose by iteratively estimating probable movements. Cai et al. (2022a) retrieved various candidate viewpoints from a target object viewpoint codebook, and then conducted in-plane 2D rotational regression on each retrieved viewpoint to obtain a set of 3D rotation estimates. These estimates were evaluated using a consistency score to generate the final rotation prediction. Meanwhile, Shugurov et al. (2022) matched the detected objects with the rendering database for initial viewpoint estimation. Then, they predicted the dense 2D-2D correspondences between the template and the image via feature matching. Pose estimation was eventually performed by using PnP+RANSAC (Fischler & Bolles, 1981) or Kabsch (Kabsch, 1976)+RANSAC.

Since estimating the full 6DoF pose of an unseen object is extremely challenging, some works focus on estimating the 3D rotation to simplify it. Different from the previous works (Wohlhart and Lepetit, 2015; Balntas et al., 2017; Corona et al., 2018; Sundermeyer et al., 2018) that exploited a global image representation to measure image similarity, Nguyen et al. (2022b) used CNN-extracted local features to compare the similarity between the input image and templates, showing better property and occlusion robustness over global representation. Another noteworthy approach is an image retrieval framework based on multi-scale local similarities developed by Zhao et al. (2022). They extracted feature maps of various sizes from the input image and devised a similarity fusion module to robustly predict image similarity scores from multi-scale pairwise feature maps. Further, Thalhammer et al. (2023) and Ausserlechner et al. (2023) extended the scheme of Nguyen et al. (2022b) and demonstrated that the pre-trained Vision-Transformer (ViT) (Dosovitskiy et al., 2021) outperforms task-specific fine-tuned CNN (LeCun et al., 1989) for template matching. However, these methods still have a noticeable performance gap between seen and unseen objects. To this end, Wang et al. (2024) introduced diffusion features that show great potential in modeling unseen objects. Furthermore, they designed three aggregation networks to efficiently capture and aggregate diffusion features at different granularities, thus improving its generalizability.

In order to further improve the generalization and robustness of the 6DoF pose estimation, Labbé et al. (2022) used a render-and-compare approach and a coarse-to-fine strategy. Notably, they leveraged a large-scale 3D model dataset to generate a synthetic dataset containing 2 million images and over 20,000 models. It achieved strong generalization by training the network on this dataset. Compared to the non-differentiable rendering pipeline of Labbé et al. (2022), Tremblay et al. (2023) utilized recent advancements in differentiable rendering to design a flexible refiner, allowing fine-tuning the setup without retraining. On the other hand, Moon et al. (2024) presented a shape-constraint recurrent flow framework, which predicts the optical flow between the template and query image and refines the pose iteratively. It took advantage of shape information directly to improve the accuracy and scalability. Recently, Wen et al. (2024) inherited the idea of Labbé et al. (2022) and developed a novel synthesis data generation pipeline using emerging large-scale 3D model databases, Large Language Models (LLMs), and diffusion models. It greatly expanded the amount and diversity of data, and ultimately achieved comparable results to instance-level methods in a render-and-compare manner.

It is well known that template matching methods are sensitive to occlusions and require considerable time to match numerous templates. Therefore, Nguyen et al. (2024b) achieved rapid and robust pose estimation by finding the suitable trade-off between the use of template matching and patch correspondences. In particular, the features of the query image and templates are extracted using the ViT (Dosovitskiy et al., 2021), followed by fast template matching using a sub-linear nearest neighbor search. The most similar template provides two DoFs for azimuth and elevation, while the remaining four DoFs are obtained by constructing correspondences between the query image and this template. Örnek et al. (2023) utilized DINOv2 (Oquab et al., 2023) to extract descriptors for the query image and templates. Moreover, they introduced a fast template retrieval method based on visual words constructed from DINOv2 patch descriptors, thereby decreasing the reliance on extensive data and enhancing matching speed compared to Labbé et al. (2022).

In summary, template matching-based methods make full use of the advantages provided by a multitude of templates, enabling high accuracy and strong generalization. Nonetheless, they have limitations in terms of time consumption, sensitivity to occlusions, and challenges posed by complex backgrounds and lighting variations.

**Brief Discussion:** The above-mentioned feature matching-based or template matching-based methods both require a CAD model of the target object to provide prior information. In practice, accurate CAD models often require specialized hardware to build, which limits the practical application of these methods to a certain extent.

### 5.2 Manual Reference View-Based Methods

Aside from these CAD model-based approaches, there are some manual reference view-based methods that do not require the unseen object CAD model as a prior condition but instead require providing some manual labeled reference views with the target object. Similar to CAD model-based methods, these methods are also categorized into two types: feature matching-based (Sec. 5.2.1) and template matching-based (Sec. 5.2.2) methods. These two categories of methods are illustrated in Fig. 11. The attributes and performance of some representative methods are shown in Table 4.

#### 5.2.1 Feature Matching-Based Methods

Different from CAD model-based feature matching methods, manual reference view-based feature matching methods primarily establish 3D-3D correspondences between the RGBD query image and RGBD reference images, or 2D-3D correspondences between the query image and sparse point cloud reconstructed by reference views. Subsequently, the object pose is solved according to the different correspondences. He et al. (2022c) proposed the first few-shot 6DoF object pose estimation method, which can estimate the pose of an unseen object by a few support views without extra training. Specifically, they desinged a dense RGBD prototype matching framework based on transformers to fully explore the semantic and geometric relationship between the query image and reference views. Corsetti et al. (2024) used a textual prompt for object segmentation and reformulated the problem as a relative pose estimation between two scenes. The relative pose was obtained via point cloud registration.

Some methods took an alternative route from the perspective of matching after reconstruction. Wu et al. (2021b) developed a global registration-based method that used reference and query images to reconstruct full-view and single-view models, and then searched for point matches between the two models. Sun et al. (2022b) drew inspiration from visual localization and revised the pipeline to adapt it for pose estimation. More precisely, they reconstructed a Structure from Motion (SfM) model of the unseen object using RGB sequences from all reference viewpoints. Then, they matched 2D keypoints in the query image with the 3D points in the SfM model by a graph attention network. Nevertheless, it performed poorly on low-textured objects because of its reliance on repeatably detected keypoints. To deal with this problem, He et al. (2022a) designed a new keypoint-free SfM method to reconstruct semi-dense point cloud models of low-textured objects based on the detector-free feature matching method LoFTR (Sun et al., 2021). Castro and Kim (2023) pointed out that these pre-trained feature matching models (Sarlin et al., 2020; Sun et al., 2021) fail to capture the optimal descriptions for pose estimation. Based on this, they redesigned the training pipeline based on a three-view system for one-shot object-to-image matching.

The aforementioned works still require dense support views (*i.e.*, $\geq$ 32 views). To address this problem, Fan et al. (2024a) turned the 6DoF object pose estimation task into relative pose estimation between the retrieved object in the target view and the reference view. Given only one reference view, they achieved it by using the DINOv2 model (Oquab et al., 2023) for global matching and the LoFTR model (Sun et al., 2021) for local matching. Note that this method cannot estimate absolute translation (or object scale), as this is an ill-posed problem when only considering two views. Beyond that, Lee et al. (2024) applied a powerful pre-trained technique tailored for 3D vision (Weinzaepfel et al., 2022) and demonstrated geometry-oriented visual pre-training can get better generalization capability with fewer reference views.

Generally, due to the lack of prior geometric information from CAD models, manual reference view-based feature matching methods often require special designs to extract the geometric features of unseen objects. The number of reference views also constrains the actual application of such approaches to a certain extent.

#### 5.2.2 Template Matching-Based Methods

Template matching-based methods mainly adopt the strategy of retrieval and refinement. There are two types: one reconstructs a 3D object representation using reference views, renders multiple templates based on this 3D representation, and employs a similarity network to compare the query image with each template for the initial pose. A refiner is then used to refine this initial pose for increased accuracy. The other directly uses reference views as templates, requiring plenty of views for retrieval and greater reliance on a refiner for accuracy. Park et al. (2020a) introduced a novel framework for pose estimation of unseen objects without the CAD model. They reconstructed 3D object representations from a few reference views, followed by estimating translation using mask bounding boxes and corresponding depth values. The initial rotation was determined by sampling angles and refined using gradient updates via a render-and-compare approach. By training the network to render and reconstruct diverse 3D shapes, it achieved excellent generalization performance on unseen objects.

Unlike Park et al. (2020a) that used the strategy of render-and-compare after reconstruction, Liu et al. (2022f) designed a pipeline for detection, retrieval, and refinement. They first designed a detector to identify object bounding boxes in the target view. Next, they compared the query and reference images at the pixel level to acquire the initial pose based on the similarity score. The pose was then refined using feature volume and multiple 3D convolution layers. However, object-centered reference images from cluttered scenes are constrained by actual segmentation or bounding box cropping, limiting its real-world applicability. To overcome this limitation, Gao et al. (2023) proposed adaptive segmentation modules to learn distinguishable representations of unseen objects, and Zhao et al. (2024) leveraged distributed reference kernels and translation estimator to achieve multi-scale correlation computation and object translation parameter prediction, thus robustly learning the prior translation of unseen objects.

**Table 4** Representative manual reference view-based methods. For each method, we report its 9 properties, which have the same meanings as described in the caption of Table. 1. D, S, F, T, P, R, and V have the same meanings as Table. 3. We report the average recall of ADD(S) within 10% of the object diameter, termed as ADD(S)-0.1d (Sec. 2). Notably, "YOLOv5" and "GT" denote the use of YOLOv5 (Ultralytics, 2022) and ground-truth bounding box/segmentation mask for object localization, respectively. For a fair comparison, we also report the number of used reference views. "Full" represents all views

| Methods | | | Published Year | Training Input | Inference Input | Pose DoF | Object Property | Task | Inference Mode | Application Area | LM ADD(S)-0.1d |
|---|---|---|---|---|---|---|---|---|---|---|---|
| Manual Reference View-Based Methods | Feature matching | He et al. (2022c) | 2022 | RGBD | RGBD | 6DoF | rigid | estimation | three-stage, D+F+P | general | 83.4 (GT+16) |
| | | Sun et al. (2022b) | 2022 | RGB | RGB | 6DoF | rigid | estimation | three-stage, D+F+P | general | 63.6 (YOLOv5+Full) |
| | | He et al. (2022a) | 2022 | RGB | RGB | 6DoF | rigid | estimation | three-stage, D+F+P | general | 76.9 (YOLOv5+Full) |
| | | Castro and Kim (2023) | 2023 | RGB | RGB | 6DoF | rigid | estimation | three-stage, D+F+P | general | 87.5 (GT+Full) |
| | | Lee et al. (2024) | 2024 | RGB | RGB | 6DoF | rigid | estimation | three-stage, D+F+P | general | 78.4 (YOLOv5+64) |
| | Template matching | Park et al. (2020a) | 2020 | RGBD | RGBD | 6DoF | rigid | estimation | three-stage, S+T+R | general | 87.1 (GT+16) |
| | | Nguyen et al. (2022a) | 2022 | RGB | RGB | 6DoF | rigid | tracking | three-stage, D+S+P | general | - |
| | | Liu et al. (2022f) | 2022 | RGB | RGB | 6DoF | rigid | estimation | three-stage, D+T+R | general | - |
| | | Gao et al. (2023) | 2023 | RGB | RGB | 6DoF | rigid | estimation | four-stage, S+D+P+R | occlusion | - |
| | | Cai et al. (2024a) | 2024 | RGB | RGB | 6DoF | rigid | estimation | two-stage, P+R | general | 81.7 (Full) |
| | | Wen et al. (2024) | 2024 | RGBD | RGBD | 6DoF | rigid | estimation, tracking | four-stage, D+T+R+V | general | 99.9 (GT+16) |

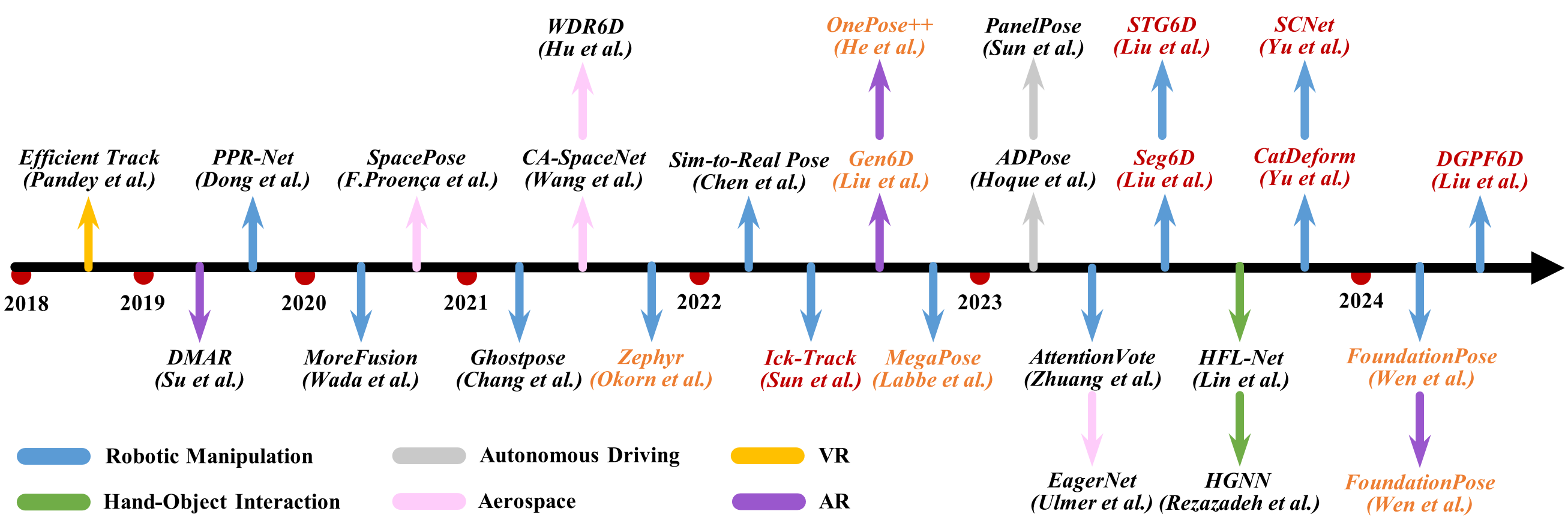

**Fig. 12** Chronological overview of some representative applications of object pose estimation methods. The black references, red references, and orange references represent the application of instance-level, category-level, and unseen object pose estimation methods, respectively. From this, we can also see the development trend, *i.e.*, from instance-level methods to category-level and unseen object pose estimation methods

To further enhance the robustness of the translation estimation for object detection, Pan et al. (2024) modified the framework of Liu et al. (2022f). Precisely, they utilized pretrained ViT (Dosovitskiy et al., 2021) to learn robust feature representations and adopted a top-K pose proposal scheme for pose initialization. Additionally, they applied a coarse-to-fine cascaded refinement process, incorporating feature pyramids and adaptive discrete pose hypotheses. Besides Pan et al. (2024), Cai et al. (2024a) revisited the pipeline of Liu et al. (2022f). They proposed a generic joint segmentation method and an efficient 3D Gaussian Splatting-based refiner, improving the performance and robustness of object localization and pose estimation.

In unseen object tracking, Nguyen et al. (2022a) proposed the first method that extended to invisible categories without requiring 3D information and extra reference images, given the ground-truth object pose in the first frame. Their transformer-based architecture outputs continuous relative object poses between consecutive frames, combined with the initial object pose, to provide the object pose for each frame. Wen et al. (2023) used the collaborative design of concurrent tracking and neural object fields to perform 6DoF tracking from RGBD sequences. Key aspects of it include online pose graph optimization, concurrent neural object fields for 3D shape and appearance reconstruction, and a memory pool facilitating communication between the two processes.

More recently, Nguyen et al. (2024a) reconsidered template matching from the perspective of generating new views. Given a single reference view, they trained a model to directly predict the discriminative embeddings of the novel viewpoints of the object. In contrast, Wen et al. (2024) applied an object-centric neural field representation for object modeling and RGBD rendering.

**Brief Discussion:** In summary, similar to template matching-based methods using CAD models, manual reference view-based methods also rely on massive templates. Moreover, due to limited reference views, these methods need to generate new templates or employ additional strategies to optimize the initial pose obtained through template matching.

## 6 Applications

With the advancement of object pose estimation technology, several applications leveraging this progress have been deployed. In this section, we elaborate on the development trends of these applications. Specifically, these applications include robotic manipulation (Sec. 6.1), Augmented Reality (AR)/Virtual Reality (VR) (Sec. 6.2), aerospace (Sec. 6.3), hand-object interaction (Sec. 6.4), and autonomous driving (Sec. 6.5). The chronological overview is shown in Fig. 12. The illustration of application scenarios is shown in Fig. 13.

### 6.1 Robotic Manipulation

We categorize the robotic manipulation application into instance-level, category-level, and unseen objects. This clas-

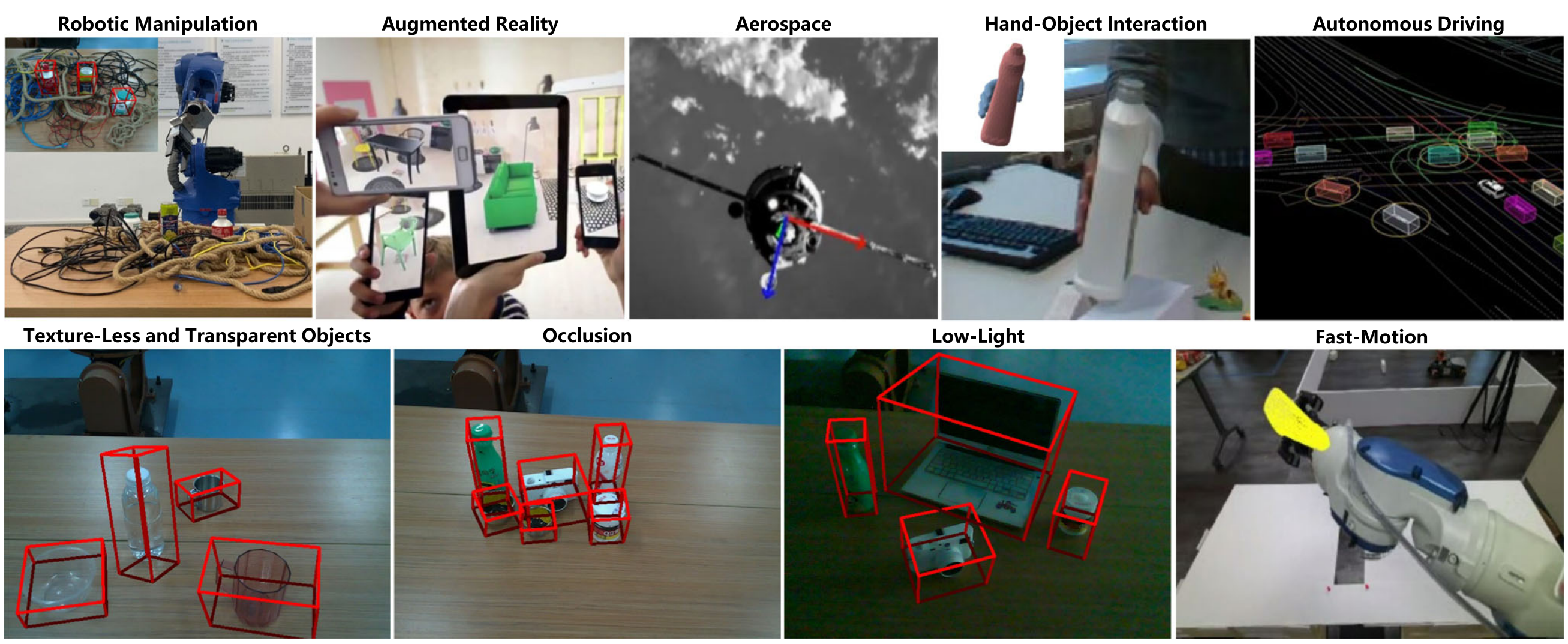

**Fig. 13** Illustration of application scenarios of object pose estimation. The first row shows its applications in robotic manipulation, augmented reality, aerospace, hand–object interaction, and autonomous driving. The second row presents examples involving challenging objects and scenes

sification helps in better understanding the challenges and requirements across different levels.

### 6.1.1 Instance-Level Manipulation

To tackle the challenge of annotating real data during training, many works utilize synthetic data for training as it is easy to acquire and annotate (Park et al., 2020b; Chen et al., 2023a). At the same time, synthetic data can simulate various scenes and environmental changes, thus helping to improve the adaptability of robotic manipulation. Li et al. (2018b) used a large-scale synthetic dataset and a small-scale weakly labeled real-world dataset to reduce the difficulty of system deployment. Additionally, Chen et al. (2022b) proposed an iterative self-training framework, using a teacher network trained on synthetic data to generate pseudo-labels for real data. Meanwhile, Fu et al. (2022) trained only on synthetic images based on physical rendering. One of the critical challenges of synthetic data is bridging the gap with reality, which Tremblay et al. (2018) addressed by combining domain randomization with real data.

Handling stacked occlusion scenes is another significant challenge, especially in industrial automation and logistics. In these scenarios, robots must accurately identify and localize objects stacked on each other, which requires an effective process of occluded objects and accurate pose estimation. Dong et al. (2019) argued that the regression poses of points from the same object should tightly reside in the pose space. Therefore, these points can be clustered into different instances, and their corresponding object poses can be estimated simultaneously. This method can handle severe object occlusion. Moreover, Zhuang et al. (2024) established an end-to-end pipeline to synchronously regress all potential object poses from an unsegmented point cloud. Most recently, Wada et al. (2020) proposed a system that fully utilized identified accurate object CAD models and non-parametric reconstruction of unrecognized structures to estimate the occluded objects pose in real-time.

Low-textured objects lack object surface texture information, making robotic manipulation challenging. Therefore, Zhang and Cao (2019) proposed a pose estimation method for texture-less industrial parts. Poor surface texture and brightness make it challenging to compute discriminative local appearance descriptors. This method achieves more accurate results by optimizing the pose in the edge image. In addition, Chang et al. (2021) carried out transparent object grasping by estimating the object pose using a proposed model-free method that relies on multiview geometry. In agricultural scenes, Kim et al. (2022) constructed an automated data collection scheme based on a 3D simulator environment to achieve three-level ripeness classification and pose estimation of target fruits.

### 6.1.2 Category-Level Manipulation

To investigate the application of category-level object pose estimation for robotic manipulation, Liu et al. (2023a) introduced a fine segmentation-guided category-level method with difference-aware shape deformation for robotic grasping. Yu et al. (2023a) proposed a shape prior-based approach and explored its application for robotic grasping. Further, Liu et al. (2023d) developed a robotic continuous grasping system with a pre-defined vector orientation-based grasping strategy, based on shape transformer-guided object pose estimation. To improve efficiency and enable the pose estimation method to be applied to tasks with higher real-time requirements, Sun et al. (2022a) utilized the inter-frame consistent keypoints to perform object pose tracking for aerial manipulation. To further avoid manual data annotation in the real-world scene, Yu et al. (2023b) built a robotic grasping platform and designed a self-supervised-based method for category-level robotic grasping. More recently, Liu et al. (2024a) explored a contrastive learning-guided prior-free object pose estimation method for domain-generalized robotic picking. Zhou et al. (2025) proposed a single-shot SSM-based framework that predicts the full 3D information (pose, size, and shape) of multiple object for end-to-end robotic scene understanding.

#### 6.1.3 Unseen Object Manipulation

Since unseen object pose estimation belongs to an emerging research, there is currently a lack of specialized designs for robotic. Here, we report several methods that validate the effectiveness of unseen object pose estimation through robotic manipulation. Okorn et al. (2021) introduced a method for zero-shot object pose estimation in clutter. By scoring pose hypotheses and choosing the highest-scoring pose, they successfully grasped a novel drill object using a robotic arm. Labbé et al. (2022) and Wen et al. (2024) adopted the render-and-compare strategy and trained the network on a large-scale synthetic dataset, resulting in an outstanding generalization. They further verified the effectiveness of their methods through robotic grasping experiments.

### 6.2 Augmented Reality/Virtual Reality

Object pose estimation has various specific applications in AR and VR fields. In AR, accurate pose estimation allows for a precise overlay of virtual objects onto the real world. The key to VR technology lies in tracking the head-mounted display pose and controller in 3D space.

Su et al. (2019) combined two CNN architectures (LeCun et al., 1989) into a network, consisting of a state estimation branch and a pose estimation branch explicitly trained on synthetic images, to achieve AR assembly applications. Pandey et al. (2018) introduced a method for automatically annotating the handheld objects pose in camera space, addressing the efficient 6Dof pose tracking problem for handheld controllers from the perspective of egocentric cameras. Liu et al. (2022f) presented a generalizable model-free 6DoF object pose estimator that has realized the complete object detection and pose estimation process. By simply capturing reference images of an unseen object and retrieving the poses of reference images, this method can predict the object pose on arbitrary query images and be easily applied to daily objects for AR/VR applications. He et al. (2022a) adopted matching after the reconstruction strategy, which establishes the correspondences between the query image and the reconstructed

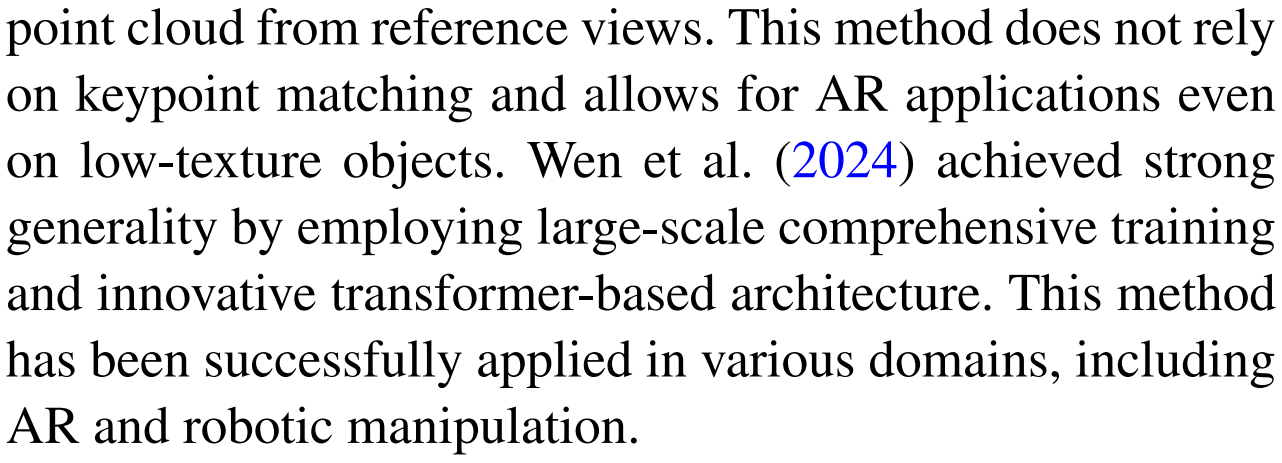
point cloud from reference views. This method does not rely on keypoint matching and allows for AR applications even on low-texture objects. Wen et al. (2024) achieved strong generality by employing large-scale comprehensive training and innovative transformer-based architecture. This method has been successfully applied in various domains, including AR and robotic manipulation.

### 6.3 Aerospace

Estimating the object pose in space presents unique challenges not commonly encountered in terrestrial environments. One of the most significant differences is the lack of atmospheric scattering, which complicates lighting conditions and makes objects invisible over long distances. In-orbit proximity operations for space rendezvous, docking, and debris removal require precise pose estimation under diverse lighting conditions and on high-texture backgrounds. Proença and Gao (2020) proposed URSO, a simulator developed on Unreal Engine 4 (UnrealEngine4, 2020) for generating annotated images of spacecraft orbiting Earth, which can be used as valuable data for aerospace application. Hu et al. (2021) proposed an encoder-decoder architecture that reliably handles large-scale changes under challenging conditions, enhancing robustness. Wang et al. (2022c) introduced a counterfactual analysis framework to achieve robust pose estimation of spaceborne targets in complex backgrounds. Ulmer et al. (2023) generated multiple pose hypotheses for objects and introduced a pixel-level posterior formula to estimate the probability of each hypothesis. This approach can handle extreme visual conditions, including overexposure, high contrast, and low signal-to-noise ratio.

### 6.4 Hand-Object Interaction

When humans/robots interact with the physical world, they primarily do so through their hands. Therefore, accurately understanding how hands interact with objects is crucial.

Hand-object interaction methods often rely on the object CAD model, and obtaining the object CAD model from daily life scenes is challenging. Patten et al. (2021) reconstructed high-quality object CAD model to mitigate the reliance on object CAD model in hand-object interaction. To further enhance hand-object interaction, Lin et al. (2023b) utilized an effective attention model to improve the representation capability of hand and object features, thereby improving the accuracy of hand and object pose estimation. However, this method has limited utilization of the underlying geometric structures, leading to an increased reliance on visual features. Performance may degrade when objects lack visual features or when these features are occluded. Therefore, Rezazadeh et al. (2023) introduced a hierarchical graph neural network architecture combined with multimodal (visual and tactile) data to compensate for visual deficiencies and improve robustness. Moreover, Qi et al. (2024) introduced a hand-object pose estimation network guided by Signed Distance Fields (SDF), which jointly leverages the SDFs of both the hand and the object to provide a complete global implicit representation. This method aids in guiding the pose estimation of hands and objects in occlusion scenarios. Yang et al. (2025b) proposed an occlusion-aware framework based on masked autoencoders, where a target-focused masking strategy and multi-scale SDF features are leveraged to enhance structural reasoning under occlusions.

### 6.5 Autonomous Driving

Object pose estimation can be used to perceive surrounding objects such as vehicles, pedestrians, and obstacles, aiding the autonomous driving system in making timely decisions.

In order to address the pose estimation problem in autonomous driving, Hoque et al. (2023) proposed a 6DoF pose hypothesis based on a deep hybrid structure composed of CNNs (LeCun et al., 1989) and RNNs (Elman, 1990). More recently, Sun et al. (2023) designed an effective keypoint selection algorithm, which takes into account the shape information of panel objects within the scene of robot cabin inspection, addressing the challenge of 6DoF pose estimation of highly variable panel objects.

### 6.6 Challenging Objects and Scenes

In practical deployment, object pose estimation models frequently encounter challenging objects and scenes that amplify the limitations of existing methods. To better guide practitioners, we provide targeted recommendations for method selection under these conditions.

**Texture-less** objects pose a major difficulty for correspondence-based methods, which depend heavily on distinctive local descriptors (Drost et al., 2010; Birdal & Ilic, 2015). In such cases, template-based (Dang et al., 2024) or regression-based (Wang et al., 2021c) approaches often provide more reliable estimates by leveraging global image evidence rather than sparse correspondences. **Transparent** and **reflective** objects present a different challenge due to noisy or missing depth. Approaches that integrate photometric and geometric reasoning (Chen et al., 2022d), or that incorporate physics-aware rendering constraints (Zhang et al., 2022a), have shown greater promise for handling these cases by compensating for unreliable sensor inputs.

Under severe **occlusions**, correspondence-based (Lian & Ling, 2023; Xu et al., 2024b) and voting-based (Peng et al., 2019; He et al., 2020) methods are typically more effective. Correspondence methods, especially dense correspondence models with strong local descriptors, can maintain stability when only partial object views are visible. Voting strategies are also advantageous in these cases, as aggregated hypotheses from visible parts can recover accurate global poses despite missing regions. Moreover, **low-light** conditions often degrade RGB-based pipelines, as appearance cues become unreliable. Since most RGBD sensors are active (using LEDs or lasers) and can provide depth and point cloud data that remain robust under challenging lighting conditions, RGBD fusion or point cloud–centric approaches are therefore recommended (Wang et al., 2019a; He et al., 2021). By explicitly integrating geometric depth information, these methods remain resilient to illumination variations and sensor noise. For improved robustness, fusion architectures that exploit cross-modality relationships rather than naive concatenation are preferable (Wang et al., 2022a; Lin et al., 2023a). Furthermore, for **fast-motion** scenarios, temporal consistency becomes critical. Methods that incorporate spatio-temporal constraints, such as keypoint tracking (Wang et al., 2020a) or pose graph optimization (Wen & Bekris, 2021), offer better stability. End-to-end regression pipelines with temporal feature integration also show promise by learning motion-aware pose embeddings that reduce prediction jitter during rapid dynamics. Finally, in tracking tasks across continuous frames, anchor-based or memory-augmented methods are particularly effective (Wang et al., 2020a; Wen & Bekris, 2021). By exploiting inter-frame coherence, these approaches reduce the reliance on per-frame detection and provide smoother, real-time performance.

Overall, practitioners should select methods by aligning the strengths of each paradigm with the specific challenges of the target application, balancing accuracy, robustness, and efficiency according to object characteristics and environmental conditions.

## 7 Emerging Trends and Future Directions

In this survey, we have provided a systematic overview of the latest deep learning-based object pose estimation methods, covering a comprehensive classification, a comparison of their strengths and weaknesses, and an exploration of their applications. Despite the great success, many challenges still exist, as discussed in Sec. 3, Sec. 4, and Sec. 5. Building on these challenges, we further highlight emerging trends and outline promising future directions to advance research in object pose estimation.

### 7.1 Emerging Trends

From the perspective of **label-efficient learning**, prevailing methodologies predominantly rely on the utilization of real-world labeled datasets for training purposes. Nevertheless, the labor-intensive nature of manually collecting and annotating training data is widely acknowledged. Hence, we advocate for the exploration of label-efficient learning techniques for object pose estimation, which can be pursued through the following avenues:
**1) LLMs/LVMs-guided weak/self-supervised learning methods.** With the rapid advancements in pre-trained LLMs/LVMs, their versatile application in various scenarios through an unsupervised manner has become feasible. Leveraging LLMs/LVMs as prior knowledge holds promise for exploring weak or self-supervised learning techniques in object pose estimation. **2) Synthesis to real-world domain adaptation and generalization methods.** Due to the high costs associated with acquiring real-world training data through manual efforts, synthetic data generation offers a cost-effective alternative. We believe that by exploring domain adaptation and generalization techniques from synthetic to real-world domains, we can mitigate domain gaps and achieve the capability to generalize synthetic data-trained models for real-world applications. **3) Novel view reconstruction for pose estimation.** Recent advances in neural rendering, including 3D Gaussians, differentiable rendering, and volumetric rendering, have enabled the synthesis of object-consistent novel views from only a few input images (Di Felice et al., 2024; Wen et al., 2024; Nguyen et al., 2024a; Legrand et al., 2024). These approaches offer a CAD-free route to densify viewpoints, regularize correspondences, and support render-and-compare estimation under sparse supervision. In category-level and unseen-category scenarios, such models can serve as priors to promote object-centric canonicalization and metric scaling, or provide uncertainty-aware multi-hypothesis images and embeddings that enhance retrieval and RANSAC-based PnP. Open challenges remain in preserving geometric fidelity under intra-class variation, mitigating generative artifacts and view inconsistency, and achieving real-time performance on robotic platforms.

In terms of **applications**, facilitating the deployment of object pose estimation methods on mobile devices and robots is crucial. We argue that enhancing the deployability of existing methods can be achieved through the following approaches:
**1) End-to-end methods integrating detection or segmentation.** Current SOTA approaches typically require initial object detection or segmentation using a pre-trained model before inputting the image into a pose estimation model (indirect pose estimation models even need to use non-differentiable PnP or Umeyama algorithms to solve pose), which complicates deployment. Future research can enhance the deployability on mobile devices and robots by exploring end-to-end object pose estimation methods that seamlessly integrate detection or segmentation.
**2) Single RGB image-based methods.** Given that most mobile devices (such as smartphones and tablets) lack depth cameras, achieving high-precision estimation of unseen object poses using a single RGB image is crucial. Due to inherent geometric limitations in 2D images, future research can explore LVMs-based monocular depth estimation methods to enhance the accuracy of monocular object pose estimation by incorporating scene-level depth information.
**3) Model lightweighting.** Existing SOTA models often have large parameter sizes and inefficient running performance, which presents challenges for deployment on mobile devices and robots with limited computational resources. Future work can explore effective lightweight methods, such as teacher-student models, to research reducing model parameter count (GPU memory) and improving model running efficiency.

Existing methods are predominantly designed for common objects and scenes, rendering them ineffective for **challenging objects and scenes**. We believe that the applicability can be enhanced through the following avenues:
**1) Articulated object pose estimation.** Articulated objects (such as clothing and drawers) exhibit multiple DoF and significant self-occlusion compared to rigid objects, making pose estimation challenging. Achieving high-precision pose estimation for articulated objects is an important research problem that remains to be addressed in the future.
**2) Transparent object pose estimation.** The simultaneous absence of texture, color, and depth information poses a significant challenge for estimating the pose of transparent objects. Future research endeavors could focus on enhancing the geometric information of transparent objects through depth augmentation or completion techniques, thereby improving the accuracy of pose estimation.
**3) Robust methods for handling occlusion.** Occlusion is the most common challenge. Currently, there exists no object pose estimation method that can effectively handle severe occlusion. Severe occlusion leads to an incomplete representation of texture and geometric features in objects, introducing uncertainty into the pose estimation model. Hence, improving the model's ability to perceive severe occlusion is crucial for enhancing its robustness.

From the aspect of **problem formulation**, recent instance-level methods have achieved high precision but exhibit poor generalization. Category-level methods demonstrate good generalization for intra-class unseen objects but fail to generalize to unseen object categories. Unseen object pose estimation methods have the potential to generalize to any unseen object, yet they still rely on object CAD models or reference views. The following paths can be explored from the problem formulation to further enhance the generalization of object pose estimation:
**1) Few-shot learning-based category-level methods for unseen categories.** Since category-level methods need to re-obtain a large amount of annotated training data for unseen object categories, their generalization is severely limited. Therefore, future research could focus on exploring how to leverage few-shot learning to enable the rapid generalization of category-level methods to unseen object categories.

**2) CAD model-free and sparse manual reference view-based unseen object pose estimation.** While current unseen object pose estimation methods do not require retraining for unseen objects, they still rely on either the CAD models or extensive annotated reference views of unseen objects, both of which still require manual acquisition. To this end, exploring CAD model-free and sparse manual reference view-based unseen object pose estimation methods is crucial.

**3) Open-vocabulary strong generalization methods.** Given the broad applicability of object pose estimation in human-machine interaction scenes, future research could leverage open vocabulary provided by humans as prompts to enhance generalization to unseen objects and scenes.

## 7.2 Future Directions

While emerging trends primarily reflect ongoing developments, future work should focus on open-ended directions and broader applications that remain underexplored. Building on the discussions in the preceding sections, we summarize the following research directions that can substantially advance the field.

**1) Improving generalization to intra-class unknown and unseen objects:** While instance- and category-level methods have shown impressive accuracy on trained objects, their performance often degrades when facing intra-class unknown or category-unseen objects. Future research could explore disentangled geometric representations that encode both category-level priors and fine-grained instance variations. Leveraging generative shape modeling, compositional learning, or implicit neural representations may enable networks to infer pose from limited supervision and adapt to new object geometries without retraining.

**2) Foundation model-based pose representation and adaptation:** The rapid progress of large-scale vision and vision-language foundation models offers new opportunities for transferable 3D understanding. Integrating these models into pose estimation pipelines could allow pretraining on massive visual data while adapting to the geometry of novel objects via lightweight finetuning or prompt-based adaptation. Developing geometry-aware adapters or SE(3)-equivariant heads for such models may bridge the gap between semantic representation learning and geometric reasoning.

**3) Label-efficient and self-supervised learning:** A major bottleneck in current methods is the dependence on dense 6D pose annotations. Future work could focus on label-efficient learning paradigms, including weakly-supervised, self-supervised, and active learning approaches that exploit geometric constraints, photometric consistency, or multi-view correspondences to learn pose without explicit labels. This will be crucial for scaling pose estimation to large and diverse object collections.

**4) Uncertainty modeling and temporal consistency:** Current models typically output a single deterministic pose, which limits their robustness in ambiguous or occluded conditions. Future methods could explicitly model pose uncertainty to better handle symmetric, transparent, or reflective objects. In dynamic environments, temporal consistency constraints through tracking or sequential optimization can also help stabilize predictions and improve overall reliability.

**5) Cross-modal and physically grounded estimation:** Beyond visual input, tactile and inertial modalities provide complementary cues for robust pose inference. Future work may explore multimodal fusion strategies that incorporate physical constraints, such as contact geometry or force sensing, to estimate pose in manipulation or interaction scenarios. The combination of visual and physical feedback could lead to more accurate and physically consistent pose estimation.

**6) Scalable benchmarks and realistic simulation:** As illustrated in Sec. 2, most benchmarks remain limited in scale, diversity, or realism. Future efforts should aim to develop large-scale, physics-aware simulation environments and real-world datasets with fine-grained annotations for both rigid and articulated objects. Such resources will support more comprehensive evaluation and training of generalizable models.

Together, these research avenues represent critical steps toward achieving robust, data-efficient, and generalizable object pose estimation across diverse domains and real-world conditions.

**Acknowledgements** This work was supported by the National Natural Science Foundation of China under Grant U22A2059 and Grant 62473141, China Scholarship Council under Grant 202306130074 and Grant 202206130048, Natural Science Foundation of Hunan Province under Grant 2024JJ5098, Open Foundation of the State Key Laboratory of Advanced Design and Manufacturing for Vehicle Body, and Open Foundation of the Engineering Research Center of Multi-Mode Control Technology and Application for Intelligent System of the Ministry of Education. Ajmal Mian was supported by the Australian Research Council Future Fellowship Award funded by the Australian Government under Project FT210100268. Nicu Sebe was supported by the EU Horizon project "ELIAS - European Lighthouse of AI for Sustainability" (No. 101120237) and the FIS project GUIDANCE (Debugging Computer Vision Models via Controlled Cross-modal Generation) (No. FIS2023-03251).

**Data Availability** The data that support the research and a more readable version of this paper are available at Awesome-Object-Pose-Estimation.

# References

Ahmadyan, A., Zhang, L., Ablavatski, A., Wei, J., & Grundmann, M. (2021) Objectron: A large scale dataset of object-centric videos in the wild with pose annotations. In *CVPR*, (pp. 7822–7831).

An, Y., Yang, D., & Song, M. (2024). Hft6d: Multimodal 6d object pose estimation based on hierarchical feature transformer. *Measurement, 224*, Article 113848.

Ausserlechner P, Haberger D, Thalhammer S, Weibel, J.-B., & Vincze, M. (2023). Zs6d: Zero-shot 6d object pose estimation using vision transformers. arXiv:2309.11986

Balntas, V., Doumanoglou, A., Sahin, C., Sock, J., Kouskouridas, R., & Kim, T.-K. (2017). Pose guided rgbd feature learning for 3d object pose estimation. In *ICCV*, (pp. 3856–3864).

Bengtson, S. H., Åström, H., Moeslund, T. B., Topp E.A., & Krueger, V. (2021). Pose estimation from rgb images of highly symmetric objects using a novel multi-pose loss and differential rendering. In *IROS*, (pp. 4618–4624).

Besl, P. J., & McKay, N. D. (1992). Method for registration of 3-d shapes. In *Sensor fusion IV: control paradigms and data structures*, (pp. 586–606).

Birdal T, Ilic S (2015) Point pair features based object detection and pose estimation revisited. In: 3DV, pp 527–535

Brachmann, E., Krull, A., Michel, F., Gumhold, S., Shotton, J., & Rother, C. (2014). Learning 6d object pose estimation using 3d object coordinates. In *ECCV*, (pp. 536–551).

Brachmann, E., Michel, F., Krull, A., Yang, M. Y., Gumhold, S. & others. (2016). Uncertainty-driven 6d pose estimation of objects and scenes from a single rgb image. In *CVPR*, (pp. 3364–3372).

Bukschat, Y., & Vetter, M. (2020). *Efficientpose: An efficient, accurate and scalable end-to-end 6d multi object pose estimation approach*. arXiv:2011.04307.

Busam, B., Jung, H. J., & Navab, N. (2020). *I like to move it: 6d pose estimation as an action decision process*. arXiv:2009.12678.

Cai, D., Heikkilä, J., & Rahtu, E. (2022a). Ove6d: Object viewpoint encoding for depth-based 6d object pose estimation. In *CVPR*, (pp. 6803–6813).

Cai, D., Heikkilä, J., & Rahtu, E. (2022b). Sc6d: Symmetry-agnostic and correspondence-free 6d object pose estimation. In *3DV*, (pp. 536–546).

Cai, D., Heikkilä, J., & Rahtu, E. (2024a). *Gs-pose: Cascaded framework for generalizable segmentation-based 6d object pose estimation*. arXiv:2403.10683.

Cai, J., He, Y., Yuan, W., Zhu, S., Dong, Z., Bo, L., & Chen, Q. (2024b). *Ov9d: Open-vocabulary category-level 9d object pose and size estimation*. arXiv:2403.12396.

Cai, M., & Reid, I. (2020). Reconstruct locally, localize globally: A model free method for object pose estimation. In *CVPR*, (pp. 3153–3163).

Calli, B., Singh, A., Walsman, A., Srinivasa, S., Abbeel, P., & Dollar, A. M. (2015). The ycb object and model set: Towards common benchmarks for manipulation research. In *ICRA*, (pp. 510–517).

Cao, T., Luo, F., Fu, Y., Zhang, W., Zheng, S., & Xiao, C. (2022) Dgecn: A depth-guided edge convolutional network for end-to-end 6d pose estimation. In *CVPR*, (pp. 3783–3792).

Caraffa, A., Boscaini, D., Hamza, A., & Poiesi, F. (2024). *Freeze: Training-free zero-shot 6d pose estimation with geometric and vision foundation models*. arXiv:2312.00947.

Castro, P., & Kim, T. K. (2023). Posematcher: One-shot 6d object pose estimation by deep feature matching. In *ICCVW*.

Chang, A. X., Funkhouser, T., Guibas, L., Hanrahan, P., Huang, Q., Li, Z., Savarese, S., Savva, M., Song, S., Su, H., & others. (2015). *Shapenet: An information-rich 3d model repository*. arXiv:1512.03012.

Chang, J., Kim, M., Kang, S., Han, H., Hong, S., Jang, K., & Kang, S. (2021). Ghostpose: Multi-view pose estimation of transparent objects for robot hand grasping. In *IROS*, (pp. 5749–5755).

Chen, D., Li, J., Wang, Z., & Xu, K. (2020a). Learning canonical shape space for category-level 6d object pose and size estimation. In *CVPR*, (pp. 11973–11982).

Chen, H., Wang, P., Wang, F., Tian, W., Xiong, L., & Li, H. (2022a) Epro-pnp: Generalized end-to-end probabilistic perspective-n-points for monocular object pose estimation. In *CVPR*, (pp. 2781–2790).

Chen, H., Manhardt, F., Navab, N., & Busam, B. (2023a). Texpose: Neural texture learning for self-supervised 6d object pose estimation. In *CVPR*, (pp. 4841–4852).

Chen, J., Sun, M., Bao, T., Zhao, R., Wu, L., & He, Z. (2023b). *Zeropose: Cad-model-based zero-shot pose estimation*. arXiv:2305.17934.

Chen, K., & Dou, Q. (2021). Sgpa: Structure-guided prior adaptation for category-level 6d object pose estimation. In *ICCV*, (pp. 2773–2782).

Chen, K., Cao, R., James, S., Li, Y., Liu, Y.-H., Abbeel, P., & Dou, Q. (2022b). Sim-to-real 6d object pose estimation via iterative self-training for robotic bin picking. In *ECCV*, (pp. 533–550).

Chen, K., James, S., Sui, C., Liu, Y.-H., Abbeel, P., & Dou, Q. (2023c). Stereopose: Category-level 6d transparent object pose estimation from stereo images via back-view nocs. In: ICRA, pp 2855–2861

Chen, L., Yang, H., Wu, C., & Wu, S. (2022). Mp6d: An rgb-d dataset for metal parts' 6d pose estimation. *IEEE RAL, 7*(3), 5912–5919.

Chen, W., Jia, X., Chang, H. J., Duan, J., & Leonardis, A. (2020b). G2l-net: Global to local network for real-time 6d pose estimation with embedding vector features. In *CVPR*, (pp. 4233–4242)

Chen, W., Jia, X., Chang, H. J., Duan, J., & Leonardis, A. (2021). Fs-net: Fast shape-based network for category-level 6d object pose estimation with decoupled rotation mechanism. In *CVPR*, (pp. 1581–1590).

Chen, X., Dong, Z., Song, J., Geiger, A., & Hilliges, O. (2020c). Category level object pose estimation via neural analysis-by-synthesis. In *ECCV*, (pp. 139–156).

Chen, X., Zhang, H., Yu, Z., Opipari, A., & Chadwicke Jenkins, O. (2022d). Clearpose: Large-scale transparent object dataset and benchmark. In *ECCV*, (pp. 381–396).

Chen, Y., Di, Y., Zhai, G., Manhardt, F., Zhang, C., Zhang, R., Tombari, F., Navab, N., & Busam, B. (2024). Secondpose: Se (3)-consistent dual-stream feature fusion for category-level pose estimation. In *CVPR*, (pp. 9959–9969).

Chi, C., & Song, S. (2021). Garmentnets: Category-level pose estimation for garments via canonical space shape completion. In *ICCV*, (pp. 3324–3333).

Choi, C., & Christensen, H. I. (2012). 3d pose estimation of daily objects using an rgb-d camera. In *IROS*, (pp. 3342–3349).

Choi, C., Trevor, A. J., & Christensen, H. I. (2013). Rgb-d edge detection and edge-based registration. In *IROS*, (pp. 1568–1575).

Corona, E., Kundu, K., & Fidler, S. (2018). Pose estimation for objects with rotational symmetry. In *IROS*, (pp. 7215–7222).

Corsetti, J., Boscaini, D., Oh, C., Cavallaro, A., & Poiesi, F. (2024). Open-vocabulary object 6d pose estimation. In *CVPR*, (pp. 18071–18080).

Crivellaro, A., Rad, M., Verdie, Y., Yi, K. M., Fua, P., & Lepetit, V. (2017). Robust 3d object tracking from monocular images using stable parts. *IEEE TPAMI, 40*(6), 1465–1479.

Dang, Z., Wang, L., Guo, Y., & Salzmann, M. (2024). Match normalization: Learning-based point cloud registration for 6d object pose estimation in the real world. *IEEE TPAMI, 46*(06), 4489–4503.

Deng, X., Mousavian, A., Xiang, Y., Xia, F., Bretl, T., & Fox, D. (2021). Poserbpf: A rao-blackwellized particle filter for 6-d object pose tracking. *IEEE TRO, 37*(5), 1328–1342.

Deng, X., Geng, J., Bretl, T., Xiang, Y., & Fox, D. (2022). icaps: Iterative category-level object pose and shape estimation. *IEEE RAL, 7*(2), 1784–1791.

Di, Y., Manhardt, F., Wang, G., Ji, X., Navab, N., & Tombari, F. (2021). So-pose: Exploiting self-occlusion for direct 6d pose estimation. In *ICCV*, (pp. 12396–12405).

Di, Y., Zhang, R., Lou, Z., Manhardt, F., Ji, X., Navab, N., & Tombari, F. (2022). Gpv-pose: Category-level object pose estimation via geometry-guided point-wise voting. In *CVPR*, (pp. 6781–6791).

Di Felice, F., Remus, A., Gasperini, S., Busam, B., Ott, L., Tombari, F., Siegwart, R., & Avizzano, C. A. (2024). *Zero123-6d: Zero-shot novel view synthesis for rgb category-level 6d pose estimation*. arXiv:2403.14279.

Do, T. T., Cai, M., Pham, T., & Reid, I. (2018). *Deep-6dpose: Recovering 6d object pose from a single rgb image*. arXiv:1802.10367.

Dong, Z., Liu, S., Zhou, T., Cheng, H., Zeng, L., Yu, X., & Liu, H. (2019). Ppr-net: Point-wise pose regression network for instance segmentation and 6dd pose estimation in bin-picking scenarios. In *IROS*, (pp. 1773–1780).

Doosti, B., Naha, S., Mirbagheri, M., & Crandall D. J. (2020). Hope-net: A graph-based model for hand-object pose estimation. In *CVPR*, (pp. 6608–6617).

Dosovitskiy, A., Beyer, L., Kolesnikov, A., Weissenborn, D., Zhai, X., Unterthiner, T., Dehghani, M., Minderer, M., Heigold, G., Gelly, S., & others (2021) An image is worth 16x16 words: Transformers for image recognition at scale. In *ICLR*.

Doucet, A., De Freitas, N., Gordon, N. J. & others. (2001). Sequential Monte Carlo Methods in Practice, vol 1, chap Rao-blackwellised particle filtering for dynamic bayesian networks.

Doumanoglou, A., Kouskouridas, R., Malassiotis, S., & Kim, T.-K. (2016). Recovering 6d object pose and predicting next-best-view in the crowd. In *CVPR*, (pp. 3583–3592).

Drost, B., Ulrich, M., Navab, N., & Ilic, S. (2010). Model globally, match locally: Efficient and robust 3d object recognition. In *CVPR*, (pp. 998–1005).

Drost, B., Ulrich, M., Bergmann, P., Hartinger, P., & Steger, C. (2017). Introducing mvtec itodd - a dataset for 3d object recognition in industry. In *ICCVW*, (pp. 2200–2208).

Du, G., & Wang, K. (2021). Vision-based robotic grasping from object localization, object pose estimation to grasp estimation for parallel grippers: A review. *Artif Intell Rev, 55*(4), 1–40.

Duffhauss, F., Demmler, T., & Neumann, G. (2022). Mv6d: Multi-view 6d pose estimation on rgb-d frames using a deep point-wise voting network. In *IROS*, (pp. 3568–3575).

Elman, J. L. (1990). Finding structure in time. *Cogn Sci, 14*(2), 179–211.

Fan, Z., Song, Z., Xu, J., Wang, Z., Wu, K., Liu, H., & He, J. (2022a). Object level depth reconstruction for category level 6d object pose estimation from monocular rgb image. In *ECCV*, (pp. 220–236).

Fan, Z., Zhu, Y., He, Y., Sun, Q., Liu, H., & He, J. (2022). Deep learning on monocular object pose detection and tracking: A comprehensive overview. *ACM Comput Surv, 55*(4), 1–40.

Fan, Z., Pan, P., Wang, P., Jiang, Y., Xu, D., & Wang, Z. (2024a). Pope: 6-dof promptable pose estimation of any object, in any scene, with one reference. *CVPR*, (pp. 7771–7781).

Fan, Z., Song, Z., Wang, Z., Xu, J., Wu, K., Liu, H., & He, J. (2024b). Acr-pose: Adversarial canonical representation reconstruction network for category level 6d object pose estimation. In *ICMR*, (pp. 55–63).

Feng, G., Xu, T. B., Liu, F., Liu, M., & Wei, Z. (2024). Nvr-net: Normal vector guided regression network for disentangled 6d pose estimation. *IEEE TCSVT, 34*(2), 1098–1113.

Fischler, M. A., & Bolles, R. C. (1981). Random sample consensus: a paradigm for model fitting with applications to image analysis and automated cartography. *COMMUN ACM, 24*(6), 381–395.

Fu, B., Leong, S. K., Lian, X., & Ji, X. (2022). 6d robotic assembly based on rgb-only object pose estimation. In *IROS*, (pp. 4736–4742).

Gao, G., Lauri, M., Wang, Y., Hu, X., Zhang, J., & Frintrop, S. (2020). 6d object pose regression via supervised learning on point clouds. In *ICRA*, (pp. 3643–3649).

Gao, G., Lauri, M., Hu, X., Zhang, J., & Frintrop, S. (2021). Cloudaae: Learning 6d object pose regression with on-line data synthesis on point clouds. In *ICRA*, (pp. 11081–11087)

Gao, N., Ngo, V. A., Ziesche, H., & Neumann, G. (2023). Sa6d: Self-adaptive few-shot 6d pose estimator for novel and occluded objects. In *CoRL*.

Garon, M., & Lalonde, J. F. (2017). Deep 6-dof tracking. *IEEE TVCG, 23*(11), 2410–2418.

Ge R, Loianno G (2021) Vipose: Real-time visual-inertial 6d object pose tracking. In *IROS*, (pp. 4597–4603).

Georgakis, G., Karanam, S., Wu, Z., & Kosecka, J. (2019). Learning local rgb-to-cad correspondences for object pose estimation. In *ICCV*, (pp. 8967–8976).

Goodwin, W., Vaze, S, Havoutis I, & Posner, I. (2022). Zero-shot category-level object pose estimation. In *ECCV*, (pp. 516–532).

Gou, M., Pan, H., Fang, H. S., Liu, Z., Lu, C., & Tan, P. (2022). *Unseen object 6d pose estimation: A benchmark and baselines*. arXiv:2206.11808.

Guan, J., Hao, Y., Wu, Q., Li, S., & Fang, Y. (2024). A survey of 6dof object pose estimation methods for different application scenarios. *Sensors, 24*(4), 1076.

Guo, J., Zhong, F., Xiong, R., Liu, Y., Wang, Y., & Liao, Y. (2022). A visual navigation perspective for category-level object pose estimation. In *ECCV*, (pp. 123–141).

Guo, S., Hu, Y, Alvarez, J. M., & Salzmann, M. (2023). Knowledge distillation for 6d pose estimation by aligning distributions of local predictions. In *CVPR*, (pp. 18633–186420.

Hagelskjær, F., & Haugaard, R. L. (2023). Keymatchnet: Zero-shot pose estimation in 3d point clouds by generalized keypoint matching. arXiv:2303.16102.

Hai, Y., Song, R., Li, J., Ferstl, D., & Hu, Y. (2023a). Pseudo flow consistency for self-supervised 6d object pose estimation. In *ICCV*, (pp. 14075–14085).

Hai, Y., Song, R., Li, J., & Hu, Y. (2023b). Shape-constraint recurrent flow for 6d object pose estimation. In *CVPR*, (pp. 4831–4840).

Haugaard, R. L., & Buch, A. G. (2022). Surfemb: Dense and continuous correspondence distributions for object pose estimation with learnt surface embeddings. In *CVPR*, (pp. 6749–6758).

He, K., Gkioxari, G., Dollár, P., & Girshick, R. (2017). Mask r-cnn. In *ICCV*, (pp. 2961–2969).

He, X., Sun, J., Wang, Y., Huang, D., Bao, H., Zhou, X. (2022a). Onepose++: Keypoint-free one-shot object pose estimation without cad models. In *NeurIPS*, (pp. 35103–35115).

He, Y., Sun, W., Huang, H., Liu, J., Fan, H., & Sun, J. (2020). Pvn3d: A deep point-wise 3d keypoints voting network for 6dof pose estimation. In *CVPR*, (pp. 11632–11641).

He, Y., Huang, H., Fan, H., Chen, Q., & Sun, J. (2021). Ffb6d: A full flow bidirectional fusion network for 6d pose estimation. In *CVPR*, (pp. 3003–3013).

He, Y., Fan, H., Huang, H., Chen, Q., & Sun, J. (2022b). Towards self-supervised category-level object pose and size estimation. arXiv:2203.02884.

He, Y., Wang, Y., Fan, H., Sun, J., & Chen, Q. (2022c) Fs6d: Few-shot 6d pose estimation of novel objects. In *CVPR*, (pp. 6814–6824).

Hinterstoisser, S., Lepetit, V., Ilic, S., Holzer, S., Bradski, G., Konolige, K., & Navab, N. (2012). Model based training, detection and pose estimation of texture-less 3d objects in heavily cluttered scenes. In *ACCV*, (pp. 548–562).

Hodan, T., Haluza, P., Obdržálek, Š., Matas, J., Lourakis, M., & Zabulis, X. (2017). T-less: An rgb-d dataset for 6d pose estimation of texture-less objects. In *WACV*, (pp. 880–888).

Hodan, T., Barath, D., & Matas, J. (2020). Epos: Estimating 6d pose of objects with symmetries. In *CVPR*, (pp. 11703–11712).

Hodan, T., Sundermeyer, M., Labbe, Y., Van Nguyen, N., Wang, G., Brachmann, E., Drost, B., Lepetit, V., Rother, C., & Matas, J.

(2024). Bop challenge 2023 on detection, segmentation and pose estimation of seen and unseen rigid objects. (pp. 5610–5619).

Hong, J. X., Zhang, H. B., Liu, J. H., Lei, Q., Yang, L.-J., & Du, J.-X. (2024). A transformer-based multi-modal fusion network for 6d pose estimation. *Information Fusion, 105*, Article 102227.

Hoque, S., Arafat, M. Y., Xu, S., Maiti, A., & Wei, Y. (2021). A comprehensive review on 3d object detection and 6d pose estimation with deep learning. *IEEE Access, 9*, 143746–143770.

Hoque, S., Xu, S., Maiti, A., Wei, Y., & Arafat, Y. (2023). Deep learning for 6d pose estimation of objects-a case study for autonomous driving. *Expert Syst Appl, 223*, Article 119838.

Hsiao, T. C., Chen, H. W., Yang, H. K., & Lee, C.-Y. (2024). Confronting ambiguity in 6d object pose estimation via score-based diffusion on se(3). In *CVPR*, (pp. 352–362).

Hu, Y., Hugonot, J., Fua, P., & Salzmann, M. (2019). Segmentation-driven 6d object pose estimation. In *CVPR*, (pp. 3385–3394)

Hu, Y., Fua, P., Wang, W., & Salzmann, M. (2020). Single-stage 6d object pose estimation. In *CVPR*, (pp. 2930–2939).

Hu, Y., Speierer, S., Jakob, W., Fua, P., & Salzmann, M. (2021). Wide-depth-range 6d object pose estimation in space. In *CVPR*, (pp. 15870–15879).

Hu, Y., Fua, P., Salzmann, M. (2022). Perspective flow aggregation for data-limited 6d object pose estimation. In *ECCV*, (pp. 89–106).

Huang, J., Yu, H., Yu, K. T., Navab, N., Ilic, S., & Busam, B. (2024). Matchu: Matching unseen objects for 6d pose estimation from rgb-d images. In *CVPR*, (pp. 10095–10105).

Huang, L., Hodan, T., Ma, L., Zhang, L., Tran, L., Twigg, C., Wu, P.-C., Yuan, J., Keskin, C., & Wang, R. (2022) Neural correspondence field for object pose estimation. In *ECCV*, (pp. 585–603).

Huang, W. L., Hung, C. Y., & Lin, I. C. (2021). Confidence-based 6d object pose estimation. *IEEE TMM, 24*, 3025–3035.

Ikeda, T., Tanishige, S., Amma, A., Sudano, M., Audren, H., & Nishiwaki, K. (2022). Sim2real instance-level style transfer for 6d pose estimation. In *IROS*, (pp. 3225–3232).

Irshad, M. Z., Kollar, T., Laskey, M., Stone, K., & Kira, Z. (2022). Centersnap: Single-shot multi-object 3d shape reconstruction and categorical 6d pose and size estimation. In *ICRA*, (pp. 10632–10640).

Iwase, S., Liu, X., Khirodkar, R., Yokota, R., & Kitani, K. M. (2021) Repose: Fast 6d object pose refinement via deep texture rendering. In *ICCV*, (pp. 3303–3312).

Jiang, H., Dang, Z., Gu, S., Xie, J., Salzmann, M., & Yang, J. (2023). Center-based decoupled point cloud registration for 6d object pose estimation. In *ICCV*, (pp. 3427–3437).

Jiang, H., Salzmann, M., Dang, Z., Xie, J., & Yang, J. (2024) Se(3) diffusion model-based point cloud registration for robust 6d object pose estimation. In *NeurIPS*, (pp. 21285–21297).

Jiang, X., Li, D., Chen, H., Zheng, Y., Zhao, R., & Wu, L. (2022). Uni6d: A unified cnn framework without projection breakdown for 6d pose estimation. In *CVPR*, (pp. 11174–11184).

Josifovski, J., Kerzel, M., Pregizer, C., Posniak, L., & Wermter, S. (2018). Object detection and pose estimation based on convolutional neural networks trained with synthetic data. In *IROS*, (pp. 6269–6276)

Jung, H., Wu, S. C., Ruhkamp, P., Zhai, G., Schieber, H., Rizzoli, G., Wang, P., Zhao, H., Garattoni, L., Meier, S., & and others (2024) Housecat6d–a large-scale multi-modal category level 6d object pose dataset with household objects in realistic scenarios. In *CVPR*, (pp. 22498–22508).

Kabsch, W. (1976). A solution for the best rotation to relate two sets of vectors. *Acta Crystallographica Section A: Crystal Physics, Diffraction, Theoretical and General Crystallography, 32*(5), 922–923.

Kaskman, R., Zakharov, S., Shugurov, I., & Ilic, S. (2019). Homebreweddb: Rgb-d dataset for 6d pose estimation of 3d objects. In *ICCVW*, (pp. 2767–2776).

Kehl, W., Manhardt, F., Tombari, F., Ilic, S., & Navab, N. (2017). Ssd-6d: Making rgb-based 3d detection and 6d pose estimation great again. In *ICCV*, (pp. 1521–1529).

Kim, J., Pyo, H., Jang, I., Kang, J., Ju, B., & Ko, K. (2022). Tomato harvesting robotic system based on deep-tomatos: Deep learning network using transformation loss for 6d pose estimation of maturity classified tomatoes with side-stem. *Comput Electron Agric, 201*, Article 107300.

Kipf, T. N., & Welling, M. (2016). Semi-supervised classification with graph convolutional networks. arXiv:1609.02907.

Kirillov, A., Mintun, E., Ravi, N., Mao, H., Rolland, C., Gustafson, L., Xiao, T., Whitehead, S., Berg, A. C., Lo, W.-Y., Dollár, P., & Girshick, R. (2023). Segment anything. In *ICCV*, (pp. 4015–4026).

Kleeberger, K., & Huber, M. F. (2020). Single shot 6d object pose estimation. In *ICRA*, (pp. 6239–6245).

Labbé, Y., Carpentier, J., Aubry, M., & Sivic, J. (2020). Cosypose: Consistent multi-view multi-object 6d pose estimation. In *ECCV*, (pp. 574–591).

Labbé, Y., Manuelli, L., Mousavian, A., Tyree, S., Birchfield, S., Tremblay, J., Carpentier, J., Aubry, M., Fox, D., & Sivic, J. (2022). Megapose: 6d pose estimation of novel objects via render & compare. In *CoRL*, (pp. 715–725).

LeCun, Y., Boser, B., Denker, J. S., Henderson, D., Howard, R. E., Hubbard, W., & Jackel, L. D. (1989). Backpropagation applied to handwritten zip code recognition. *Neural Comput, 1*(4), 541–551.

Lee, J., Cabon, Y., Brégier, R., Yoo, S., & Revaud, J. (2024). Mfos: Model-free & one-shot object pose estimation. In *AAAI*, (pp. 2911–2919).

Lee, T., Lee, B. U., Kim, M., & Kweon, I. S. (2021). Category-level metric scale object shape and pose estimation. *IEEE RAL, 6*(4), 8575–8582.

Lee, T., Lee, B. U., Shin, I., Choe, J., Shin, U., Kweon, I. S., & Yoon, K.-J. (2022). Uda-cope: Unsupervised domain adaptation for category-level object pose estimation. In *CVPR*, (pp. 14891–14900).

Lee, T., Tremblay, J., Blukis, V., Wen, B., Lee, B.-U., Shin, I., Birchfield, S., Kweon, I. S., & Yoon, K.-J. (2023). Tta-cope: Test-time adaptation for category-level object pose estimation. In *CVPR*, (pp. 21285–21295).

Legrand, A., Detry, R., & De Vleeschouwer, C. (2024). Domain generalization for 6d pose estimation through nerf-based image synthesis. arXiv:2407.10762.

Li, C., Bai, J., & Hager, G. D. (2018a). A unified framework for multi-view multi-class object pose estimation. In *ECCV*, (pp. 254–269).

Li, F., Vutukur, S. R., Yu, H., Shugurov, I., Busam, B., Yang, S., & Ilic, S. (2023a). Nerf-pose: A first-reconstruct-then-regress approach for weakly-supervised 6d object pose estimation. In *ICCV*, (pp. 2123–2133).

Li, G., Li, Y., Ye, Z., Zhang, Q., Kong, T., Cui, Z., & Zhang, G. (2023b). Generative category-level shape and pose estimation with semantic primitives. In *CoRL*, (pp. 1390–1400).

Li, G., Zhu, D., Zhang, G., Shi, W., Zhang, T., Zhang, X., & Li, J. (2023c). Sd-pose: Structural discrepancy aware category-level 6d object pose estimation. In *WACV*, (pp. 5685–5694).

Li, H., Lin, J., & Jia, K. (2022). Dcl-net: Deep correspondence learning network for 6d pose estimation. In *ECCV*, (pp. 369–385).

Li, X., Wang, H., Yi, L., Guibas, L. J., Abbott, A. L., & Song, S. (2020). Category-level articulated object pose estimation. In *CVPR*, (pp. 3706–3715).

Li, X., Weng, Y., Yi, L., Guibas, L. J., Abbott, A., Song, S., & Wang, H. (2021a). Leveraging se (3) equivariance for self-supervised category-level object pose estimation from point clouds. In *NeurIPS*, (pp. 15370–15381).

Li, Y., Sun, J., Li, X., Zhang, Z., Cheng, H., & Wang, X. (2018b). Weakly supervised 6d pose estimation for robotic grasping. In *SIGGRAPH*, (pp. 1–8).

Li, Y., Wang, G., Ji, X., Xiang, Y., & Fox, D. (2018c). Deepim: Deep iterative matching for 6d pose estimation. In *ECCV*, (pp. 683–698).

Li, Y., Mao, Y., Bala, R., & Hadap, S. (2024a). Mrc-net: 6-dof pose estimation with multiscale residual correlation. In *CVPR*, (pp. 10476–10486).

Li, Y., Mo, K., Duan, Y., Wang, H., Zhang, J., & Shao, L. (2024b). Category-level multi-part multi-joint 3d shape assembly. In *CVPR*, (pp. 3281–3291).

Li, Z., & Ji, X. (2020). Pose-guided auto-encoder and feature-based refinement for 6-dof object pose regression. In *ICRA*, (pp. 8397–8403).

Li, Z., Wang, G., & Ji, X. (2019). Cdpn: Coordinates-based disentangled pose network for real-time rgb-based 6-dof object pose estimation. In *ICCV*, (pp. 7678–7687).

Li, Z., Hu, Y., Salzmann, M., & Ji, X. (2021b). Sd-pose: Semantic decomposition for cross-domain 6d object pose estimation. In *AAAI*, (pp. 2020–2028).

Lian, R., & Ling, H. (2023). Checkerpose: Progressive dense keypoint localization for object pose estimation with graph neural network. In *ICCV*, (pp. 14022–14033).

Lin, H., Liu, Z., Cheang, C., Fu, Y., Guo, G., & Xue, X. (2022a). Sar-net: Shape alignment and recovery network for category-level 6d object pose and size estimation. In *CVPR*, (pp. 6707–6717).

Lin, H., Peng, S., Zhou, Z., & Zhou, X. (2022b). Learning to estimate object poses without real image annotations. In *IJCAI*, (pp. 1159–1165).

Lin, J., Li, H., Chen, K., Lu, J., & Jia, K. (2021a). Sparse steerable convolutions: An efficient learning of se (3)-equivariant features for estimation and tracking of object poses in 3d space. In *NeurIPS*, (pp. 16779–16790).

Lin, J., Wei, Z., Li, Z., Xu, S., Jia, K., & Li, Y. (2021b). Dualposenet: Category-level 6d object pose and size estimation using dual pose network with refined learning of pose consistency. In *ICCV*, (pp. 3560–3569).

Lin, J., Wei, Z., Ding, C., & Jia, K. (2022c). Category-level 6d object pose and size estimation using self-supervised deep prior deformation networks. In *ECCV*, (pp. 19–34).

Lin, J., Wei, Z., Zhang, Y., & Jia, K. (2023a). Vi-net: Boosting category-level 6d object pose estimation via learning decoupled rotations on the spherical representations. In *ICCV*, (pp. 14001–14011).

Lin J, Liu L, Lu D, et al (2024a) Sam-6d: Segment anything model meets zero-shot 6d object pose estimation. In: CVPR, pp 27906–27916

Lin M, Murali V, Karaman S (2021c) 6d object pose estimation with pairwise compatible geometric features. In: ICRA, pp 10966–10973

Lin, X., Yang, W., Gao, Y., & Zhang, T. (2024b). Instance-adaptive and geometric-aware keypoint learning for category-level 6d object pose estimation. In *CVPR*, (pp. 21040–21049).

Lin, X., Zhu, M., Dang, R., Zhou, G., Shu, S., Lin, F., Liu, C., & Chen, Q. (2024c). Clipose: Category-level object pose estimation with pre-trained vision-language knowledge. *IEEE TCSVT*.

Lin, Y., Tremblay, J., Tyree, S., Vela, P. A., & Birchfield, S. (2022d). Keypoint-based category-level object pose tracking from an rgb sequence with uncertainty estimation. In *ICRA*, (pp. 1258–1264).

Lin, Y., Tremblay, J., Tyree, S., Vela, P. A., & Birchfield, S. (2022e). Single-stage keypoint-based category-level object pose estimation from an rgb image. In *ICRA*, (pp. 1547–1553).

Lin, Y., Su, Y., Nathan, P., Inuganti, S., Di, Y., Sundermeyer, M., Manhardt, F., Stricker, D., Rambach, J., & Zhang, Y. (2024d). Hipose: Hierarchical binary surface encoding and correspondence pruning for rgb-d 6dof object pose estimation. In *CVPR*, (pp. 10148–10158).

Lin, Z., Ding, C., Yao, H., Kuang, Z., & Huang, S. (2023b). Harmonious feature learning for interactive hand-object pose estimation. In *CVPR*, (pp. 12989–12998).

Lipson, L., Teed, Z., Goyal, A., & Deng, J. (2022). Coupled iterative refinement for 6d multi-object pose estimation. In *CVPR*, (pp. 6728–6737).

Liu, C., Sun, W., Liu, J., Zhang, X., Fan, S., & Fu, Q. (2023). Fine segmentation and difference-aware shape adjustment for category-level 6dof object pose estimation. *Appl Intell, 53*(20), 23711–23728.

Liu, F., Hu, Y., & Salzmann, M. (2023b). Linear-covariance loss for end-to-end learning of 6d pose estimation. In *ICCV*, (pp. 14107–14117).

Liu, J., Cao, Z., Tang, Y., Liu, X., & Tan, M. (2022). Category-level 6d object pose estimation with structure encoder and reasoning attention. *IEEE TCSVT, 32*(10), 6728–6740.

Liu, J., Sun, W., Liu, C., Zhang, X., Fan, S., & Wu, W. (2022). Hff6d: Hierarchical feature fusion network for robust 6d object pose tracking. *IEEE TCSVT, 32*(11), 7719–7731.

Liu, J., Chen, Y., Ye, X., & Qi, X. (2023c). Ist-net: Prior-free category-level pose estimation with implicit space transformation. In *ICCV*, (pp. 13978–13988).

Liu, J., Sun, W., Liu, C., Zhang, X., & Fu, Q. (2023). Robotic continuous grasping system by shape transformer-guided multi-object category-level 6d pose estimation. *IEEE TII, 19*(11), 11171–11181.

Liu, J., Sun, W., Yang, H., Liu, C., Zhang, X., & Mian, A. (2024). Domain-generalized robotic picking via contrastive learning-based 6-d pose estimation. *IEEE TII, 20*(6), 8650–8661.

Liu, J., Sun, W., Liu, C., Yang, H., Zhang, X., & Mian, A. (2025). Mh6d: Multi-hypothesis consistency learning for category-level 6-d object pose estimation. *IEEE TNNLS, 36*(3), 4820–4833.

Liu, J., Sun, W., Yang, H., Zheng, J., Geng, Z., Rahmani, H., & Mian, A. (2025). Diff9d: Diffusion-based domain-generalized category-level 9-dof object pose estimation. *IEEE Transactions on Pattern Analysis and Machine Intelligence, 47*(7), 5520–5537.

Liu, J., Sun, W., Yang, H., Zheng, J., Geng, Z., Rahmani, H., & Mian, A. (2025c). Monodiff9d: Monocular category-level 9d object pose estimation via diffusion model. In *ICRA*, (pp. 8687–8694).

Liu, L., Xue, H., Xu, W., Fu, H., & Lu, C. (2022). Toward real-world category-level articulation pose estimation. *IEEE TIP, 31*, 1072–1083.

Liu, L., Du, J., Wu, H., Yang, X., Liu, Z., Hong, R., & Wang, M. (2023e). Category-level articulated object 9d pose estimation via reinforcement learning. In *ACM MM*, (pp. 728–736).

Liu, P., Zhang, Q., Zhang, J., Wang, F., & Cheng, J. (2021a). Mfpn-6d: Real-time one-stage pose estimation of objects on rgb images. In *ICRA*, (pp. 12939–12945).

Liu, P., Zhang, Q., & Cheng, J. (2023). Bdr6d: Bidirectional deep residual fusion network for 6d pose estimation. *IEEE TASE, 21*(2), 1793–1804.

Liu, S., Jiang, H., Xu, J., Liu, S., & Wang, X. (2021b). Semi-supervised 3d hand-object poses estimation with interactions in time. In *CVPR*, (pp. 14687–14697).

Liu, W., Anguelov, D., Erhan, D., Szegedy, C., Reed, S., Fu, C.-Y., & Berg, A. C. (2016). Ssd: Single shot multibox detector. In *ECCV*, (pp. 21–37).

Liu, X., & Zhang, J. (2023). Self-supervised category-level articulated object pose estimation with part-level se (3) equivariance. *ICLR*.

Liu, X., Zhang, J., He, X., Song, X., & Qin, X. (2019a). 6dof pose estimation with object cutout based on a deep autoencoder. In *ISMAR-Adjunct*, (pp. 360–365).

Liu, X., Jonschkowski, R., Angelova, A., & Konolige, K. (2020). Keypose: Multi-view 3d labeling and keypoint estimation for transparent objects. In *CVPR*, (pp. 11602–11610).

Liu, X., Iwase, S., & Kitani, K. M. (2021c). Kdfnet: Learning keypoint distance field for 6d object pose estimation. In *IROS*, (pp. 4631–4638).

Liu, X., Wang, G., Li, Y., & Ji, X. (2022d) Catre: Iterative point clouds alignment for category-level object pose refinement. In *ECCV*, (pp. 499–516).

Liu, X., Yuan, X., Zhu, Q., Wang, Y., Feng, M., Zhou, J., & Zhou, Z. (2024). A depth adaptive feature extraction and dense prediction network for 6-d pose estimation in robotic grasping. *IEEE TII, 20*(2), 2727–2737.

Liu, Y., Zhou, L., Zong, H., Gong, X., Wu, Q., Liang, Q., & Wang, J. (2019). Regression-based three-dimensional pose estimation for texture-less objects. *IEEE TMM, 21*(11), 2776–2789.

Liu, Y., Liu, Y., Jiang, C., Lyu, K., Wan, W., Shen, H., Liang, B., Fu, Z., Wang, H., & Yi, L. (2022e). Hoi4d: A 4d egocentric dataset for category-level human-object interaction. In *CVPR*, (pp. 21013–21022).

Liu, Y., Wen, Y., Peng, S., Lin, C., Long, X., Komura, T., & Wang, W. (2022f). Gen6d: Generalizable model-free 6-dof object pose estimation from rgb images. In *ECCV*, (pp. 298–315).

Liu, Z., Wang, Q., Liu, D., & Tan, J. (2024c). Pa-pose: Partial point cloud fusion based on reliable alignment for 6d pose tracking. PR 148:110151

Long, J., Shelhamer, E., & Darrell, T. (2015). Fully convolutional networks for semantic segmentation. In *CVPR*, (pp. 3431–3440).

Lowe, D. G. (2004). Distinctive image features from scale-invariant keypoints. *IJCV, 60*, 91–110.

Ma, W., Wang, A., Yuille, A., & Kortylewski, A. (2022). Robust category-level 6d pose estimation with coarse-to-fine rendering of neural features. In *ECCV*, (pp. 492–508).

Manhardt, F., Kehl, W., Navab, N., & Tombari, F. (2018). Deep model-based 6d pose refinement in rgb. In *ECCV*, (pp. 800–815)

Manhardt, F., Arroyo, D. M., Rupprecht, C., Busam, B., Birdal, T., Navab, N., & Tombari, F. (2019). Explaining the ambiguity of object detection and 6d pose from visual data. In *ICCV*, (pp. 6841–6850).

Manuelli, L., Gao, W., Florence, P., & Tedrake, R. (2019). kpam: Key-point affordances for category-level robotic manipulation. In *ISRR*, (pp. 132–157).

Martín-Martín, R., Eppner, C., & Brock, O. (2019). The rbo dataset of articulated objects and interactions. *IJRR, 38*(9), 1013–1019.

Marullo, G., Tanzi, L., Piazzolla, P., & Vezzetti, E. (2023). 6d object position estimation from 2d images: A literature review. *Multimed Tools Appl, 82*(16), 24605–24643.

Mei, J., Jiang, X., & Ding, H. (2022). Spatial feature mapping for 6dof object pose estimation. *PR* 131:108835

Michel, F., Krull, A., Brachmann, E., Yang, M. Y., Gumhold, S., & Rother, C. (2015) Pose estimation of kinematic chain instances via object coordinate regression. In *BMVC*, (pp. 181–1).

Mo, K., Zhu, S., Chang, A. X., Yi, L., Tripathi, S., Guibas, L. J., & Su, H. (2019). Partnet: A large-scale benchmark for fine-grained and hierarchical part-level 3d object understanding. In *CVPR*, (pp. 909–918).

Mo, N., Gan, W., Yokoya, N., & Chen, S. (2022). Es6d: A computation efficient and symmetry-aware 6d pose regression framework. In *CVPR*, (pp. 6718–6727).

Moon, S., Son, H., Hur, D., & Kim, S. (2024). Genflow: Generalizable recurrent flow for 6d pose refinement of novel objects. In *CVPR*, (pp. 10039–10049).

Moré, J. J. (2006). The levenberg-marquardt algorithm: Implementation and theory. In *Numerical Analysis*, (pp. 105–116).

Mu, F., Huang, R., Luo, A., Li, X., Qiu, J., & Cheng, H. (2021). Temporalfusion: Temporal motion reasoning with multi-frame fusion for 6d object pose estimation. In *IROS*, (pp. 5930–5936).

Nguyen, V. N., Du, Y., Xiao, Y., Ramamonjisoa, M., & Lepetit, V. (2022a). Pizza: A powerful image-only zero-shot zero-cad approach to 6 dof tracking. In *3DV*, (pp. 515–525).

Nguyen, V. N., Hu, Y., Xiao, Y., Salzmann, M., & Lepetit, V. (2022b). Templates for 3d object pose estimation revisited: Generalization to new objects and robustness to occlusions. In *CVPR*, (pp. 6771–6780).

Nguyen, V. N., Groueix, T., Ponimatkin, G., Lepetit, V., & Hodan, T. (2023). Cnos: A strong baseline for cad-based novel object segmentation. In *ICCV*, (pp. 2134–2140).

Nguyen, V. N., Groueix, T., Ponimatkin, G., Hu, Y., Marlet, R., Salzmann, M., & Lepetit, V. (2024a). Nope: Novel object pose estimation from a single image. In *CVPR*, (pp. 17923–17932).

Nguyen, V. N., Groueix, T., Salzmann, M., & Lepetit, V. (2024b). Gigapose: Fast and robust novel object pose estimation via one correspondence. In *CVPR*, (pp. 9903–9913).

Nie, T., Ma, J., Zhao, Y., Fan, Z., Wen, J., & Sun, M. (2023). Category-level 6d pose estimation using geometry-guided instance-aware prior and multi-stage reconstruction. *IEEE RAL, 8*(4), 2381–2388.

Oberweger, M., Rad, M., & Lepetit, V. (2018). Making deep heatmaps robust to partial occlusions for 3d object pose estimation. In *ECCV*, (pp. 119–134).

Okorn, B., Gu, Q., Hebert, M., & Held, D. (2021). Zephyr: Zero-shot pose hypothesis rating. In *ICRA*, (pp. 14141–14148).

Ono, Y., Trulls, E., Fua, P., & Yi, K. M. (2018). Lf-net: Learning local features from images. In *NeurIPS*, (pp. 6237–6247).

Oquab, M., Darcet, T., Moutakanni, T., Vo H., Szafraniec M., Khalidov, V., Fernandez, P., Haziza, D., Massa, F., El-Nouby, A., & others (2023) Dinov2: Learning robust visual features without supervision. arXiv:2304.07193.

Örnek, E. P., Labbé, Y., Tekin, B., Ma, L., Keskin, C., Forster, C., & Hodan, T. (2023). Foundpose: Unseen object pose estimation with foundation features. arXiv:2311.18809.

Pan, P., Fan, Z., Feng, B. Y., Wang, P., Li, C., & Wang, Z. (2024). Learning to estimate 6dof pose from limited data: A few-shot, generalizable approach using rgb images. In *3DV*, (pp. 1059–1071).

Pandey, R., Pidlypenskyi, P., Yang, S., & Kaeser-Chen, C. (2018). Efficient 6-dof tracking of handheld objects from an egocentric viewpoint. In *ECCV*, (pp. 416–431).

Papaioannidis, C., & Pitas, I. (2019). 3d object pose estimation using multi-objective quaternion learning. *IEEE TCSVT, 30*(8), 2683–2693.

Papaioannidis, C., Mygdalis, V., & Pitas, I. (2020). Domain-translated 3d object pose estimation. *IEEE TIP, 29*, 9279–9291.

Park, J., & Cho, N. (2022). Dprost: Dynamic projective spatial transformer network for 6d pose estimation. In *ECCV*, (pp. 363–379).

Park, J. J., Florence, P., Straub, J., Newcombe, R., & Lovegrove, S. (2019a) Deepsdf: Learning continuous signed distance functions for shape representation. In *CVPR*, (pp. 165–174)

Park, K., Patten, T., & Vincze, M. (2019b). Pix2pose: Pixel-wise coordinate regression of objects for 6d pose estimation. In *ICCV*, (pp. 7668–7677).

Park, K., Mousavian, A., Xiang, Y., & Fox, D. (2020a). Latentfusion: End-to-end differentiable reconstruction and rendering for unseen object pose estimation. In *CVPR*, (pp. 10710–10719).

Park, K., Patten, T., & Vincze, M. (2020b). Neural object learning for 6d pose estimation using a few cluttered images. In *ECCV*, (pp. 656–673).

Patten, T., Park, K., Leitner, M., Wolfram, K., & Vincze, M. (2021). Object learning for 6d pose estimation and grasping from rgb-d videos of in-hand manipulation. In *IROS*, (pp. 4831–4838).

Pavlakos, G., Zhou, X., Chan, A., Derpanis, K. G., & Daniilidis, K. (2017). 6-dof object pose from semantic keypoints. In *ICRA*, (pp. 2011–2018).

Peng, S., Liu, Y., Huang, Q., Zhou, X., & Bao, H. (2019). Pvnet: Pixel-wise voting network for 6dof pose estimation. In *CVPR*, (pp. 4561–4570).

Peng, W., Yan, J., Wen, H., & Sun, Y. (2022) Self-supervised category-level 6d object pose estimation with deep implicit shape representation. In *AAAI*, (pp. 2082–2090).

Pitteri, G., Ilic, S., & Lepetit, V. (2019). Cornet: Generic 3d corners for 6d pose estimation of new objects without retraining. In *ICCVW*.

Pitteri, G., Bugeau, A., Ilic, S., & Lepetit, V. (2020). 3d object detection and pose estimation of unseen objects in color images with local surface embeddings. In *ACCV*, (pp. 38–54).

Proença, P. F., & Gao, Y. (2020). Deep learning for spacecraft pose estimation from photorealistic rendering. In *ICRA*, (pp. 6007–6013).

Qi, C. R., Su, H., Mo, K., & Guibas, L. J. (2017) Pointnet: Deep learning on point sets for 3d classification and segmentation. In *CVPR*, (pp. 652–660).

Qi, H., Zhao, C., Salzmann, M., & Mathis, A. (2024). Hoisdf: Constraining 3d hand-object pose estimation with global signed distance fields. In *CVPR*, (pp. 10392–10402).

Qin, Y., Su, H., & Wang, X. (2022). From one hand to multiple hands: Imitation learning for dexterous manipulation from single-camera teleoperation. *IEEE RAL, 7*(4), 10873–10881.

Qin, Z., Yu, H., Wang, C., Guo, Y., Peng, Y., & Xu, K. (2022b). Geometric transformer for fast and robust point cloud registration. In *CVPR*, (pp. 11143–11152)

Rad, M., & Lepetit, V. (2017), Bb8: A scalable, accurate, robust to partial occlusion method for predicting the 3d poses of challenging objects without using depth. In *ICCV*, (pp. 3828–3836).

Rambach, J., Deng, C., Pagani, A., & Stricker, D. (2018). Learning 6dof object poses from synthetic single channel images. In *ISMAR-Adjunct*, (pp. 164–169).

Redmon, J., Divvala, S., Girshick, R., & Farhadi, A. (2016). You only look once: Unified, real-time object detection. In *CVPR*, (pp. 779–788).

Remus, A., & D'Avella, S. (2023). I2c-net: Using instance-level neural networks for monocular category-level 6d pose estimation. *IEEE RAL, 8*(3), 1515–1522.

Rennie, C., Shome, R., Bekris, K. E., & De Souza, A. F. (2016). A dataset for improved rgbd-based object detection and pose estimation for warehouse pick-and-place. *IEEE RAL, 1*(2), 1179–1185.

Rezazadeh, A., Dikhale, S., Iba, S., & Jamali, N. (2023). Hierarchical graph nural networks for proprioceptive 6d pose estimation of in-hand objects. In *ICRA*, (pp. 2884–2890).

Rusu, R. B., Blodow, N., & Beetz, M. (2009). Fast point feature histograms (fpfh) for 3d registration. In *ICRA*, (pp. 3212–3217).

Rusu, R. B., Bradski, G., Thibaux, R., & Hsu J. (2010). Fast 3d recognition and pose using the viewpoint feature histogram. In *IROS*, (pp. 2155–2162).

Sahin, C., & Kim, T. K. (2018). Category-level 6d object pose recovery in depth images. In *ECCVW*.

Sarlin, P. E., DeTone, D., Malisiewicz, T., & Rabinovich, A. (2020). Superglue: Learning feature matching with graph neural networks. In *CVPR*, (pp. 4938–4947).

Sarode, V., Li, X., Goforth, H., Aoki, Y., Srivatsan, R. A., Lucey, S., & Choset, H. (2019). Pcrnet: Point cloud registration network using pointnet encoding. arXiv:1908.07906.

Shi, Y., Huang, J., Xu, X., Zhang, Y., & Xu, K. (2021). Stablepose: Learning 6d object poses from geometrically stable patches. In *CVPR*, (pp. 15222–15231).

Shotton, J., Glocker, B., Zach, C., Izadi, S., Criminisi, A., & Fitzgibbon, A. (2013). Scene coordinate regression forests for camera relocalization in rgb-d images. In *CVPR*, (pp. 2930–2937).

Shugurov, I., Zakharov, S., & Ilic, S. (2021). Dpodv2: Dense correspondence-based 6 dof pose estimation. *IEEE TPAMI, 44*(11), 7417–7435.

Shugurov, I., Li, F., Busam, B., & Ilic, S. (2022). Osop: A multi-stage one shot object pose estimation framework. In *CVPR*, (pp. 6835–6844).

Sock, J., Garcia-Hernando, G., Armagan, A., & Kim, T.-K. (2020). Introducing pose consistency and warp-alignment for self-supervised 6d object pose estimation in color images. In *3DV*, (pp. 291–300).

Song, C., Song, J., & Huang, Q. (2020). Hybridpose: 6d object pose estimation under hybrid representations. In *CVPR*, (pp. 431–440).

Stevšič, S., & Hilliges, O. (2020). Spatial attention improves iterative 6d object pose estimation. In *3DV*, (pp. 1070–1078).

Su, Y., Rambach, J., Minaskan, N., Lesur, P., Pagani, A., & Stricker, D. (2019). Deep multi-state object pose estimation for augmented reality assembly. In *ISMAR-Adjunct*, (pp. 222–227).

Su, Y., Saleh, M., Fetzer, T., Rambach, J., Navab, N., Busam, B., Stricker, D., & Tombari, F. (2022). Zebrapose: Coarse to fine surface encoding for 6dof object pose estimation. In *CVPR*, (pp. 6738–6748).

Sun, H., Ni, P., Li, Z., Wang, Y., Zhu, X., & Cao, Q. (2023). Panelpose: A 6d pose estimation of highly-variable panel object for robotic robust cockpit panel inspection. In *IROS*, (pp. 3214–3221).

Sun, J., Shen, Z., Wang, Y., Bao, H., & Zhou, X. (2021). Loftr: Detector-free local feature matching with transformers. In *CVPR*, (pp. 8922–8931).

Sun, J., Wang, Y., Feng, M., Wang, D., Zhao, J., Stachniss, C., & Chen, X. (2022a). Ick-track: A category-level 6-dof pose tracker using inter-frame consistent keypoints for aerial manipulation. In *IROS*, (pp. 1556–1563).

Sun, J., Wang, Z., Zhang, S., He, X., Zhao, H., Zhang, G., & Zhou, X. (2022b). Onepose: One-shot object pose estimation without cad models. In *CVPR*, (pp. 6825–6834).

Sundermeyer, M., Marton, Z. C., Durner, M., Brucker, M., & Triebel, R. (2018). Implicit 3d orientation learning for 6d object detection from rgb images. In *ECCV*, (pp. 699–715).

Sundermeyer, M., Durner, M., Puang, E. Y., Marton, Z.-C., Vaskevicius, N., Arras, K. O., & Triebel, R. (2020). Multi-path learning for object pose estimation across domains. In *CVPR*, (pp. 13916–13925).

Tan, T., & Dong, Q. (2023). Smoc-net: Leveraging camera pose for self-supervised monocular object pose estimation. In *CVPR*, (pp. 21307–21316).

Tejani, A., Tang, D., Kouskouridas, R., & Kim, T.-K. (2014). Latent-class hough forests for 3d object detection and pose estimation. In *ECCV*, (pp. 462–477).

Tekin, B., Sinha, S. N., & Fua, P. (2018). Real-time seamless single shot 6d object pose prediction. In *CVPR*, (pp. 292–301).

Thalhammer, S., Leitner, M., Patten, T., & Vincze, M. (2021). Pyrapose: Feature pyramids for fast and accurate object pose estimation under domain shift. In *ICRA*, (pp. 13909–13915).

Thalhammer, S., Weibel, J. B., Vincze, M., & Garcia-Rodriguez, J. (2023). Self-supervised vision transformers for 3d pose estimation of novel objects. *Image Vision Comput, 139*, Article 104816.

Tian, M., Ang, M. H., & Lee, G. H. (2020a). Shape prior deformation for categorical 6d object pose and size estimation. In *ECCV*, (pp. 530–546).

Tian, M., Pan, L., Ang, M. H., & Lee, G. H.(2020b). Robust 6d object pose estimation by learning rgb-d features. In *ICRA*, (pp. 6218–6224).

Tremblay, J., To, T., Sundaralingam, B., Xiang, Y., Fox, D., & Birchfield, S. (2018). Deep object pose estimation for semantic robotic grasping of household objects. arXiv:1809.10790.

Tremblay, J., Wen, B., Blukis, V., Sundaralingam, B., Tyree, S., & Birchfield, S. (2023). Diff-dope: Differentiable deep object pose estimation. arXiv preprint arXiv:2310.00463

Tyree, S., Tremblay, J., To, T., Cheng, J., Mosier, T., Smith, J., & Birchfield, S. (2022). 6-dof pose estimation of household objects for robotic manipulation: An accessible dataset and benchmark. In *IROS*, (pp. 13081–13088).

Ulmer, M., Durner, M., Sundermeyer, M., Stoiber, M., & Triebel, R. (2023). 6d object pose estimation from approximate 3d models for orbital robotics. In *IROS*, (pp. 10749–10756).

Ultralytics. (2022). GitHub - ultralytics/yolov5. https://github.com/ultralytics/yolov5

Umeyama, S. (1991). Least-squares estimation of transformation parameters between two point patterns. *IEEE TPAMI, 13*(04), 376–380.

UnrealEngine4. (2020). Unreal engine 4. https://www.unrealengine.com

Wada, K., Sucar, E., James, S., Lenton, D., & Davison, A. J. (2020). Morefusion: Multi-object reasoning for 6d pose estimation from volumetric fusion. In *CVPR*, (pp. 14540–14549).

Wan, B., Shi, Y., & Xu, K. (2023). Socs: Semantically-aware object coordinate space for category-level 6d object pose estimation under large shape variations. In *ICCV*, (pp. 14065–14074).

Wang, C., Xu, D., Zhu, Y., Martín-Martín, R., Lu, C., Fei-Fei, L., & Savarese, S. (2019a). Densefusion: 6d object pose estimation by iterative dense fusion. In *CVPR*, (pp. 3343–3352).

Wang, C., Martín-Martín, R., Xu, D., Lv, J., Lu, C., Fei-Fei, L., Savarese, S., & Zhu, Y. (2020a). 6-pack: Category-level 6d pose tracker with anchor-based keypoints. In *ICRA*, (pp. 10059–10066).

Wang, D., Zhou, G., Yan, Y., Chen, H., & Chen, Q. (2021). Geopose: Dense reconstruction guided 6d object pose estimation with geometric consistency. *IEEE TMM, 24*, 4394–4408.

Wang, F., Zhang, X., Chen, T., Shen, Z., Liu, S., & He, Z. (2023). Kvnet: An iterative 3d keypoints voting network for real-time 6-dof object pose estimation. *Neurocomputing, 530*, 11–22.

Wang, G., Manhardt, F., Shao, J., Ji, X., Navab, N., & Tombari, F. (2020b). Self6d: Self-supervised monocular 6d object pose estimation. In *ECCV*, (pp. 108–125).

Wang, G., Manhardt, F., Liu, X., Ji, X., Tombari, F., & Ji, X. (2021). Occlusion-aware self-supervised monocular 6d object pose estimation. *IEEE TPAMI, 46*(3), 1788–1803.

Wang, G., Manhardt, F., Tombari, F., & Ji, X. (2021c). Gdr-net: Geometry-guided direct regression network for monocular 6d object pose estimation. In *CVPR*, (pp. 16611–16621).

Wang, H., Sridhar, S., Huang, J., Valentin, J., Song, S., & Guibas, L. J. (2019b) Normalized object coordinate space for category-level 6d object pose and size estimation. In *CVPR*, (pp. 2642–2651).

Wang, H., Li, W., Kim, J., & Wang, Q. (2022a). Attention-guided rgb-d fusion network for category-level 6d object pose estimation. In *IROS*, (pp. 10651–10658).

Wang, H., Fan, Z., Zhao, Z., Che, Z., Xu, Z., Liu, D., Feng, F., Huang, Y., Qiao, X., & Tang, J. (2023b). Dtf-net: Category-level pose estimation and shape reconstruction via deformable template field. In *ACM MM*, (pp. 3676–3685).

Wang, J., Chen, K., & Dou, Q. (2021d). Category-level 6d object pose estimation via cascaded relation and recurrent reconstruction networks. In *IROS*, (pp. 4807–4814).

Wang, P., Jung, H., Li, Y., Shen, S., Srikanth, R. P., Garattoni, L., Meier, S., Navab, N., & Busam, B. (2022b). Phocal: A multi-modal dataset for category-level object pose estimation with photometrically challenging objects. In *CVPR*, (pp. 21222–21231).

Wang, R., Wang, X., Li, T., Yang, R., Wan, M., & Liu, W. (2023c). Query6dof: Learning sparse queries as implicit shape prior for category-level 6dof pose estimation. In *ICCV*, (pp. 14055–14064)

Wang, S., Wang, S., Jiao, B., Yang, D., Su, L., Zhai, P., Chen, C., & Zhang, L. (2022c) Ca-spacenet: Counterfactual analysis for 6d pose estimation in space. In *IROS*, (pp. 10627–10634).

Wang, T., Hu, G., & Wang, H. (2024). Object pose estimation via the aggregation of diffusion features. In *CVPR*, (pp. 10238–10247).

Wei, J., Song, X., Liu, W., Kneip, L., Li, H., & Ji, P. (2023). Rgb-based category-level object pose estimation via decoupled metric scale recovery. arXiv:2309.10255.

Wei, J., Song, X., Liu, W., Kneip, L., Li, H., & Ji, P. (2024). Rgb-based category-level object pose estimation via decoupled metric scale recovery. In *ICRA*, (pp. 2036–2042).

Weinzaepfel, P., Leroy, V., Lucas, T., Brégier, R., Cabon, Y., Arora, V., Antsfeld, L., Chidlovskii, B., Csurka, G., & Revaud, J. (2022). Croco: Self-supervised pre-training for 3d vision tasks by cross-view completion. In *NeurIPS*, (pp. 3502–3516).

Wen, B., & Bekris, K. (2021). Bundletrack: 6d pose tracking for novel objects without instance or category-level 3d models. In *IROS*, (pp. 8067–8074).

Wen, B., Mitash, C., Ren, B., & Bekris, K. E. (2020). Se(3)-tracknet: Data-driven 6d pose tracking by calibrating image residuals in synthetic domains. In *IROS*, (pp. 10367–10373).

Wen, B., Tremblay, J., Blukis, V., et al (2023). Bundlesdf: Neural 6-dof tracking and 3d reconstruction of unknown objects. In *CVPR*, (pp. 606–617).

Wen, B., Yang, W., Kautz, J., & Birchfield, S. (2024). Foundationpose: Unified 6d pose estimation and tracking of novel objects. In *CVPR*, (pp. 17868–17879).

Wen, Y., Fang, Y., Cai, J., Tung, K., & Cheng, H. (2021). Gccn: Geometric constraint co-attention network for 6d object pose estimation. In *ACM MM*, (pp. 2671–2679).

Wen, Y., Li, X., Pan, H., Yang, L., Wang, Z., Komura, T., & Wang, W. (2022). Disp6d: Disentangled implicit shape and pose learning for scalable 6d pose estimation. In *ECCV*, (pp. 404–421).

Weng, Y., Wang, H., Zhou, Q., Qin, Y., Duan, Y., Fan, Q., Chen, B., Su, H., & Guibas, L. J. (2021). Captra: Category-level pose tracking for rigid and articulated objects from point clouds. In *ICCV*, (pp. 13209–13218).

Wohlhart, P., & Lepetit, V. (2015). Learning descriptors for object recognition and 3d pose estimation. In *CVPR*, (pp. 3109–3118).

Wu, C., Chen, L., He, Z., & Jiang, J. (2021). Pseudo-siamese graph matching network for textureless objects' 6-d pose estimation. *IEEE TIE, 69*(3), 2718–2727.

Wu C, Chen L, Wang S, Yang H, Jiang J (2023) Geometric-aware dense matching network for 6d pose estimation of objects from rgb-d images. PR 137:109293

Wu, J., Zhou, B., Russell, R., Kee, V., Wagner, S., Hebert, M., Torralba, A., & Johnson, D. M. S. (2018). Real-time object pose estimation with pose interpreter networks. In *IROS*, (pp. 6798–6805).

Wu, J., Wang, Y., & Xiong, R. (2021b). Unseen object pose estimation via registration. In *RCAR*, (pp. 974–979).

Wu, Y., Zand, M., Etemad, A., & Greenspan, M. (2022). Vote from the center: 6 dof pose estimation in rgb-d images by radial keypoint voting. In *ECCV*, (pp. 335–352).

Xiang, Y., Schmidt, T., Narayanan, V., & Fox, D. (2017). Posecnn: A convolutional neural network for 6d object pose estimation in cluttered scenes. arXiv:1711.00199.

Xu, L., Qu, H., Cai, Y., & Liu, J. (2024a). 6d-diff: A keypoint diffusion framework for 6d object pose estimation. In *CVPR*, (pp. 9676–9686).

Xu, Y., Lin, K. Y., Zhang, G., Wang, X., & Li, H. (2024). Rnnpose: 6-dof object pose estimation via recurrent correspondence field estimation and pose optimization. *IEEE TPAMI, 46*(7), 4669–4683.

Xu, Z., Zhang, Y., Chen, K., & Jia, K. (2022). Bico-net: Regress globally, match locally for robust 6d pose estimation. In *IJCAI*, (pp. 1509–1515).

Yang, H., Sun, W., Liu, J., Zheng, J., Dai, Z., & Mian, A. (2025). Lancope: Language-guided category-level object pose estimation from a single rgb image. *IEEE RAL, 10*(7), 7555–7562.

Yang, H., Sun, W., Liu, J., Zheng, J., Xiao, J., & Mian, A. (2025b). Occlusion-aware 3d hand-object pose estimation with masked autoencoders. arXiv:2506.10816.

Yang, H., Sun, W., Liu, J., Zheng, J., Zeng, Z., & Mian, A. (2025). Rgb-based category-level object pose estimation via depth recovery and adaptive refinement. *IEEE RAL, 10*(6), 5377–5384.

Yang, Z., Yu, X., & Yang, Y. (2021). Dsc-posenet: Learning 6dof object pose estimation via dual-scale consistency. In *CVPR*, (pp. 3907–3916).

Yen-Chen, L., Florence, P., Barron, J. T., Rodriguez, A., Isola, P., & Lin, T.-Y. (2021) inerf: Inverting neural radiance fields for pose estimation. In *IROS*, (pp. 1323–1330).

You, Y., Shi, R., Wang, W., & Lu, C. (2022). Cppf: Towards robust category-level 9d pose estimation in the wild. In *CVPR*, (pp. 6866–6875).

Yu, S., Zhai, D. H., Guan, Y., & Xia, Y. (2023a). Category-level 6-d object pose estimation with shape deformation for robotic grasp detection. *IEEE TNNLS*.

Yu, S., Zhai, D. H., & Xia, Y. (2023). Robotic grasp detection based on category-level object pose estimation with self-supervised learning. *IEEE TMEC, 29*(1), 625–635.

Yu, S., Zhai, D. H., Xia, Y., Li, D., & Zhao, S. (2023). Cattrack: Single-stage category-level 6d object pose tracking via convolution and vision transformer. *IEEE TMM, 26*, 1665–1680.

Yu, S., Zhai, D. H., & Xia, Y. (2024). Catformer: Category-level 6d object pose estimation with transformer. In *AAAI*, (pp. 6808–6816).

Yu, X., Zhuang, Z., Koniusz, P., & Li, H. (2020). 6dof object pose estimation via differentiable proxy voting loss. arXiv preprint arXiv:2002.03923.

Zaccaria, M., & Manhardt, F. (2023). Self-supervised category-level 6d object pose estimation with optical flow consistency. *IEEE RAL, 8*(5), 2510–2517.

Zakharov S, Shugurov I, Ilic S (2019) Dpod: 6d pose object detector and refiner. In: ICCV, pp 1941–1950

Ze Y, Wang X (2022) Category-level 6d object pose estimation in the wild: A semi-supervised learning approach and a new dataset. In: NeurIPS, pp 27469–27483

Zeng, L., Lv, W. J., Dong, Z. K., & Liu, Y. J. (2021). Ppr-net++: Accurate 6-d pose estimation in stacked scenarios. *IEEE TASE, 19*(4), 3139–3151.

Zeng, L., Lv, W. J., Zhang, X. Y., & Liu, Y. J. (2021b) Parametricnet: 6dof pose estimation network for parametric shapes in stacked scenarios. In *ICRA*, (pp. 772–778).

Zhang, H., & Cao, Q. (2019). Detect in rgb, optimize in edge: Accurate 6d pose estimation for texture-less industrial parts. In *ICRA*, (pp. 3486–3492).

Zhang, H., Opipari, A., Chen, X., Zhu, J., Yu, Z., & Jenkins, Odest C. (2022a). Transnet: Category-level transparent object pose estimation. In *ECCV*, (pp. 148–164).

Zhang, J., Wu, M., & Dong, H. (2023). Generative category-level object pose estimation via diffusion models. In *NeurIPS*, (pp. 54627–54644).

Zhang, K., & Fu, Y. (2023). Self-supervised geometric correspondence for category-level 6d object pose estimation in the wild. In *ICLR*.

Zhang, R., Di, Y., Lou, Z., Manhardt, F., Tombari, F., & Ji, X. (2022b). Rbp-pose: Residual bounding box projection for category-level pose estimation. In *ECCV*, (pp. 655–672).

Zhang, R., Di, Y., Manhardt, F., Tombari, F., & Ji, X. (2022c). Ssp-pose: Symmetry-aware shape prior deformation for direct category-level object pose estimation. In *IROS*, (pp. 7452–7459).

Zhang, S., Zhao, W., Guan, Z., Peng, X., & Peng, J. (2021). Keypoint-graph-driven learning framework for object pose estimation. In *CVPR*, (pp. 1065–1073).

Zhang, Y., Wu, Z., Peng, H., & Lin, S. (2020a). A transductive approach for video object segmentation. In *CVPR*, (pp. 6949–6958).

Zhang, Y., Zhang, C., Rosenberger, M., & Notni, G. (2020b). 6d object pose estimation algorithm using preprocessing of segmentation and keypoint extraction. In *I2MTC*, (pp. 1–6).

Zhang, Z., Chen, W., Zheng, L., Leonardis, A., & Chang, H. J. (2022d). Trans6d: Transformer-based 6d object pose estimation and refinement. In *ECCV*, (pp. 112–128).

Zhao, C., Hu, Y., & Salzmann, M. (2022). Fusing local similarities for retrieval-based 3d orientation estimation of unseen objects. In *ECCV*, (pp. 106–122).

Zhao, C., Hu, Y., & Salzmann, M. (2024). Locposenet: Robust location prior for unseen object pose estimation. In *3DV*, (pp. 1072–1081).

Zhao, H., Wei, S., Shi, D., Tan, W., Li, Z., Ren, Y., Wei, X., Yang, Y., & Pu, S. (2023). Learning symmetry-aware geometry correspondences for 6d object pose estimation. In *ICCV*, (pp. 14045–14054).

Zhao, W., Zhang, S., Guan, Z., Zhao, W., Peng, J., & Fan, J. (2020). Learning deep network for detecting 3d object keypoints and 6d poses. In *CVPR*, (pp. 14134–14142).

Zheng, L., Wang, C., Sun, Y., Dasgupta, E., Chen, H., Leonardis, A., Zhang, W., & Chang, H. J. (2023). Hs-pose: Hybrid scope feature extraction for category-level object pose estimation. In: CVPR, (pp. 17163–17173)

Zheng, L., Tse, T. H. E., Wang, C., Sun, Y., Chen, H., Leonardis, A., Zhang, W., & Chang, H. J. (2024). Georef: Geometric alignment across shape variation for category-level object pose refinement. In *CVPR*, (pp. 10693–10703).

Zhou, G., Yan, Y., Wang, D., & Chen, Q. (2020). A novel depth and color feature fusion framework for 6d object pose estimation. *IEEE TMM, 23*, 1630–1639.

Zhou, G., Wang, D., Yan, Y., Chen, H., & Chen, Q. (2021). Semi-supervised 6d object pose estimation without using real annotations. *IEEE TCSVT, 32*(8), 5163–5174.

Zhou, G., Wang, H., Chen, J., & Huang, D. (2021b). Pr-gcn: A deep graph convolutional network with point refinement for 6d pose estimation. In *ICCV*, (pp. 2793–2802).

Zhou, J., Chen, K., Xu, L., Dou, Q., & Qin, J. (2023a). Deep fusion transformer network with weighted vector-wise keypoints voting for robust 6d object pose estimation. In *ICCV*, (pp. 13967–13977).

Zhou, J., Zhu, Q., Wang, Y., Feng, M., Liu, J., Huang, J., & Mian, A. (2025). A state space model for multiobject full 3-d information estimation from rgb-d images. *IEEE TCYB, 55*(5), 2248–2260.

Zhou, L., Liu, Z., Gan, R., Wang, H., & Ang, M. H. (2023b). Dr-pose: A two-stage deformation-and-registration pipeline for category-level 6d object pose estimation. In *IROS*, (pp. 1192–1199).

Zhu, Z., Wang, J., Qin, Y., Sun, D., Jampani, V., & Wang, X. (2024). Contactart: Learning 3d interaction priors for category-level articulated object and hand poses estimation. (pp. 201–212).

Zhuang, C., Wang, H., & Ding, H. (2024). Attentionvote: A coarse-to-fine voting network of anchor-free 6d pose estimation on point cloud for robotic bin-picking application. *Robot Cim-int Manuf, 86*, Article 102671.

Zou, L., Huang, Z., Gu, N., & Wang, G. (2022). 6d-vit: Category-level 6d object pose estimation via transformer-based instance representation learning. *IEEE TIP, 31*, 6907–6921.

Zou, L., Huang, Z., Gu, N., & Wang, G. (2023). Gpt-cope: A graph-guided point transformer for category-level object pose estimation. *IEEE TCSVT, 34*(4), 2385–2398.